\documentclass[10pt,twocolumn,letterpaper]{article}

\usepackage{cvpr}  % switch to [final] for camera-ready

\usepackage{graphicx}
\usepackage{amsmath}
\usepackage{amssymb}
\usepackage{amsthm}
\usepackage{booktabs}
\usepackage{multirow}
\usepackage{xcolor}
\usepackage{subcaption}
\usepackage{xspace}
\usepackage{bm}
\usepackage{enumitem}
\usepackage{makecell}

\newif\ifshowplaceholders
\showplaceholdersfalse

\newcommand{\ci}[2]{[#1, #2]}           % confidence interval

\theoremstyle{plain}
\newtheorem{proposition}{Proposition}
\theoremstyle{remark}
\newtheorem{remark}{Remark}
\AtEndPreamble{\crefname{remark}{Remark}{Remarks}\crefname{proposition}{Prop.}{Props.}}

\makeatletter
\ifdefined\IfFormatAtLeastTF\IfFormatAtLeastTF{2025-06-01}{%
  \RemoveFromHook{build/column/before}[lineno]%
  \AddToHook{build/column/before}{%
    \protected@write\@auxout{}{\string\def\string\@LN@column{\if@firstcolumn2\else1\fi}}}%
}{}\fi
\makeatother

\makeatletter
\newbox\shipfix@labels
\newif\ifshipfix@inlabel
\pretocmd\@EveryShipout@Output{%
  \global\setbox\shipfix@labels\box\@labels
  \if@inlabel\global\shipfix@inlabeltrue\global\@inlabelfalse\else\global\shipfix@inlabelfalse\fi}{}{\PackageError{preamble}{shipout guard not applied}{}}
\apptocmd\@EveryShipout@Output{%
  \global\setbox\@labels\box\shipfix@labels
  \ifshipfix@inlabel\global\@inlabeltrue\fi}{}{\PackageError{preamble}{shipout guard not applied}{}}
\makeatother

\usepackage{pifont}
\usepackage{colortbl}                     % shaded Swap-Pair rows in Tab. 2 and Tab. 3 (2026-11-08, PI)
\definecolor{swapfill}{RGB}{231,240,253}  % light tint of the Swap-Pair blue
\newcommand{\List}{\textsf{List}\xspace}
\newcommand{\PWC}{\textsf{Swap-Pair}\xspace}
\newcommand{\GCCR}{\textsf{GCC-Rating}\xspace}
\newcommand{\Rating}{\textsf{Rating}\xspace}
\newcommand{\IsoList}{\textsf{Iso-List}\xspace}
\newcommand{\IsoPair}{\textsf{Iso-Pair}\xspace}
\newcommand{\ScreenList}{\textsf{Screen+List}\xspace}
\newcommand{\pp}{pp\xspace}
\makeatletter
\newcommand{\fcdef}[2]{\expandafter\def\csname fc@#1\endcsname{#2}}
\newcommand{\fc}[1]{\ifcsname fc@#1\endcsname\csname fc@#1\endcsname\else\PackageError{fc}{Unknown forecast key #1}{}\fi}
\makeatother

\newcommand{\Eone}{E_1}
\newcommand{\beats}{\triangleright}

\newcommand{\figfile}[4]{\IfFileExists{#1}{\includegraphics[width=#2]{#1}}{\fbox{\parbox[c][#3][c]{\dimexpr#2-2\fboxsep-2\fboxrule\relax}{\centering\small\textsf{#4}}}}}
\newcommand{\yes}{\ding{51}}
\newcommand{\no}{\ding{55}}

\theoremstyle{plain}
\newtheorem{lemma}{Lemma}

\AtEndPreamble{\crefname{lemma}{Lemma}{Lemmas}\crefname{corollary}{Cor.}{Cors.}}

\newcommand{\ListAvg}{\textsf{List-avg}\xspace}   % List averaged over all cyclic rotations (4 at N=4, 8 at N=8)
\newcommand{\PairOne}{\textsf{Pair-1}\xspace}     % one-order pairwise win rate (lower index shown first)
\newcommand{\PRP}{\textsf{PRP}\xspace}            % pairwise ranking prompting: both orders, hard wins, paper tie rule

\definecolor{cvprblue}{rgb}{0.21,0.49,0.74}
\usepackage[pagebackref,breaklinks,colorlinks,allcolors=cvprblue]{hyperref}

\def\confName{CVPR}
\def\confYear{2027}

\newenvironment{promptbox}[2]{\par\medskip\noindent\textbf{\small #1}\par\nobreak\noindent{\scriptsize source: #2}\par\nobreak
  \vspace{2pt}\hrule\vspace{3pt}\begingroup\ttfamily\scriptsize\raggedright\frenchspacing\setlength{\parskip}{0pt}}%
  {\par\endgroup\vspace{2pt}\hrule\smallskip}
\newcommand{\pline}[1]{\noindent\hangindent=1.5em\hangafter=1 #1\par}
\newcommand{\mainref}[1]{\cref{#1} of the main paper}
\newcommand{\GCC}{\textsf{GCC}\xspace}
\newcommand{\CHIP}{\textsf{CHIP}\xspace}
\newcommand{\CalPair}{\textsf{CalPair}\xspace}
\newcommand{\PriDe}{\textsf{PriDe}\xspace}

\usepackage[T1]{fontenc}

\title{One Worse Image Flips the Winner: Candidate-Set Sensitivity of Multimodal Judges and Swap-Pair Rewards for Image Generation}

\author{Yu Chang, Anzhe Cheng, Puquan Pan, Heng Ping, Peiyu Zhang, Shahin Nazarian, Paul Bogdan\\[3pt]
University of Southern California}

\begin{document}
% Teaser under the title (standard CVPR pattern).
\twocolumn[{%
\renewcommand\twocolumn[1][]{#1}%
\maketitle
\begin{center}
  \captionsetup{type=figure}
  \figfile{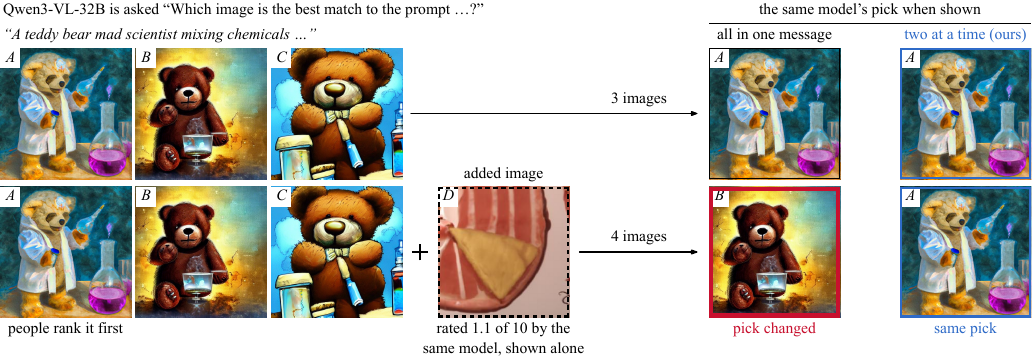}{\textwidth}{6.1cm}{fig1\_L6.pdf: one HPDv2 set; the listwise pick with three and with four images, and the Swap-Pair pick}
  \vspace{-1.5em}
  \captionof{figure}{\textbf{One worse image flips the winner.}
  In this selected HPDv2 example, people rank $A$ first. Qwen3-VL-32B, shown $A$, $B$, $C$ in one message (\List), picks $A$; once $D$ is added, an image made for another prompt that the same judge rates 1.1 of 10 alone, it picks $B$.
  \PWC shows the same judge two images at a time, in both orders, and picks $A$ with and without $D$.
  On five open-weight judges, fifty such images change the \List winner in 22--31 percentage points more sets than fifty near-identical copies of one image (\cref{tab:judges}).}
  \label{fig:teaser}
  \vspace{0.2em}
\end{center}

}]

\begin{abstract}
Multimodal large language models judge text-to-image generation, both to pick the best of several samples and to reward the generative model during fine-tuning.
Pointwise judging rates each image alone, too coarsely to separate close candidates, whereas listwise judging shows all candidates in one message so that the judge can compare them.
We find that adding an image that loses to two candidates, $A$ and $B$, in two-image comparisons can switch the listwise pick from $A$ to $B$, an effect we call \emph{candidate-set sensitivity}.
On each of five open-weight judges, such images change the winner in 22--31 percentage points more candidate sets than near-identical copies of one worse image, a control for chance.
The flaw also appears among the rollouts of one GRPO group, where exchanging a rollout that loses every comparison changes the listwise pick among the others.
To pick better samples and train better generative models with the same judge, we propose \PWC, which keeps the direct comparison of listwise judging but never shows the judge a third image.
It asks about each pair in both orders, drops a candidate whenever some rival beats it and also beats every candidate it beats, and ranks the rest by wins.
Adding such a candidate thus provably leaves the winner unchanged, and candidate-set sensitivity all but disappears on every judge.
In best-of-N selection, the \PWC pick scores higher on HPSv2.1 and under human ratings than the pick of any other way of asking the judge.
As the GRPO reward, \PWC trains a better generative model than the pointwise, listwise and one-order pairwise rewards of the same judge, with an HPSv2.1 win rate over the base model 11.4 points above that of the listwise reward.
People prefer its images, and the lead is larger after longer training and holds at the batch size of a published configuration, on FLUX.1-dev and on benchmarks that no reward optimises.
\end{abstract}

\section{Introduction}
\label{sec:intro}

Text-to-image generative models are now fine-tuned with Group Relative Policy Optimization (GRPO), which scores several samples, or rollouts, of one prompt against each other with a learned reward~\cite{wang2025prefgrpo,liu2025flowgrpo,xue2025dancegrpo,wang2025unifiedreward}.
The reward comes from models that score one image against its prompt~\cite{xu2023imagereward,wu2023hpsv2,kirstain2023pickapic} or from multimodal large language models used as judges~\cite{chen2024mjbench}.
Whether such a judge picks the best of $N$ samples~\cite{seo2026judge} or rewards rollouts, it chooses among the candidate images of one prompt.
Whatever sways that choice besides image quality therefore passes into the selected image and the reward.

Readouts, the ways of asking the judge to choose, differ in how many candidates one message shows, which decides what the judge can compare and what can sway it.
Pointwise rating (\Rating)~\cite{chen2024mjbench} shows the judge one image at a time, but its scores from one to ten are too coarse to separate close candidates.
Pairwise readouts show two images and let the judge compare them directly, but Pref-GRPO~\cite{wang2025prefgrpo} asks about each pair in one order (\PairOne), so the image shown first keeps an advantage.
Pairwise ranking prompting (\PRP)~\cite{qin2024prp} asks in both orders but keeps only a vote from each answer, which lets the top candidates tie.
The listwise readout (\List)~\cite{yang2026only,katkuri2026pairwise,chen2026amber} asks for the winner among all candidates in one message, but its answer depends on how they are presented.
Position~\cite{zheng2023judging,wang2024fair} and option labels~\cite{zheng2024mcq,pezeshkpour2024order} sway LLM judges, and multimodal judges prefer the image shown first~\cite{seo2026judge} and reverse their pick when two images trade places~\cite{huang2026mllm}.
The usual remedy averages the answer over presentation orders~\cite{sargent2026vlic,qin2024prp,zhou2024batch}, here over slot rotations (\ListAvg).

From one call to the next, however, the set of candidates changes too.
A judge that prefers $A$ to $B$ should keep this preference when an image $D$ that it considers worse than both is added.
Sen's condition for rational choice requires this~\cite{sen1971choice}, people violate it~\cite{huber1982adding}, and so does the multimodal judge of \cref{fig:teaser}.
Shown in one listwise message the three images that humans rank highest for a prompt, $A$, $B$ and $C$, Qwen3-VL-32B confidently picks $A$.
We then add an image $D$ that was generated for another prompt, scores 1.1 out of 10 when the same judge rates it alone, and loses to both $A$ and $B$ in pairs.
With $D$ shown alongside them, the judge switches to $B$ with near certainty.
We call this failure \emph{candidate-set sensitivity}.

We present the first measurement of candidate-set sensitivity in multimodal judges and find it in each of five open-weight judges (\cref{sec:measure}).
To each candidate set we add fifty different worse images, one at a time, and record the share of sets in which the winner depends on which of them is added.
This share exceeds the one for fifty near-identical copies of one worse image, which reflect the judge's own noise, by 22--31 percentage points on every judge.
On Qwen3-VL-8B, the gap stays in this range when the candidates are eight SD-XL samples of one prompt from a public benchmark (\cref{sec:measure-groups}).
Among the rollouts of a GRPO group, exchanging one that the judge ranks below all the others changes the listwise pick among the rest in 17.6 points more groups than re-rendering it does.
On both Qwen3-VL judges, the flips survive the usual remedy of order averaging as well as label debiasing and attention blocking between images (\cref{sec:measure-diag}).
Winners built from pair messages, which never show a third image, rarely change when the same images are added.

The readout we seek compares candidates directly and never shows the judge more than two of them.
Pairwise readouts already have this form, yet they keep the order advantage or the ties noted above.

We propose \PWC, which asks about each pair in both orders and sums the judge's two log-odds for one image into a margin, whose sign says which image beats the other (\cref{sec:pairwise}).
In this sum the advantage of the first-shown image cancels, whereas \PairOne's single order keeps it.
Margins can form cycles, so an added candidate can beat the winner while some rival beats it and everything it beats; counting wins or summing margins then lets it move the winner (\cref{fig:anatomy}).
Tournament theory calls such a candidate covered~\cite{miller1980new,laslier1997tournament}, as it does any image that loses to all others, and the cover rule that \PWC applies drops every covered candidate before ranking the rest by their wins.
For fixed pair margins, a covered addition leaves the winner unchanged, and so does a reordering except in an exact tie (Theorem~\ref{thm:select}), a guarantee that \PairOne and \PRP lack.

On all five judges, \PWC makes the flips nearly disappear (\cref{sec:pair-results}).
We follow one judge, Qwen3-VL-8B, into both uses on groups of eight SD3.5-Medium rollouts and take HPSv2.1~\cite{wu2023hpsv2}, a learned preference score independent of the judge, as the yardstick.
In best-of-8 selection the \PWC pick recovers 64.2\% of the best achievable HPSv2.1 gain, against 47.5\% for the listwise and 43.1\% for the pointwise pick, and beats every other readout, order-averaged listwise included (\cref{sec:validity}).
Under human ratings on a public benchmark, its pick again recovers the largest share of the gain of all readouts.
When the same judge instead supplies the GRPO reward, we count a prompt as won if HPSv2.1 scores the fine-tuned model's image above the base model's.
With the \PWC reward this win rate is 11.4 percentage points higher than with the listwise reward and 10.2 points higher than with the pointwise reward (\cref{sec:training}).
\PWC also beats the order-averaged listwise reward and the one-order and tournament pairwise rewards.
It leads the two-order-vote reward and the reward released with Pref-GRPO on its own judge, and, on yardsticks that no reward optimises, the scorer rewards of published GRPO pipelines.
People prefer its images, and its lead is larger after three times longer training and holds at Flow-GRPO's public batch size and KL weight, on FLUX.1-dev, on unseen prompts and on public benchmarks.
How the judge is asked is part of the reward.

\section{Related Work}
\label{sec:related}

\paragraph{Judges and rewards over candidate sets.}
Text-to-image reward models score one image against a prompt~\cite{xu2023imagereward,wu2023hpsv2,kirstain2023pickapic,wang2025unifiedreward}; MLLM judges are benchmarked on scores, pairs and batch rankings~\cite{chen2024mjbench,chen2024mllmasajudge,li2024vlrewardbench,yasunaga2025multimodal}, select the best of several generations~\cite{seo2026judge}, and serve as listwise or group-wise rewards that score $N$ responses jointly~\cite{yang2026only,katkuri2026pairwise,jia2026prioritizing,chen2026amber}.
They build and benchmark judges and rewards; we measure whether a judge's winner depends on the other candidates it sees, and follow the answer into selection and training.

\paragraph{GRPO rewards for text-to-image.}
Flow-GRPO~\cite{liu2025flowgrpo} and DanceGRPO~\cite{xue2025dancegrpo} bring GRPO to flow-matching generators and reward each rollout with a learned scorer such as PickScore or HPSv2.1, or with a task reward.
Pref-GRPO~\cite{wang2025prefgrpo} observes that normalising small pointwise score differences within a group produces illusory advantages that the policy exploits, and rewards each rollout with its win rate under a pairwise judge instead; UnifiedReward~\cite{wang2025unifiedreward} supplies such judges, and Tournament-GRPO~\cite{yang2026tournamentgrpo} builds group rewards from judge tournaments.
These works choose the reward model, or the judge and its prompt; we hold the judge and its prompt fixed, vary only how its answers are read, and compare the readouts with the published rewards in one recipe.

\paragraph{Position, label and candidate-set effects.}
Decoy effects, long known in human choice~\cite{huber1982adding}, appear in instruction-tuned language models~\cite{valkanova2024irrelevant,itzhak2024instructed,liu2024decoy}, and \citet{tripathi2025pairwise} show that a third text response can reverse a text judge's pairwise preference.
Position bias is documented in LLM judges~\cite{zheng2023judging,wang2024fair} and listwise rankers~\cite{zhang2026ranked,frohmann2026equal}, and multiple-choice preferences attach to the option label~\cite{zheng2024mcq,pezeshkpour2024order}.
Multimodal judges favour the first generation~\cite{seo2026judge} or late images~\cite{tian2025identifying,xian2026logitattention}, reverse their pick between two edits under an order swap~\cite{huang2026mllm} and flip under irrelevant context~\cite{omar2026its,zhang2026illusion,sharma2024losing}.
These studies vary order, labels or context on a fixed set, or add a text response for a text judge.
We add images that the judge under test has itself rated below the contenders and measure their effect against a matched noise floor.

\paragraph{Order averaging, pairwise protocols and invariant constructions.}
Order averaging is the standard remedy: VLIC judges both orders of a pair~\cite{sargent2026vlic}, and batch calibration~\cite{zhou2024batch} and PriDe~\cite{zheng2024mcq} correct an estimated prior.
Architectural fixes isolate the candidates from one another by shared positions~\cite{mcilroyyoung2024orderindependence,egressy2025setllm}, blocked attention~\cite{bito2026one,amit2026permutationrobust} or noise in place of the other images~\cite{park2026mitigating}; blocked attention keeps the excess we measure, and noise in place of the other images lets the added image itself win (\cref{sec:measure-diag}).
Pairwise protocols are closest to ours.
\PRP ranks text with both-order pairwise prompts aggregated by sorting or win counts~\cite{qin2024prp}; Pref-GRPO replaces pointwise reward scores with each rollout's within-group win rate under a pairwise judge~\cite{wang2025prefgrpo}, querying each unordered pair once in its released implementation (Supp.~A); Tournament-GRPO turns judge tournaments among rollouts into group rewards for text generation~\cite{yang2026tournamentgrpo}.
The cover rule comes from tournament theory~\cite{miller1980new,dutta1988covering,laslier1997tournament}.
\PWC differs from all three in what it does with the pair answers: it queries both orders of every pair and removes covered candidates before any score is read, so a covered addition leaves its winner unchanged (\cref{thm:select}), a guarantee that sorting, win counts or a one-order win rate lack.

\section{Measuring Candidate-Set Sensitivity}
\label{sec:measure}

\subsection{Readouts, property and endpoints}
\label{sec:measure-setup}

\paragraph{Readouts.}
A judge receives a prompt and $N$ candidates in labelled slots (prompts in Supp.~A), and a readout turns the probabilities of its first answer token over the slot labels into a winner.
The listwise readout \List shows all $N$ in one message and picks the most probable label; \ListAvg averages these probabilities per image over every cyclic rotation of the slots.
A pair message shows two candidates and yields the log-odds of the first label over the second.
Following Pref-GRPO's released implementation~\cite{wang2025prefgrpo}, \PairOne asks each unordered pair once, lower index first, and scores a win 1, a loss 0 and a tie $\tfrac12$; \PRP~\cite{qin2024prp} asks both orders and counts hard votes, while \PWC sums both orders' log-odds into a margin and selects with a cover rule (\cref{sec:pairwise}).

\paragraph{The property.}
Sen's condition $\alpha$~\cite{sen1971choice} forbids a candidate $d\notin S$ that the judge itself ranks below the contenders $A$ and $B$ to move the winner among the contenders $A$, $B$, $C$ of $S$: $w(S\cup\{d\})\in\{w(S),d\}$.
A rubric may legitimately re-weigh $A$ against $B$ when a strong $d$ arrives, so we add only \emph{dominated decoys}, verified by the judge under test: same-prompt images that humans rank below $A$ and $B$ or human-last images of other prompts, which the judge rates alone at least one point below both and prefers $A$ and $B$ over in both pairwise orders.
Every flipping decoy is re-verified and leaves the pool if it fails (Supp.~B).

\paragraph{Endpoints and judges.}
\emph{Fifty tries} (\cref{tab:judges}) ask whether the winner depends on which image is added: for a family $\mathcal{P}$ of up to fifty fourth images for set $S$, $f(S;\mathcal{P})=1$ if the winner is not the same for every $d\in\mathcal{P}$.
Read on fifty re-renders of one fixed decoy (Supp.~B), the same $f$ gives the noise floor, the judge's own noise under fifty near-identical additions.
The primary endpoint is the \emph{searched excess} $\Eone=\frac{1}{|\mathcal{S}|}\sum_{S}\big[f(S;\text{decoys})-f(S;\text{re-renders})\big]$, which is 0 for an ideal judge and for a judge whose flips are pure noise (95\% bootstrap intervals; further endpoints in Supp.~D).
HPDv2's test split~\cite{wu2023hpsv2} supplies 400 prompts with nine generated images each under a full human ranking, 300 for development and 100 held out; the contenders $A$, $B$, $C$ are the images of human rank 1--3, and the fourth slot, balanced over positions, holds the added image.
The study covers five open-weight judges (closed-model check in Supp.~D): Qwen3-VL-8B and 32B~\cite{bai2025qwen3vl} (q8, q32), InternVL3.5-38B~\cite{wang2025internvl35} (ivl38), Mistral-Small-3.1-24B~\cite{mistral2025small} (mst24) and UnifiedReward-qwen-7B~\cite{wang2025unifiedreward} (ur7), from the reward-model family of Pref-GRPO, all under one family-agnostic decision prompt.

\begin{table}[t]
  \centering
  \scriptsize
  \setlength{\tabcolsep}{4pt}
  \caption{\textbf{Decoys flip listwise judges 22 to 31\,\pp above the noise floor.}
  HPDv2 development sets; \% of sets whose \List winner depends on which of fifty decoys, or of fifty re-renders, is added; $\Eone$ with 95\% interval. Last column: $\Eone$ of \PWC.}
  \label{tab:judges}
  \begin{tabular}{@{}l cc l c@{}}
    \toprule
    & \multicolumn{3}{c}{Fifty tries, \List} & \PWC \\
    \cmidrule(lr){2-4}\cmidrule(lr){5-5}
    Judge & decoys & re-renders & $\Eone$ [95\% CI] & $\Eone$ \\
    \midrule
    q8    & 40.1 & 9.0  & $+31.1$ [$+26.0$, $+36.7$] & 1.4 \\
    q32   & 39.5 & 17.1 & $+22.4$ [$+17.4$, $+27.4$] & 1.3 \\
    ivl38 & 33.6 & 10.1 & $+23.5$ [$+18.8$, $+28.5$] & 0.0 \\
    mst24 & 45.3 & 18.1 & $+27.2$ [$+21.8$, $+32.6$] & 0.3 \\
    ur7   & 46.4 & 17.6 & $+28.9$ [$+23.0$, $+34.7$] & 0.4 \\
    \bottomrule
  \end{tabular}
\end{table}

\subsection{Fifty tries against the noise floor}
\label{sec:measure-add}

With fifty tries, dominated decoys move the listwise winner in 22 to 31\,\pp more sets than fifty re-renders do, on every judge (\cref{tab:judges}).
These flips run between contenders: the decoy itself wins in 0.1\% of q32 trials and is never a set's most frequent winner.
The excess replicates for q32 on the held-out prompts and also appears for q8 with text decoys on VL-RewardBench~\cite{li2024vlrewardbench}.
What else the message shows moves the listwise winner: one added decoy moves it about as often as one re-render or a grey square, and unscreened natural images as often as decoys (Supp.~D); dominance makes each such move a violation of $\alpha$, and fifty tries find the decoys that make it.

\subsection{Real groups: reordering and added images}
\label{sec:measure-groups}

Even with nothing added, reordering the images of one listwise message changes which rollout is selected and which rollouts are pushed up or down.
On 128 audit groups of eight SD3.5-Medium rollouts judged by q8 (Supp.~E), a seeded permutation changes \List's top-ranked rollout in 62.5\% of groups (\cref{tab:protocols}).
The same permutation flips the sign of at least one rollout's rank-based GRPO advantage in 89\% of groups.
People see the change: on GenAI-Rank, when reordering changes \List's pick, the two picked images differ by 0.48 points of mean human rating on a five-point scale on which the best image of a prompt beats a random one by 0.62 (Supp.~E).

Exchanging one rollout changes the pick as well.
For each audit group we fill the eighth slot with other rollouts of the same prompt that the judge rates alone at least a point below the top two of the seven kept ones and that lose both pair orders against each of the seven, and read the fifty-tries endpoint against fifty re-renders of one of them.
In the 119 groups with at least ten such rollouts, which of them is drawn decides \List's pick among the other seven in 17.6\,\pp more groups than the re-renders do, and \ListAvg's in 11.8\,\pp (\cref{tab:protocols}, Exchange).
Across these exchanges, on average 1.05 more of the seven kept rollouts change the sign of their \List advantage than across the re-renders (Supp.~E).

The searched excess of \cref{sec:measure-add} also appears in public groups.
Here the candidates are eight of the nine SD-XL samples of a GenAI-Rank prompt~\cite{li2024genaibench}, a public benchmark with human ratings per image, and the dominated decoys are other prompts' human-last images, verified against all eight; 243 of the 256 prompts drawn for this test have at least ten such decoys.
Under q8, $\Eone=+24.3$ \ci{+18.5}{+30.0}\,\pp, rising with the number of tries, and averaging each message over eight orders leaves most of it (Supp.~D).

\subsection{Presentation remedies leave the excess}
\label{sec:measure-diag}

Order averaging, label debiasing and blocked attention each leave the excess: \ListAvg keeps $+24.7$\,\pp on q32, PriDe~\cite{zheng2024mcq} $+30.8$ on q8 and InvariRank's block attention mask~\cite{bito2026one} 110\% of \List's excess on q8, while noise in place of the other images, as in FOCUS~\cite{park2026mitigating}, lets the added image itself win (Supp.~D).
A pair message never shows a third image: asked in pairs, q8 and q32 leave a searched excess of at most 3.8\,\pp on the same sets (Supp.~D), and what remains is to aggregate the pair answers into one winner (\cref{sec:pairwise}).

\theoremstyle{plain}
\newtheorem{theorem}{Theorem}

\section{\PWC: Pairwise Margins with a Cover Rule}
\label{sec:pairwise}

\PWC turns the judge's answers to two-image messages into a two-order margin per pair, selects with the cover rule of tournament theory~\cite{miller1980new,laslier1997tournament} and rewards each GRPO rollout by its Copeland score.

\subsection{The rule}
\label{sec:rule}

A pair message shows no third candidate, but asked in one order, as by \PairOne, it favours the first-shown image on q8 (\cref{sec:training}).
Asking about each pair $\{x,y\}$ in both orders, \PWC reads the two-order margin $m(x,y)$, the log-odds for $x$ summed over the two messages, so $m(y,x)=-m(x,y)$ and a constant first-slot preference cancels.
Each message holds only $x$ and $y$, so $m(x,y)$ depends on $x$, $y$ and the judge, not on the rest of the set (hypothesis~H).
H holds by construction for one forward pass per message, as in training (batched serving in Supp.~C).
$x$ \emph{beats} $y$, $x\beats y$, when $m(x,y)>0$; a zero or unreadable margin is a tie.

\begin{figure}[t]
  \centering
  \figfile{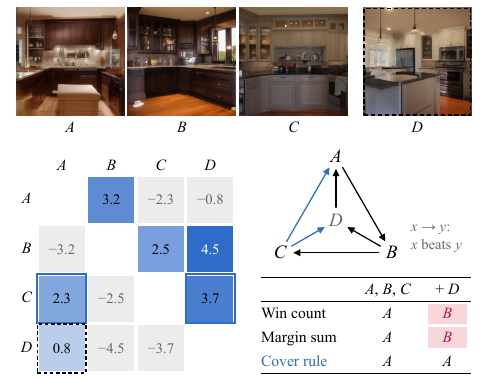}{\linewidth}{6.0cm}{figC\_v9.pdf: four samples of one prompt, two-order margin matrix, tournament with D covered by C, winners of three rules}
  \vspace{-1.5em}
  \caption{\textbf{A covered image that beats the winner.}
  Four SD3.5-Medium samples of one prompt, judged pairwise by q8. \emph{Left:}~two-order margins (blue: row wins). \emph{Right:}~$A$, $B$, $C$ form a cycle; $D$ beats $A$ but $C$ covers it. The cover rule picks $A$ with or without $D$; counting wins or summing margins picks $B$ once $D$ is added (red).}
  \label{fig:anatomy}
  \vspace{-0.6em}
\end{figure}

Margins can form cycles, and a candidate can beat the winner of the others while a rival beats it and everything it beats.
Counting wins or summing margins over all candidates then lets its addition move the winner; the cover rule removes such candidates first (\cref{fig:anatomy}).
In a set $T$, $y$ \emph{covers} $x$ when it beats $x$ and everything $x$ beats,
\begin{equation}
  y\beats x\quad\text{and}\quad \forall z\in T:\ x\beats z\Rightarrow y\beats z, \label{eq:cover}
\end{equation}
the downward covering relation of a weak tournament~\cite{bordes1983possibility,duggan2013uncovered}.
The winner $w(T)$ is read in three steps: (1) remove every covered candidate at once and repeat on the survivors until none is covered; (2) score each survivor by Copeland among the survivors, one point per win and half per tie; (3) break a top tie by each tied survivor's summed margins against the other tied survivors, then by the lowest slot.

\subsection{Selection}
\label{sec:select}

\begin{theorem}[Selection]\label{thm:select}\label{prop:select}
Fix the margins among $S$ (H), and let $S\cup\{d\}$ hold the members of $S$ in the same relative slot order as $S$.
(a) If $d$ is covered in $S\cup\{d\}$, then $w(S\cup\{d\})=w(S)$, and no margin involving $d$ enters the Copeland or tie-break step; in particular for every $d$ that loses every pair, ties among the others allowed.
(b) For any relabelling $\pi$ of the slots, $w(\pi S)=\pi\,w(S)$: exact unless both the Copeland score and the margin sum among the tied survivors tie exactly, so that the lowest slot decides.
\end{theorem}

Covering is transitive and survives the removal of another covered candidate, so every order of single removals ends at the same survivors, the iterated uncovered set~\cite{laslier1997tournament,brandt2008computing}; removing $d$ first is one such order, and the later steps read survivors only.
Part (a) is thus the known stability of this set under a covered entrant~\cite{laslier1997tournament}, here for weak tournaments with unreadable margins and with the tie-break over survivors (Supp.~C).
Only the lowest-slot tie-break depends on the labels, so (b) can fail only on the double tie, which occurred in none of the 384 stored groups, 256 selection and 128 audit groups (Supp.~C).
Only the cover step carries (a) beyond a universal loser, a candidate that loses every pair: every other aggregation we enumerated moves the contender winner for some covered candidate (Supp.~C).

\subsection{The training reward}
\label{sec:reward}\label{prop:reward}

GRPO needs a score for every rollout, so the \PWC reward is each rollout's Copeland score over its group; the cover step serves selection and is not part of the reward.
If a candidate enters and loses every pair while the margins among the rollouts are unchanged, all Copeland rewards go through one positive affine map, so the rollouts keep their ranking, and their advantages, normalised over the original rollouts, change only through the small constant added to the standard deviation (Supp.~C).
The exchanged rollouts of \cref{sec:measure-groups} are such candidates, so the \PWC reward ranks the seven kept rollouts the same whichever of them is drawn (\cref{tab:protocols}).

\begin{table}[t]
  \centering
  \scriptsize
  \setlength{\tabcolsep}{2.5pt}
  \renewcommand{\arraystretch}{1.06}
  \caption{\textbf{Readouts of q8: \PWC is the stable readout and picks best.}
  Msgs: judge messages per group of eight. Exch.: excess \% of groups whose pick depends on which dominated rollout is exchanged in; Reorder: \% whose pick changes under a permutation (\cref{sec:measure-groups}); zeros hold by construction. $\rho$: Spearman correlation with HPSv2.1. Best-of-$N$: \% of the oracle gain, $N=8$ on 256 SD3.5-M groups (HPSv2.1) and $N=9$ on GenAI-Rank (human ratings). Best-of-8, \PWC minus each other readout: $+6.2$ to $+21.0$ points, every 95\% interval above 0.}
  \label{tab:protocols}
  \begin{tabular}{@{}l r r r r r r@{}}
    \toprule
    & & \multicolumn{3}{c}{SD3.5-M audit groups} & \multicolumn{2}{c}{Best-of-$N$} \\
    \cmidrule(lr){3-5}\cmidrule(lr){6-7}
    Readout & Msgs & Exch.$\downarrow$ & Reorder$\downarrow$ & $\rho\uparrow$ & HPSv2.1$\uparrow$ & People$\uparrow$ \\
    \midrule
    \Rating        & 8  & 0.0  & 0.0  & .33 & 43.1 & 31.0 \\
    \List          & 1  & 17.6 & 62.5 & .36 & 47.5 & 27.6 \\
    \ListAvg       & 8  & 11.8 & 25.0 & .39 & 58.0 & 33.8 \\
    \textsf{PSC}~\cite{tang2024found} & 8 & 10.9 & 32.8 & .43 & 57.8 & 33.1 \\
    \PRP           & 56 & 0.0  & 15.6 & .48 & 56.1 & 35.1 \\
    \PairOne       & 28 & 0.0  & 48.4 & .41 & 47.7 & 30.7 \\
    \textsf{Knockout}~\cite{hu2025mmrb2} & 7 & 0.0 & 50.8 & .30 & 48.7 & 29.8 \\
    \textsf{Tourn.}~\cite{yang2026tournamentgrpo} & 21 & 0.0 & 44.7 & .39 & 50.1 & 30.4 \\
    \rowcolor{swapfill}
    \PWC           & 56 & 0.0  & 0.0  & \textbf{.51} & \textbf{64.2} & \textbf{38.0} \\
    \bottomrule
  \end{tabular}
\end{table}

\begin{table*}[t]
  \centering
  \footnotesize
  \setlength{\tabcolsep}{4.2pt}
  \caption{\textbf{GRPO on SD3.5-Medium: \PWC leads every column except a scorer reward's own or a closely related metric (grey).}
  Win rate (\%) against the base model; pinned: mean (sd) over seeds; fresh: 992 held-out prompts; GenEval: score $\times$100 (base 67.0). Bold: best outside grey.}
  \label{tab:training}
  \begin{tabular}{l c c c c c c c}
    \toprule
    & \multicolumn{2}{c}{Pinned HPSv2.1} & \multicolumn{3}{c}{Fresh, 992 prompts} & \multicolumn{2}{c}{Benchmarks} \\
    \cmidrule(lr){2-3}\cmidrule(lr){4-6}\cmidrule(lr){7-8}
    Reward & 200 steps & 600 steps & HPSv2.1 & PickScore & ImageReward & GenEval & VQAScore \\
    \midrule
    \multicolumn{8}{l}{\emph{Readouts of Qwen3-VL-8B (q8): 16 seeds to step 200; 8 more seeds to step 600}} \\
    \Rating, each image alone & 56.2 (6.0) & -- & 53.6 & 53.2 & 55.9 & 68.0 & 49.6 \\
    \List, one message & 55.0 (6.2) & 58.6 & 52.0 & 53.1 & 57.6 & 68.1 & 48.4 \\
    \ListAvg, 8 rotations & 62.1 (5.5) & 67.2 & 58.0 & 54.2 & 58.6 & 68.2 & 50.1 \\
    \PairOne, one order & 62.6 (5.0) & 68.1 & 58.0 & 55.4 & 59.8 & 68.3 & 50.4 \\
    Tournament-GRPO~\cite{yang2026tournamentgrpo} & 60.4 (5.7) & -- & 56.6 & 54.4 & 58.8 & 68.2 & 49.8 \\
    \PRP, two-order votes & 63.2 (5.3) & 69.9 & 59.4 & 56.0 & 60.6 & 68.6 & 51.3 \\
    \rowcolor{swapfill}
    \PWC, Copeland score & \textbf{66.4} (5.0) & \textbf{74.3} & \textbf{62.4} & \textbf{57.8} & \textbf{61.9} & \textbf{69.1} & \textbf{52.6} \\
    \midrule
    \multicolumn{8}{l}{\emph{Readouts of the UnifiedReward judge of Pref-GRPO: 8 seeds}} \\
    Pref-GRPO~\cite{wang2025prefgrpo}, as released & 58.5 (5.6) & -- & 56.9 & 55.6 & 59.6 & 68.5 & 50.7 \\
    \rowcolor{swapfill}
    \PWC, Copeland score & 60.7 (5.1) & -- & 58.9 & 56.2 & 60.4 & 68.4 & 51.4 \\
    \midrule
    \multicolumn{8}{l}{\emph{Published scorer rewards: 8 seeds}} \\
    PickScore, as in Flow-GRPO~\cite{liu2025flowgrpo} & \cellcolor{black!8}68.7 (4.1) & -- & \cellcolor{black!8}65.9 & \cellcolor{black!8}79.3 & 59.6 & 67.5 & 49.2 \\
    HPSv2.1, as in DanceGRPO~\cite{xue2025dancegrpo} & \cellcolor{black!8}83.6 (3.2) & -- & \cellcolor{black!8}80.2 & \cellcolor{black!8}63.1 & 59.1 & 67.3 & 48.6 \\
    UnifiedReward score~\cite{wang2025unifiedreward} & 57.4 (5.8) & -- & 55.1 & 54.0 & 58.7 & 68.7 & 51.5 \\
    \bottomrule
  \end{tabular}
  \vspace{2pt}
  \begin{minipage}{\textwidth}\scriptsize
  Pinned HPSv2.1, \PWC (q8) minus arm, 95\% interval over seeds: \List $+11.4$ [$+7.3$, $+15.5$], \Rating $+10.2$ [$+6.2$, $+14.2$], Tournament-GRPO $+6.0$ [$+2.1$, $+9.9$]; with 24 seeds, \ListAvg $+4.6$ [$+1.6$, $+7.7$] and \PairOne $+3.4$ [$+0.5$, $+6.4$]. Holm-adjusted $p\le.046$ for all five.
  \end{minipage}
\end{table*}

\subsection{Pairwise readouts on the measurement sets}
\label{sec:pair-results}

Read by \PWC, the five judges nearly stop flipping between contenders, with a searched excess of at most 1.4\,\pp (\cref{tab:judges}).
On the GenAI-Rank groups of \cref{sec:measure-groups}, 99.5\% of the added decoys lose all sixteen pair messages against the eight samples, and the \PWC winner is the same for every added decoy in all 243 groups.
On MJ-Bench~\cite{chen2024mjbench}, a pair asked in one order favours the image shown first on every judge, which \PWC's two orders cancel, and the \PWC winner agrees more often than \List's with the human HPDv2 ranking, on q8, on q32 and held out (Supp.~D).

\section{Best-of-N Selection}
\label{sec:consequences}\label{sec:downstream}\label{sec:validity}

With Qwen3-VL-8B (q8) as the judge, each readout picks one of the eight SD3.5-Medium rollouts in each of 256 selection groups, and a pick is scored by the share of the oracle gain it recovers, the HPSv2.1 gain of the best of the eight rollouts over a random one.

The \PWC pick scores higher on HPSv2.1 than the pick of every other readout (\cref{tab:protocols}).
It recovers 64.2\% of the oracle gain, against 47.5\% for one \List message and 43.1\% for \Rating, and it beats the order-correcting readouts \ListAvg, PSC~\cite{tang2024found} and \PRP and the single-order pairwise readouts \PairOne, Knockout~\cite{hu2025mmrb2} and Tournament-GRPO~\cite{yang2026tournamentgrpo}.
Each of the other readouts fails in one of four ways (\cref{tab:protocols}, Supp.~D).
The listwise ones depend on which rollouts share the message and on their order, and a pair asked in one order favours the image shown first.
\PRP removes the order advantage by asking both orders but scores a pair whose two answers disagree as a tie, as 45\% of the HPDv2 contender pairs do on q8, and \Rating's one-image scores separate close rollouts poorly.
\PWC's summed margin orders the pairs that \PRP ties, and its comparisons separate the close pairs that \Rating cannot: on pairs whose ratings differ by less than half a point, \PWC orders more pairs as people do.

\PWC also leads under human ratings.
On GenAI-Rank~\cite{li2024genaibench}, a public benchmark with nine SD-XL and nine DALL-E~3 images per prompt and three human ratings per image, the \PWC pick recovers the largest share of the human-rating gain of all nine readouts (\cref{tab:protocols}, last column).
With Qwen3-VL-32B it also beats the picks of VQAScore, the benchmark's own metric, and of four published scorers (Supp.~E).

\section{Training a Better Generative Model}
\label{sec:training}

\begin{figure*}[t]
  \centering
  \figfile{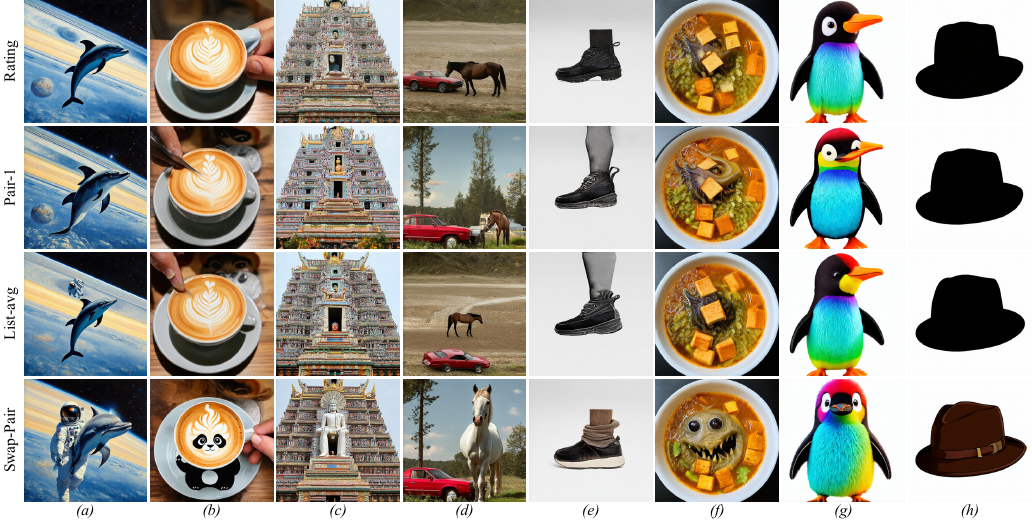}{\textwidth}{8.9cm}{fig5\_samples\_v4.pdf: eight selected fresh prompts; Rating-, Pair-1-, List-avg- and Swap-Pair-trained samples under shared noise}
  \vspace{-0.2em}
  \caption{\textbf{Samples of the \Rating-, \PairOne-, \ListAvg- and \PWC-trained models.}
  Same prompt, noise and seed per column; selected among the 727 of 4{,}960 fresh images of seeds 1 to 5 on which the \PWC image beats the other three on HPSv2.1 and PickScore (unselected samples in Supp.~E).
  \emph{(a)}~a dolphin in an astronaut suit on Saturn; \emph{(b)}~a panda making latte art; \emph{(c)}~a massive statue in a temple; \emph{(d)}~a tall horse next to a red car; \emph{(e)}~a shoe with a sock draped over it; \emph{(f)}~a monster soup with tofu; \emph{(g)}~a rainbow coloured penguin; \emph{(h)}~a hat.}
  \label{fig:samples}
\end{figure*}

\begin{table}[t]
  \centering
  \footnotesize
  \setlength{\tabcolsep}{3.2pt}
  \caption{\textbf{People prefer the \PWC-trained model.}
  Share (\%) of 1{,}500 judgments, five raters on 300 fresh prompts, that prefer its image; ties count one half; 95\% interval over prompts and raters.}
  \label{tab:human}
  \begin{tabular}{l c c}
    \toprule
    \PWC-trained against & Overall & Prompt match \\
    \midrule
    base model & 62.0 [58.2, 65.8] & 58.3 \\
    \List-trained & 58.6 [55.0, 62.2] & 56.4 \\
    \ListAvg-trained & 55.9 [52.4, 59.4] & 54.5 \\
    \PairOne-trained & 54.6 [50.8, 58.3] & 53.5 \\
    \PRP-trained & 53.1 [49.4, 56.9] & 52.6 \\
    Pref-GRPO-trained & 56.8 [53.2, 60.4] & 55.3 \\
    PickScore-trained & 56.5 [53.0, 60.0] & 58.8 \\
    HPSv2.1-trained & 58.1 [54.6, 61.6] & 59.7 \\
    \List-trained, both 600 steps & 61.7 [58.0, 65.4] & 58.0 \\
    \bottomrule
  \end{tabular}
\end{table}

\begin{table}[t]
  \centering
  \footnotesize
  \setlength{\tabcolsep}{2.6pt}
  \caption{\textbf{The lead holds at the public batch and KL weight and on FLUX.1-dev.}
  Top: SD3.5-Medium, 1{,}152 images per step and KL weight 0.04, 1{,}000 steps, 3 seeds; means on the pinned prompts (HPSv2.1 $\times$100; win: HPSv2.1 win rate). Bottom: FLUX.1-dev, 200 steps, 6 seeds; win rates (\%). Grey and bold as in \cref{tab:training}.}
  \label{tab:scale}
  \begin{tabular}{l c c c c}
    \toprule
    \multicolumn{5}{l}{\emph{SD3.5-Medium, Flow-GRPO's public batch and KL weight}} \\
    Reward & HPSv2.1 & PickScore & GenEval & Win \\
    \midrule
    base model & 29.79 & 22.51 & 67.0 & -- \\
    PickScore reward & \cellcolor{black!8}30.96 & \cellcolor{black!8}23.38 & 68.8 & \cellcolor{black!8}71.2 \\
    Pref-GRPO, as released & 31.02 & 22.95 & 70.4 & 73.9 \\
    \ListAvg, q8 & 30.89 & 22.86 & 70.0 & 72.4 \\
    \rowcolor{swapfill}
    \PWC, q8 & \textbf{31.47} & \textbf{23.05} & \textbf{72.3} & \textbf{79.1} \\
    \midrule
    \multicolumn{5}{l}{\emph{FLUX.1-dev, same recipe as \cref{tab:training}}} \\
    Reward & HPSv2.1 & ImageReward & \multicolumn{2}{c}{VQAScore} \\
    \midrule
    \List, q8 & 53.4 & 57.1 & \multicolumn{2}{c}{49.3} \\
    \PairOne, q8 & 57.6 & 58.4 & \multicolumn{2}{c}{50.2} \\
    Pref-GRPO, as released & 55.9 & 58.0 & \multicolumn{2}{c}{50.0} \\
    \rowcolor{swapfill}
    \PWC, q8 & \textbf{62.7} & \textbf{60.8} & \multicolumn{2}{c}{\textbf{52.4}} \\
    \bottomrule
  \end{tabular}
\end{table}

\paragraph{Setup.}
We fine-tune SD3.5-Medium~\cite{esser2024sd3,stability2024sd35} with LoRA at 512\,px under Flow-GRPO~\cite{liu2025flowgrpo}, with 8 prompts $\times$ 8 rollouts per step and KL weight 0.4, and change only the reward.
Seven rewards are read from q8: \Rating; \List and \ListAvg, which rank each group by one listwise message or by its eight rotations; and four pairwise rewards, the one-order win rate of Pref-GRPO~\cite{wang2025prefgrpo} (\PairOne), Tournament-GRPO's shuffled knockouts~\cite{yang2026tournamentgrpo}, \PRP's two-order votes and \PWC's Copeland score.
Two more are read from the UnifiedReward judge of Pref-GRPO, its released reward and \PWC, and three are the scorer rewards of published pipelines: PickScore as in Flow-GRPO, HPSv2.1 as in DanceGRPO~\cite{xue2025dancegrpo}, and the UnifiedReward score~\cite{wang2025unifiedreward}.
The q8 arms run 16 seeds to step 200, and 8 further seeds of \List, \ListAvg, \PairOne, \PRP and \PWC run to step 600; every other arm of \cref{tab:training} runs 8 seeds.
We report the win rate against the base model, the share of prompts on which a metric prefers the fine-tuned image to the base model's image from the same noise, on 1{,}200 pinned prompts (the HPDv2 test prompts and withheld HPDv2 training prompts) and on fresh prompts.
HPSv2.1 is the yardstick for the judge rewards; the scorer rewards optimise HPSv2.1 or a close relative, so against them we compare on ImageReward~\cite{xu2023imagereward}, GenEval~\cite{ghosh2023geneval} and VQAScore on GenAI-Bench~\cite{li2024genaibench}, which no reward optimises (recipe and absolute scores in Supp.~E).

\begin{figure}[t]
  \centering
  \figfile{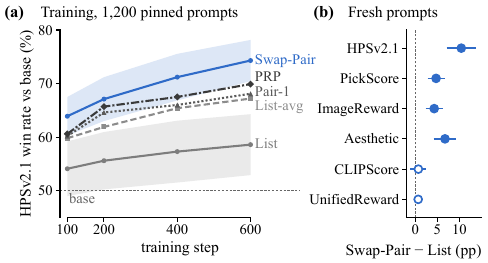}{\linewidth}{4.7cm}{fig3\_v11.pdf: (a) HPSv2.1 win rate vs base at steps 100 to 600, five arms; (b) Swap-Pair minus List on six metrics, 992 fresh prompts}
  \vspace{-1.5em}
  \caption{\textbf{The \PWC lead after longer training.}
  \emph{(a)}~Pinned HPSv2.1 win rate of the five q8 arms trained to step 600, mean over seeds 17 to 24; band: 95\% interval (\PWC, \List).
  \emph{(b)}~\PWC $-$ \List on 992 fresh prompts (hollow: interval includes 0).}
  \label{fig:training}
  \vspace{-0.6em}
\end{figure}

\paragraph{Readouts of one judge.}
Among the seven readouts of q8, \PWC has the highest value on every yardstick (\cref{tab:training}).
Its policies win 66.4\% of the pinned prompts, 11.4 points more than \List-trained and 10.2 more than \Rating-trained ones, and its leads over these two, Tournament-GRPO, \ListAvg and \PairOne hold after a Holm--Bonferroni correction~\cite{holm1979simple} (note of \cref{tab:training}).
The rewards inherit the failures of their readouts (\cref{sec:consequences}), and in training these failures become advantages that do not follow the image.
In \PairOne, for example, the first-shown image wins 62.5\% of the pair messages, and since the lower index is shown first, its mean reward falls from 0.625 for the first rollout of a group to 0.376 for the last, a reward that follows the sampling order rather than the image; \PWC's mean reward is 0.50 at every index.
Read in order, the pairwise rows of \cref{tab:training} separate the two changes that lead from \PairOne to \PWC: asking the second order with hard votes, as \PRP does, raises the pinned win rate from 62.6\% to 63.2\%, and summing the two orders' log-odds instead of counting votes raises it to 66.4\%.
The leads are larger after three times longer training (\cref{fig:training}a), and the longer runs show no sign of reward hacking: mean saturation stays within 1.4\% of the base model and diversity at 0.94 of it (Supp.~E).

\paragraph{Cost.}
A \PWC step takes 81.2\,s on one GPU against 43.6\,s with the \List reward, because the judge reads 56 pair messages instead of one; generation and the policy update take the same 41.8\,s in both.
The lead remains when the other rewards get the extra compute: \List trained for 600 steps, at 7.3 GPU-hours, still trails \PWC at step 200, at 4.5, and so do \ListAvg and \PairOne at the GPU-hours of \PWC's 200 steps (Supp.~E).

\paragraph{Published rewards and a published configuration.}
On Pref-GRPO's own judge, \PWC leads the reward released with Pref-GRPO on five of the six yardsticks (\cref{tab:training}).
The PickScore and HPSv2.1 rewards win the metrics they optimise and their close relatives (grey in \cref{tab:training}), but on ImageReward, GenEval and VQAScore, which no arm optimises, \PWC leads all three scorer rewards.
At the batch size and KL weight of Flow-GRPO's public configuration, 1{,}152 images per step and 0.04, and over 1{,}000 steps, \PWC again has the highest HPSv2.1, GenEval and win rate of the four arms (\cref{tab:scale}; recipe in Supp.~E).

\paragraph{People and a second generator.}
Raters prefer the images of the \PWC-trained model to those of the base model and of the models trained with \List, \ListAvg, \PairOne, Pref-GRPO and the two scorer rewards (\cref{tab:human}).
Against the \List-trained model they choose it in 58.6\% of 1{,}500 judgments, and in 61.7\% when both models train for 600 steps.
On FLUX.1-dev~\cite{flux2024}, \PWC again has the highest value on all three yardsticks, 9.3 points of HPSv2.1 win rate ahead of \List (\cref{tab:scale}).

\paragraph{Unseen prompts and benchmarks.}
The leads carry over to 992 fresh prompts held out from development, drawn from PartiPrompts~\cite{yu2022parti} and DrawBench~\cite{saharia2022imagen}.
Against \List-trained policies, \PWC leads on HPSv2.1, PickScore, ImageReward and the LAION aesthetic score after a Holm--Bonferroni correction over six metrics (\cref{fig:training}b), and on these four it leads every other readout of q8 (Supp.~E).
Prompt adherence improves as well, with the highest GenEval score and VQAScore win rate of all rewards (\cref{tab:training}), and diversity stays at the level of the \List-trained model (Supp.~E).
The samples of \cref{fig:samples} keep most of the layout and differ in prompt content: only the \PWC-trained model draws the panda in the latte art (b), puts a statue in the temple (c) and drapes the sock over the shoe (e), and its penguin is a rainbow from head to feet (g).\label{sec:samples}

\section{Conclusion}
\label{sec:conclusion}

The same multimodal judge names a different winner depending on which candidates it sees together.
On five open-weight judges, fifty images that the judge rates below both contenders change the listwise winner in 22 to 31\,\pp more sets than its noise floor, and on both Qwen3-VL judges this survives order averaging, label debiasing and isolating the encodings.
Asked one pair at a time in both orders, the same judge keeps its winner under covered additions, its \PWC pick beats every other readout on HPSv2.1, and its \PWC reward leads every other reward of the same judge.
As a GRPO reward, \PWC lifts the HPSv2.1 win rate 11.4\,\pp above one listwise message and 10.2\,\pp above the pointwise reward, leads the reward released with Pref-GRPO and trains a model whose images people prefer, and the lead holds at a published batch size and on FLUX.1-dev.
How the judge is asked is part of the reward, for selection and for training.
In practice this means three changes to a pipeline that already uses a multimodal judge: show it two images per message, ask both orders and sum the log-odds, and give each rollout its Copeland score; the cover step matters only when one image must be picked.

\paragraph{Limitations and future work.}
\label{sec:limits}
Searching from sampled answers instead of first-token label distributions would extend the study to API-served and reasoning judges, and computing the \PWC reward from a subset of the pairs would lower its cost.
Larger groups, larger judges and video are natural next targets.

\clearpage
\section*{LLM Usage Statement}
We used a large language model, Claude (Anthropic), only at the assist level, \ie, to edit or refine content produced by the authors.
This applied to writing (language, restructuring, and prose), related work, code and implementation, the design, execution, and analysis of experiments, figures and visual material, data, and the presentation of research ideas, hypotheses, claims, and conclusions.
We also provided the full manuscript to Claude to obtain reviewer-style feedback.
All research ideas, methods, code, data, experiments, and results originate from the authors, who checked all LLM-assisted content and take full responsibility for the submitted paper.

{
    \small
    \bibliographystyle{ieeenat_fullname}
    \bibliography{main}
}

\clearpage
\csletcs{r@M-tab:protocols}{r@tab:protocols}
\csletcs{r@M-tab:protocols@cref}{r@tab:protocols@cref}
\csletcs{r@M-sec:pair-results}{r@sec:pair-results}
\csletcs{r@M-sec:pair-results@cref}{r@sec:pair-results@cref}
\section*{Supplementary Material}
\appendix
\ifcsname theHsection\endcsname\expandafter\renewcommand\csname theHsection\endcsname{supp.\arabic{section}}\fi
\ifcsname theHtable\endcsname\expandafter\renewcommand\csname theHtable\endcsname{supp.\arabic{table}}\fi
\ifcsname theHfigure\endcsname\expandafter\renewcommand\csname theHfigure\endcsname{supp.\arabic{figure}}\fi
\ifcsname theHequation\endcsname\expandafter\renewcommand\csname theHequation\endcsname{supp.\arabic{equation}}\fi
\ifcsname theHlemma\endcsname\expandafter\renewcommand\csname theHlemma\endcsname{supp.\arabic{lemma}}\fi
\ifcsname theHproposition\endcsname\expandafter\renewcommand\csname theHproposition\endcsname{supp.\arabic{proposition}}\fi
\ifcsname theHremark\endcsname\expandafter\renewcommand\csname theHremark\endcsname{supp.\arabic{remark}}\fi
\ifcsname theHtheorem\endcsname\expandafter\renewcommand\csname theHtheorem\endcsname{supp.\arabic{theorem}}\fi
\ifcsname c@section\endcsname\setcounter{section}{0}\fi
\ifcsname c@table\endcsname\setcounter{table}{0}\fi
\ifcsname c@figure\endcsname\setcounter{figure}{0}\fi
\ifcsname c@equation\endcsname\setcounter{equation}{0}\fi
\ifcsname c@lemma\endcsname\setcounter{lemma}{0}\fi
\ifcsname c@proposition\endcsname\setcounter{proposition}{0}\fi
\ifcsname c@remark\endcsname\setcounter{remark}{0}\fi
\ifcsname c@theorem\endcsname\setcounter{theorem}{0}\fi
\renewcommand{\thesection}{\Alph{section}}
\renewcommand{\thetable}{S\arabic{table}}
\renewcommand{\thefigure}{S\arabic{figure}}
\renewcommand{\theequation}{S\arabic{equation}}
\renewcommand{\thelemma}{S\arabic{lemma}}
\renewcommand{\theproposition}{S\arabic{proposition}}
\renewcommand{\theremark}{S\arabic{remark}}
\renewcommand{\thetheorem}{S\arabic{theorem}}

\noindent This supplement follows the order of the main paper, except that the proofs (\cref{sec:s-proofs}) precede the extended measurements (\cref{sec:s-attack}).
\Cref{sec:s-notation} gives the notation, the readouts, every judge prompt verbatim and the rubrics (\mainref{sec:measure-setup}).
\Cref{sec:s-search} describes the decoy construction, the search and the matched floors (\mainref{sec:measure-add}).
\Cref{sec:s-proofs} proves the statements of \mainref{sec:pairwise}, classifies aggregation rules by the guarantee they keep, and gives the cost of \PWC with the certified-leader replay (\cref{sec:s-leader}).
\Cref{sec:s-attack} extends the measurement of \mainref{sec:measure}: every endpoint per judge, one added image, budget curves, transfer, text decoys, natural and content-free images, the closed-model check, the defences that failed, the label, isolation and shared-position controls, and the variants of the fix.
\Cref{sec:s-training} gives the training recipe and the extended results of \mainref{sec:consequences} and \mainref{sec:training}: reward audits with the exchange test, the seven q8 arms with \PRP and Tournament-GRPO and their seed statistics, every fresh-prompt metric, GenEval by task, the \Rating arm, the results on GenAI-Rank and GenAI-Bench, and the published rewards, longer training and equal compute, the runs at the public batch and KL weight, the five-rater human study and the FLUX.1-dev runs.
\Cref{sec:s-protocol} states which analyses were fixed before which data were seen.
Main-paper items are cited with their main-paper numbers; tables, figures and statements of this supplement are numbered S1, S2, \dots

\section{Notation, readouts, prompts and rubrics}
\label{sec:s-notation}

\subsection{Judges, data and estimators}
\label{sec:s-estimators}

\paragraph{Judges.}
We abbreviate the five open-weight judges as q8 (Qwen3-VL-8B~\cite{bai2025qwen3vl}), q32 (Qwen3-VL-32B), ivl38 (InternVL3.5-38B~\cite{zhu2025internvl3}), mst24 (Mistral-Small-3.1-24B~\cite{mistral2025small}) and ur7 (UnifiedReward-qwen-7b~\cite{wang2025unifiedreward}, run with the same family-agnostic prompts as the others rather than its own).
q8 is the judge of the main training study; the Pref-GRPO arms and one \PWC arm read the judge that Pref-GRPO releases (\cref{sec:s-further}).
All five are read through first-token log-probabilities (\cref{sec:s-readouts}); the one closed model we checked returns text only and is treated separately (\cref{sec:s-closed}).

\paragraph{Sets and clusters.}
A \emph{set} is one HPDv2 test prompt~\cite{wu2023hpsv2} with its three human-best images $A$, $B$, $C$ (human ranks 1--3, the \emph{contenders}) and one further slot for an added candidate.
The 300 development sets and the 100 held-out sets are disjoint; the held-out sets were judged once, after every analysis applied to them had been fixed (\cref{sec:s-protocol}).
A set is the cluster and the resampling unit of every attack interval; a prompt is the cluster for agreement with human rankings; the training study resamples seeds, and prompts within seeds (\cref{sec:s-train-eval}).

\paragraph{Budget-matched rates.}
For a set whose effective pool has $f$ flipping trials out of $n$, the probability that at least one of $k$ trials drawn without replacement flips is
\begin{equation}
  \mathrm{any}_k = 1 - \binom{n-f}{k}\Big/\binom{n}{k}, \label{eq:anyk}
\end{equation}
which equals the mean flip at $k=1$ and $[f>0]$ at $k=n$.
A best-of-$k$ search succeeds exactly when one of its $k$ decoys flips, so the searched rate at budget $k$ is the pool's $\mathrm{any}_k$; the main paper's ``fifty tries'' is the $k=n$ point; its ``one added image'' is the genuine addition against the decoy-free triple (\cref{tab:s-k1}), not the $k=1$ point of this curve.
An any-of-$k$ rate is compared only with an any-of-$k$ rate at the same $k$ and on the same sets.

\paragraph{Intervals.}
Every attack interval is a percentile set bootstrap with 5{,}000 resamples, paired across readouts that share sets.
Intervals are 95\% unless marked.
A few development rows were computed for an early decision on two judges at once at 97.5\%; we recomputed the per-judge attack rows at 95\% from the stored rows (\cref{tab:s-budget,tab:s-k1}), which changes no point estimate, and the rows still at 97.5\% (the isolation rows of \cref{tab:s-iso} and a few differences) are marked where they appear.
Agreement with human rankings uses a prompt-cluster bootstrap.
Training contrasts are two-sided; the interval is the wider of a two-level bootstrap (seeds, then prompts) and a Welch $t$ interval over seeds, and the $p$-value is the larger of the two (\cref{sec:s-train-eval}).

\begin{table}[t]
  \centering
  \scriptsize
  \setlength{\tabcolsep}{2.5pt}
  \caption{\textbf{Notation.} Per-set quantities; every reported rate is the mean over sets with a set-bootstrap CI.}
  \label{tab:s-notation}
  \begin{tabular}{@{}>{\raggedright\arraybackslash}p{0.21\linewidth}p{0.73\linewidth}@{}}
    \toprule
    Symbol & Meaning \\
    \midrule
    $A,B,C$ & the contenders: human ranks 1, 2, 3 of the set's prompt \\
    $d$ & the candidate added in the set's fourth slot: a searched decoy, a re-render, a grey image or a natural image \\
    decoy & a pool image that passes the screen and loses its pair against $A$ and against $B$ in both orders (\cref{sec:s-search}) \\
    $n$ & effective pool size after the dominance closure and exclusions; search and floor are cut to the same $n$ \\
    $W_0$ & modal winner over the effective pool \\
    $W_0^{\mathrm{rr}}$ & modal winner over the re-renders of the set's fixed decoy \\
    $s$ & searched flip: some decoy $d$ of the pool gives a winner $\ne W_0$ (any of $n$) \\
    $f$ & re-render floor: some re-render gives a winner $\ne W_0^{\mathrm{rr}}$ (any of $n$) \\
    $\Eone$ & searched excess, $\mathrm{mean}(s-f)$; the main paper's ``excess'' \\
    residual & some decoy gives a winner that is neither $W_0$ nor $d$: a flip to another contender \\
    decoy wins & some decoy is itself the winner (any of $n$; no floor) \\
    contender-only & $s$ and $f$ counting only flips to another contender \\
    one added image & $k=1$: mean over a set's added candidates $x$ of $[w(S\cup\{x\})\notin\{w(S),x\}]$, against the decoy-free triple $S=\{A,B,C\}$ (a genuine addition; a win of $x$ itself does not count) \\
    $R$ & $\Eone$ of a readout over $\Eone$ of \List on the same sets and resamples \\
    knife-edge & a set with $f=1$ \\
    $x\beats y$ & $x$ beats $y$: positive two-order margin $m(x,y)$ (\cref{sec:s-proofs}) \\
    covered & $y\beats x$ and $y$ beats everything $x$ beats \\
    universal loser & a candidate that loses every pair; a special case of covered \\
    K3+ & agreement with the HPDv2 human order on 1{,}200 three-image sets (winner; pairs), no decoy \\
    reorder & share of groups whose top pick changes when the same candidates are shown in another order \\
    exchange & excess share of groups whose pick among seven kept rollouts depends on which dominated rollout fills the eighth slot (\cref{sec:s-exchange}) \\
    $\rho$ & within-group Spearman correlation of a reward with HPSv2.1~\cite{wu2023hpsv2} \\
    best-of-8 & mean HPSv2.1 gain of the selected rollout over a random one, in \% of the oracle's gain \\
    win rate & share of prompts on which a trained policy's image outscores the base model's image from the same noise (ties count one half) \\
    \bottomrule
  \end{tabular}
\end{table}

\subsection{Readouts}
\label{sec:s-readouts}

\paragraph{First-token readouts.}
Every open-judge answer is the distribution of the first generated token over the readout's label set, read after the prompt's forced prefix (\cref{sec:s-prompts}), with greedy decoding in bf16.
A readout is \emph{unreadable} when its committed label mass is below 0.5; it is excluded and counted, never read as an answer.
The listwise index readouts have no assistant prefix, and a judge can abstain (``None of the images\dots''); every such readout is therefore also read with the message continued by the answer stem ``Image'', and with $\pi$ the first-token and $q$ the post-stem label probabilities and $r=1-\sum_k\pi_k$, $P(k)=\pi_k+r\,q_k$ and the committed mass is $\sum_k\pi_k+r\sum_k q_k$.
A sensitivity analysis that instead counts a readout as unreadable when its first-token mass is below 0.5 leaves the searched-excess decision unchanged on q8, q32, mst24 and ur7; ivl38 abstains often and is reported in \cref{sec:s-ivl}.

\paragraph{The readouts of the main paper.}
All use the same judge and the same rubric, ``best match to the prompt with the highest overall quality''.
\begin{itemize}[leftmargin=*,itemsep=1pt,topsep=2pt]
\item \List: all $N$ candidates in one message; $P(\text{winner}=i)$ over the labels $1,\dots,N$, renormalised; the winner is the arg\,max. As a training reward, the normalised rank of $P(\text{winner})$ within the group.
\item \ListAvg: the \List message in all $N$ cyclic rotations of the slots ($N=4$ in the attack, $N=8$ in training); per rotation the label distribution is mapped back to the candidates, and the reward or score is the mean over rotations; rotation 0 is the \List message byte for byte. As a training reward, the same rank transform as \List.
\item \Rating: each candidate alone, the expected 1--10 rating (the label ``10'' is read through a cached second token). As a training reward, the within-group rank.
\item \PairOne: a one-order pairwise reward that follows Pref-GRPO's released implementation~\cite{wang2025prefgrpo}: each unordered pair is asked once in a two-image message with the lower index shown first; the first-shown image wins if its margin $\log P(\text{``1''})-\log P(\text{``2''})$ is positive, a tie or an unreadable answer gives one half to each, and the score is the win count over the $N-1$ opponents. Pref-GRPO reads the winner from UnifiedReward-Think's text reply; we read our judge's margin.
\item \PRP~\cite{qin2024prp}: every pair in both orders; a hard win when both orders prefer the same image, otherwise (conflicting, exactly zero or unreadable) one half each; the score is the sum, and ties of the score fall back to the initial order (the lowest slot). As a training reward, the sum over the $N-1$ opponents divided by $N-1$.
\item \PWC: every pair in both orders; the two-order margin $m(x,y)$ is the sum of the two log-odds for $x$, so $m(y,x)=-m(x,y)$; the selection is the cover rule with Copeland scores over the survivors and the tie convention of \cref{sec:s-proofs}; the training reward is the Copeland score over all rollouts, one point per win and one half per tie, divided by $N-1$.
\end{itemize}
The pairwise messages hold exactly two images and the pair question of \cref{sec:s-pair-prompts}.
The published selection baselines of \mainref{tab:protocols}, permutation self-consistency, the knockout and Tournament-GRPO's repeated tournament, and the Tournament-GRPO training reward are read from the same \List and pair messages (\cref{sec:s-audit,sec:s-recipe}).

\paragraph{Diagnostic readouts.}
These appear only in \cref{sec:s-iso,sec:s-variants}.
\IsoList keeps the listwise message but masks attention so that every image block is encoded in isolation from the others; \IsoPair reads, from one isolated prefill, the label distribution of each pair tail (``Consider only Image $i$ and Image $j$\dots'') restricted to the labels $i,j$, summed over both mention orders; the forward-tail variant reads only the order with the lower slot named first.
The yes/no readout asks of each image in turn ``Does Image $k$ match the prompt well and have high overall quality?'' and reads the yes--no log-odds, without comparing labels.
The per-candidate readout asks the same question of each image alone.

\subsection{Screen, dominance check and rubric scope}
\label{sec:s-rubrics}

\paragraph{Screen (D-point).}
For a multimodal judge, the screen score of an image under the set's prompt is the mean of its expected MJ-Bench~\cite{chen2024mjbench} alignment and artifacts ratings.
A pool image qualifies when its score is at least 1.0 below both $A$'s and $B$'s; for UnifiedReward the score is its own point score and margin.
A candidate with an unreadable screen readout never qualifies, and a set whose $A$ or $B$ readout is unreadable is excluded.

\paragraph{Dominance check (D-pair).}
For each contender $X\in\{A,B\}$ and each presentation order, the summed log-odds margin of $X$ over the candidate under the two MJ-Bench two-image prompts (alignment and artifacts) must be positive: eight queries per checked candidate (UnifiedReward: its pair-rank prompt, four queries).
A check is complete only with all its parts; an unreadable part fails the candidate.

\paragraph{Rubric scope.}
The screen and dominance rubrics (alignment and artifacts) differ from the decision rubric (overall quality), as they do in practice, where a pointwise scorer and a listwise judge often use different prompts.
The screen therefore certifies that a decoy is scalar-inferior to $A$ and $B$ and that the judge prefers $A$ and $B$ to it in both orders on its own rubric, not that the decision readout ranks it last.
For the pairwise readouts the relevant relation is the readout's own pair winners: a decoy that is not covered under them is counted apart (\cref{sec:s-proofs}), and every residual flip of \PWC goes through such a decoy.
Under the pairwise readouts' own pair winners, the stored decoy is a universal loser in 99.6\% of the attack messages (\cref{sec:s-pairwise-attack}).

\paragraph{Agreement with human rankings (K3+).}
On 1{,}200 three-image sets drawn from the 300 development prompts, and on 1{,}200 sets from the 100 held-out prompts, every readout judges every set once with no added candidate.
Winner agreement is the share of sets whose selected image is the human-best of the three; pair agreement is the share of within-set pairs ordered as the humans order them, a tie counting one half.
Each readout uses its full decision rule (arg\,max for \List and \Rating; the cover rule for \PWC).
Intervals are 95\% prompt-cluster bootstraps, paired across readouts.

\subsection{Prompts, verbatim}
\label{sec:s-prompts}
Every string below is shown exactly as the judge receives it; placeholders in braces are filled by plain replacement (\texttt{\{prompt\}} is the set's HPDv2 prompt, \texttt{\{n\}} the number of candidates, \texttt{\{k\}}, \texttt{\{i\}}, \texttt{\{j\}} slot numbers).
Images are inserted where the bracketed markers stand.
Literature strings were byte-compared with their public sources before any row was written; strings marked ``ours'' have no literature counterpart.
The blocks of this subsection are generated from the frozen prompt module.

{\let\subsection\subsubsection
% Generated by make_prompts.py from the frozen prompt module; do not edit by hand.
% prompts SHA 9032192f4a89
\subsection{Listwise decision prompt and the isolated readouts (ours)}

\begin{promptbox}{Listwise, MLLM judges (decision cell; also the closed judges)}{ours}
\pline{Here are \{n\} images generated for the prompt \textquotedbl{}\{prompt\}\textquotedbl{}.}
\pline{Image 1: \emph{\textrm{[image 1]}}}
\pline{Image 2: \emph{\textrm{[image 2]}}}
\pline{\emph{\textrm{(one label and image per slot, one per line)}}}
\pline{Image \{n\}: \emph{\textrm{[image \{n\}]}}}
\pline{Which image is the best match to the prompt with the highest overall quality? Answer with the image number only.}
\end{promptbox}

\begin{promptbox}{R-faith tail (secondary rubric), replacing the listwise tail}{ours}
\pline{Which image matches the prompt most faithfully (ignore aesthetics)? Answer with the image number only.}
\end{promptbox}

\begin{promptbox}{Packed 2x2 contact sheet, head (the tail is the listwise tail)}{ours}
\pline{This contact sheet shows \{n\} images generated for the prompt \textquotedbl{}\{prompt\}\textquotedbl{}, each with its number printed in the grey strip directly below it, as \textquotedbl{}\#k\textquotedbl{}.}
\pline{\emph{\textrm{[one sheet with four numbered tiles]}}}
\end{promptbox}

\begin{promptbox}{V2 tail and P (per candidate k); also the sibling-swap query}{ours}
\pline{Does Image \{k\} match the prompt well and have high overall quality? Answer yes or no.}
\end{promptbox}

\begin{promptbox}{V4 tail (per ordered pair i, j)}{ours}
\pline{Consider only Image \{i\} and Image \{j\}. Which of the two better matches the prompt and has higher overall quality? Answer with the image number only.}
\end{promptbox}

\begin{promptbox}{UnifiedReward listwise (adaptation of its pair-rank prompt to n images)}{ours}
\pline{\emph{\textrm{[n images]}}}
\pline{You are given a text caption and \{n\} generated images based on that caption. Your task is to evaluate and compare these images based on two key criteria:}
\pline{1. Alignment with the Caption: Assess how well each image aligns with the provided caption. Consider the accuracy of depicted objects, their relationships, and attributes as described in the caption.}
\pline{2. Overall Image Quality: Examine the visual quality of each image, including clarity, detail preservation, color accuracy, and overall aesthetic appeal.}
\pline{Compare all images using the above criteria and select the one that best aligns with the caption while exhibiting superior visual quality.}
\pline{Provide a clear conclusion such as \textquotedbl{}Image 1 is the best.\textquotedbl{}}
\pline{Your task is provided as follows:}
\pline{Text Caption: {[}\{prompt\}{]}}
\end{promptbox}

\subsection{Literature prompts (byte-verified against their sources)}

\begin{promptbox}{D-point, alignment rating (MJ-Bench, single image)}{\url{https://raw.githubusercontent.com/MJ-Bench/MJ-Bench/main/prompt_template/prompts_single_image/alignment_single_number_scale10.txt}}
\pline{\emph{\textrm{[image]}}}
\pline{As a professional \textquotedbl{}Text-{}to-{}Image\textquotedbl{} quality assessor, your task is to evaluate the text-{}image alignment quality of an image generated from a specific prompt. The criteria for evaluation are as follows:~}
\pline{-{} *{}*{}Alignment*{}*{}: Assess how accurately the image reflects the given prompt. Check if all elements, including the background, object attributes such as color, gender, skin texture, and count, along with actions and spatial relationships like positioning, are correctly represented.}
\pline{\mbox{}}
\pline{\mbox{}}
\pline{Please analyze step by step and provide the ALIGNMENT-{}RATING using the following scale: {[}1, 2, 3, 4, 5, 6, 7, 8, 9, 10{]}. ~In this scale, 1 represents the worst alignment quality, and 10 represents the best alignment quality.}
\pline{\mbox{}}
\pline{\mbox{}}
\pline{The output of the rating should be in the following two-{}string format without line breaks and indentation. Here is an example:}
\pline{\textasciigrave{}\textasciigrave{}\textasciigrave{}}
\pline{\{\textquotedbl{}ANALYSIS\textquotedbl{}: \textquotedbl{}YOUR ANALYSIS\textquotedbl{}\}~}
\pline{\mbox{}}
\pline{\{\textquotedbl{}ALIGNMENT-{}RATING\textquotedbl{}: \textquotedbl{}YOUR RATING\textquotedbl{}\}}
\pline{\textasciigrave{}\textasciigrave{}\textasciigrave{}}
\pline{\mbox{}}
\pline{Now, proceed with evaluating the image based on the prompt description provided. The prompt is:~}
\pline{\{prompt\}}
\end{promptbox}

\begin{promptbox}{D-point, artifacts rating (MJ-Bench, single image)}{\url{https://raw.githubusercontent.com/MJ-Bench/MJ-Bench/main/prompt_template/prompts_single_image/artifacts_single_number_scale10.txt}}
\pline{\emph{\textrm{[image]}}}
\pline{As a professional \textquotedbl{}Text-{}to-{}Image\textquotedbl{} quality assessor, your task is to evaluate the quality of an image generated from a specific prompt. The criteria for evaluation are as follows:~}
\pline{-{} *{}*{}Artifacts*{}*{}: Identify if any artifacts in the image, such as distortion, blurriness, or illogical representation of facial features, limbs, fingers, objects, or text. These artifacts should be avoided in the image.}
\pline{\mbox{}}
\pline{\mbox{}}
\pline{Please analyze step by step and provide the ARTIFACTS-{}RATING using the following scale: {[}1, 2, 3, 4, 5, 6, 7, 8, 9, 10{]}. In this scale, 1 represents the worst quality, and 10 represents the best quality.}
\pline{\mbox{}}
\pline{Please do not generate any other opening, closing, and explanations. The output of the analysis and rating should be strictly adhere to the following format:}
\pline{\textasciigrave{}\textasciigrave{}\textasciigrave{}}
\pline{ANALYSIS: Provide your analysis here.}
\pline{\mbox{}}
\pline{ARTIFACTS-{}RATING: Only provide your rating here.}
\pline{\textasciigrave{}\textasciigrave{}\textasciigrave{}}
\pline{\mbox{}}
\pline{Now, proceed with evaluating the image based on the prompt:~}
\pline{\textasciigrave{}\textasciigrave{}\textasciigrave{}}
\pline{\{prompt\}}
\pline{\textasciigrave{}\textasciigrave{}\textasciigrave{}}
\end{promptbox}

\begin{promptbox}{D-pair, alignment (MJ-Bench, two images)}{\url{https://raw.githubusercontent.com/MJ-Bench/MJ-Bench/main/prompt_template/prompts_multi_image/alignment_multi_number_scale10.txt}}
\pline{\emph{\textrm{[two images]}}}
\pline{As a professional \textquotedbl{}Text-{}to-{}Image\textquotedbl{} quality inspector, your task is to assess the quality of two images generated from the same prompt. The criteria for evaluation are as follows:~}
\pline{-{} *{}*{}Alignment*{}*{}: Assess how accurately the image reflects the given prompt. Check if all elements, including the background, object attributes such as color, gender, skin texture, and count, along with actions and spatial relationships like positioning, are correctly represented.}
\pline{\mbox{}}
\pline{Please analyze each image step by step and provide the IMAGE-{}1 RATING and IMAGE-{}2 RATING using the following scale: {[}1, 2, 3, 4, 5, 6, 7, 8, 9, 10{]}. In this scale, 1 represents the worst, and 10 represents the best.}
\pline{Finally, based on your ratings, choose the overall BETTER IMAGE by stating your preference with a number in {[}0, 1, 2{]}. Here, 1 means IMAGE-{}1 is better than IMAGE-{}2, 2 means IMAGE-{}2 is better than IMAGE-{}1, and 0 means you cannot decide which one is better (or they are equal). However, please try to avoid giving a \textquotedbl{}tie\textquotedbl{} preference and be as decisive as possible.~}
\pline{\mbox{}}
\pline{\mbox{}}
\pline{Please do not generate any other opening, closing, and explanations. The output of the analysis and rating should be strictly adhere to the following format:}
\pline{\textasciigrave{}\textasciigrave{}\textasciigrave{}}
\pline{IMAGE-{}1 RATING: YOUR IMAGE-{}1 RATING}
\pline{\mbox{}}
\pline{IMAGE-{}2 RATING: YOUR IMAGE-{}2 RATING}
\pline{\mbox{}}
\pline{BETTER IMAGE: YOUR CHOICE USING A NUMBER}
\pline{\mbox{}}
\pline{ANALYSIS OF CHOICE: YOUR ANALYSIS}
\pline{\textasciigrave{}\textasciigrave{}\textasciigrave{}}
\pline{\mbox{}}
\pline{\mbox{}}
\pline{Now, proceed with evaluating these images based on the prompt:~}
\pline{\textasciigrave{}\textasciigrave{}\textasciigrave{}}
\pline{\{prompt\}}
\pline{\textasciigrave{}\textasciigrave{}\textasciigrave{}}
\end{promptbox}

\begin{promptbox}{D-pair, artifacts (MJ-Bench, two images)}{\url{https://raw.githubusercontent.com/MJ-Bench/MJ-Bench/main/prompt_template/prompts_multi_image/artifacts_multi_number_scale10.txt}}
\pline{\emph{\textrm{[two images]}}}
\pline{As a professional \textquotedbl{}Text-{}to-{}Image\textquotedbl{} quality inspector, your task is to evaluate the quality of two images generated from a same prompt. The criteria for evaluation are as follows:~}
\pline{-{} *{}*{}Artifacts*{}*{}: Identify if any artifacts in the image, such as distortion, blurriness, or illogical representation of facial features, limbs, fingers, objects, or text. These artifacts should be avoided in the picture.}
\pline{\mbox{}}
\pline{Please analyze each image step by step and provide the IMAGE-{}1 RATING and IMAGE-{}2 RATING using the following scale: {[}1, 2, 3, 4, 5, 6, 7, 8, 9, 10{]}. In this scale, 1 represents the worst, and 10 represents the best.}
\pline{Finally, based on your ratings, choose the overall BETTER IMAGE by stating your preference with a number in {[}0, 1, 2{]}. Here, 1 means IMAGE-{}1 is better than IMAGE-{}2, 2 means IMAGE-{}2 is better than IMAGE-{}1, and 0 means you cannot decide which one is better (or they are equal). However, please try to avoid giving a \textquotedbl{}tie\textquotedbl{} preference and be as decisive as possible.}
\pline{\mbox{}}
\pline{\mbox{}}
\pline{Please do not generate any other opening, closing, and explanations. The output of the analysis and rating should be strictly adhere to the following format:}
\pline{\textasciigrave{}\textasciigrave{}\textasciigrave{}}
\pline{IMAGE-{}1 RATING: YOUR IMAGE-{}1 RATING}
\pline{\mbox{}}
\pline{IMAGE-{}2 RATING: YOUR IMAGE-{}2 RATING}
\pline{\mbox{}}
\pline{BETTER IMAGE: YOUR CHOICE USING A NUMBER}
\pline{\mbox{}}
\pline{ANALYSIS OF CHOICE: YOUR ANALYSIS}
\pline{\textasciigrave{}\textasciigrave{}\textasciigrave{}}
\pline{\mbox{}}
\pline{Now, proceed with evaluating these images based on the prompt:~}
\pline{\textasciigrave{}\textasciigrave{}\textasciigrave{}}
\pline{\{prompt\}}
\pline{\textasciigrave{}\textasciigrave{}\textasciigrave{}}
\end{promptbox}

\begin{promptbox}{UnifiedReward point score (D-point)}{\url{https://raw.githubusercontent.com/CodeGoat24/UnifiedReward/main/inference_qwen/image_generation/qwen_point_score_image_generation.py}}
\pline{\emph{\textrm{[image]}}}
\pline{You are given a text caption and a generated image based on that caption. Your task is to evaluate this image based on two key criteria:}
\pline{1. Alignment with the Caption: Assess how well this image aligns with the provided caption. Consider the accuracy of depicted objects, their relationships, and attributes as described in the caption.}
\pline{2. Overall Image Quality: Examine the visual quality of this image, including clarity, detail preservation, color accuracy, and overall aesthetic appeal.}
\pline{Extract key elements from the provided text caption, evaluate their presence in the generated image using the format: \textquotesingle{}element (type): value\textquotesingle{} (where value=0 means not generated, and value=1 means generated), and assign a score from 1 to 5 after \textquotesingle{}Final Score:\textquotesingle{}.}
\pline{Your task is provided as follows:}
\pline{Text Caption: {[}\{prompt\}{]}}
\end{promptbox}

\begin{promptbox}{UnifiedReward pair rank (D-pair; the Pref-GRPO group reward)}{\url{https://raw.githubusercontent.com/CodeGoat24/UnifiedReward/main/inference_qwen/image_generation/qwen_pair_rank_image_generation.py}}
\pline{\emph{\textrm{[two images]}}}
\pline{You are given a text caption and two generated images based on that caption. Your task is to evaluate and compare these images based on two key criteria:}
\pline{1. Alignment with the Caption: Assess how well each image aligns with the provided caption. Consider the accuracy of depicted objects, their relationships, and attributes as described in the caption.}
\pline{2. Overall Image Quality: Examine the visual quality of each image, including clarity, detail preservation, color accuracy, and overall aesthetic appeal.}
\pline{Compare both images using the above criteria and select the one that better aligns with the caption while exhibiting superior visual quality.}
\pline{Provide a clear conclusion such as \textquotedbl{}Image 1 is better.\textquotedbl{}, \textquotedbl{}Image 2 is better.\textquotedbl{} and \textquotedbl{}Both images are equally good.\textquotedbl{}}
\pline{Your task is provided as follows:}
\pline{Text Caption: {[}\{prompt\}{]}}
\end{promptbox}

\begin{promptbox}{VL-RewardBench pairwise (text-decoy family)}{\url{https://raw.githubusercontent.com/vl-rewardbench/VL_RewardBench/main/prompt.py}}
\pline{~You are a highly capable multimodal AI assistant tasked with evaluating answers to visual questions. Please analyze the following image and question, then determine which of the two provided answers is better.}
\pline{\mbox{}}
\pline{Question: \{question\}}
\pline{\mbox{}}
\pline{Answer 1: \{answer1\}}
\pline{\mbox{}}
\pline{Answer 2: \{answer2\}}
\pline{\mbox{}}
\pline{Please evaluate both answers based on the following criteria:}
\pline{1. Accuracy: How well does the answer align with the visual information in the image?}
\pline{2. Completeness: Does the answer fully address all aspects of the question?}
\pline{3. Clarity: Is the answer easy to understand and well-{}articulated?}
\pline{4. Relevance: Does the answer directly relate to the question and the image?}
\pline{\mbox{}}
\pline{After your evaluation, please:}
\pline{1. Explain your reasoning for each criterion.}
\pline{2. Provide an overall judgment on which answer is better (Answer 1 or Answer 2). For example: Overall Judgment: Answer X is better.}
\pline{\mbox{}}
\pline{Your response should be structured and detailed, demonstrating your understanding of both the visual and textual elements of the task.}
\pline{\emph{\textrm{[image]}}}
\end{promptbox}

\begin{promptbox}{VL-RewardBench listwise (ours: its pairwise prompt for n answers)}{ours}
\pline{~You are a highly capable multimodal AI assistant tasked with evaluating answers to visual questions. Please analyze the following image and question, then determine which of the \{n\} provided answers is best.}
\pline{\mbox{}}
\pline{Question: \{question\}}
\pline{\mbox{}}
\pline{\{answers\}}
\pline{\mbox{}}
\pline{Please evaluate all answers based on the following criteria:}
\pline{1. Accuracy: How well does the answer align with the visual information in the image?}
\pline{2. Completeness: Does the answer fully address all aspects of the question?}
\pline{3. Clarity: Is the answer easy to understand and well-{}articulated?}
\pline{4. Relevance: Does the answer directly relate to the question and the image?}
\pline{\mbox{}}
\pline{After your evaluation, please:}
\pline{1. Explain your reasoning for each criterion.}
\pline{2. Provide an overall judgment on which answer is best. For example: Overall Judgment: Answer X is the best.}
\pline{\mbox{}}
\pline{Your response should be structured and detailed, demonstrating your understanding of both the visual and textual elements of the task.}
\pline{\emph{\textrm{[image]}}}
\end{promptbox}

\begin{promptbox}{VL-RewardBench pointwise (ours: its criteria, one answer)}{ours}
\pline{~You are a highly capable multimodal AI assistant tasked with evaluating answers to visual questions. Please analyze the following image and question, then rate the provided answer.}
\pline{\mbox{}}
\pline{Question: \{question\}}
\pline{\mbox{}}
\pline{Answer: \{answer\}}
\pline{\mbox{}}
\pline{Please evaluate the answer based on the following criteria:}
\pline{1. Accuracy: How well does the answer align with the visual information in the image?}
\pline{2. Completeness: Does the answer fully address all aspects of the question?}
\pline{3. Clarity: Is the answer easy to understand and well-{}articulated?}
\pline{4. Relevance: Does the answer directly relate to the question and the image?}
\pline{\mbox{}}
\pline{After your evaluation, provide an overall rating of the answer from 1 (worst) to 10 (best). For example: Overall Rating: X.}
\pline{\emph{\textrm{[image]}}}
\end{promptbox}

\subsection{Forced answer prefixes}

Where a literature prompt asks for analysis before its answer, the answer field is forced as an assistant-side prefix, so the first generated token is the answer (\cref{sec:s-rubrics}).

\begin{center}\small\begin{tabular}{@{}ll@{}}\toprule Prompt & Forced prefix \\ \midrule
Listwise (MLLM) & \texttt{\emph{\textrm{(none)}}} \\
MJ-Bench alignment, single & \texttt{\{\textquotedbl{}ALIGNMENT-{}RATING\textquotedbl{}: \textquotedbl{}} \\
MJ-Bench artifacts, single & \texttt{ARTIFACTS-{}RATING:~} \\
MJ-Bench, two images & \texttt{BETTER IMAGE:~} \\
UnifiedReward point & \texttt{Final Score:~} \\
UnifiedReward pair / list & \texttt{Image~} \\
VL-RewardBench pair / list & \texttt{Overall Judgment: Answer~} \\
VL-RewardBench point & \texttt{Overall Rating:~} \\
\bottomrule\end{tabular}\end{center}
}

\subsubsection{Rating and pair prompts (ours)}
\label{sec:s-pair-prompts}
\Rating and the pairwise readouts (\PairOne, \PRP, \PWC, the knockout and the repeated tournament) use two further templates.
The pair message is the listwise head with $n=2$ and the pair tail of the isolated readouts with $i=1$, $j=2$; its committed readout uses the answer stem ``Image''.

\begin{promptbox}{\Rating (one image alone)}{ours}
\pline{Here is an image generated for the prompt \textquotedbl{}\{prompt\}\textquotedbl{}.}
\pline{Image 1: \emph{\textrm{[image]}}}
\pline{Rate how well Image 1 matches the prompt and how high its overall quality is, on a scale from 1 to 10. Answer with the number only.}
\end{promptbox}

\begin{promptbox}{Pair message (one ordered pair $x$, $y$; \PairOne, \PRP, \PWC)}{ours}
\pline{Here are 2 images generated for the prompt \textquotedbl{}\{prompt\}\textquotedbl{}.}
\pline{Image 1: \emph{\textrm{[image $x$]}}}
\pline{Image 2: \emph{\textrm{[image $y$]}}}
\pline{Consider only Image 1 and Image 2. Which of the two better matches the prompt and has higher overall quality? Answer with the image number only.}
\end{promptbox}

\clearpage
\section{Decoy construction, search and floors}
\label{sec:s-search}

\subsection{The search, step by step}
\label{sec:s-steps}

\begin{enumerate}[leftmargin=*,itemsep=2pt,topsep=2pt]
\item \textbf{Sets.} From HPDv2's test prompts, a seeded shuffle assigns 300 development sets and 100 held-out sets. $A$, $B$ and $C$ are the human ranks 1--3 among the prompt's generated images. Within each split a seeded order assigns the added candidate's slot so that every slot holds it equally often (75 development sets per slot), and $A$, $B$, $C$ fill the other three slots in a permutation drawn once per set; their relative order is the same in every message of the set.
\item \textbf{Candidate pool.} One seeded list per set, shared by every judge: the same-prompt human losers (ranks below $C$) and the human-last images of other prompts. Each image is resized once to the judge's canonical input, and the same file serves every readout, so a candidate has the same token count within a judge.
\item \textbf{Screen.} D-point (\cref{sec:s-rubrics}) for the pool images and for $A$, $B$, $C$ under the set's prompt. Qualifiers keep the pool order.
\item \textbf{Fixed decoy.} The qualifiers are checked with D-pair in pool order until one passes; it is the set's fixed decoy and the source of its re-render floor. A set with no verified decoy is excluded from the endpoints and counted (\cref{tab:s-dpair}).
\item \textbf{Searched pool.} Up to fifty further qualifiers in pool order.
\item \textbf{Search.} One listwise message per searched decoy, with the decoy in the set's slot; one message per re-render of the fixed decoy, as many as the pool has decoys; one message of $\{A,B,C\}$ with no added candidate, in the same relative order.
\item \textbf{Dominance closure.} D-pair is run on a seeded random sample of the searched pool and on every decoy whose winner differs from $W_0$. Failures leave the pool, $W_0$ is recomputed, and new flips are checked, to a fixed point; every counted flip of the final pool therefore carries a complete check, or the judge's row is incomplete. Each readout that is attacked gets the same closure on its own rows.
\item \textbf{Endpoints.} Unreadable rows are dropped; the effective pool and the readable re-renders are cut to the same length $n$, and the estimators of \cref{tab:s-notation} are computed per set. When a budget $k$ exceeds a set's $n$, \cref{eq:anyk} is evaluated at $k=n$.
\item \textbf{Re-search per readout.} For every readout attacked in \cref{sec:s-attack}, the decoys and re-renders are re-judged under that readout and the closure is rerun, because a readout changes which decoys flip.
\end{enumerate}

\subsection{Floors and matched controls}
\label{sec:s-floors}

\paragraph{Re-render floor.}
Each re-render of the fixed decoy is a re-encoding with a small seeded translation and brightness change, regenerated bit for bit from (set, index).
The re-renders flip against their own modal winner, so a set whose winner sits on a knife edge is charged to the floor, and the floor is the best of the same number of tries as the search.
The floor is not pure noise: under \List a re-render reaches every later candidate and the readout, so knife-edge sets flip under it.
Subtracting it makes the searched excess conservative; the floor itself is reported in every row (\cref{tab:s-budget}).

\paragraph{Order floor and rotation averaging.}
\ListAvg judges every pool, re-render and decoy-free message in all four cyclic rotations of the slots and averages the label distributions mapped back to the candidates; rotation 0 reproduces the stored single-message rows bitwise (2{,}691 of 2{,}691 rows on q32, 2{,}601 of 2{,}601 on q8).
Averaging lowers the q32 re-render floor from 17.1 to 13.7 points, because it removes order noise, but not the searched flip rate, 39.5 against 38.5 (\cref{sec:s-listavg}).

\paragraph{One added image ($k=1$).}
At $k=1$ the swap floor is degenerate: a re-render of the fixed decoy is itself a dominated decoy and is exchangeable with a random one.
One added image is therefore measured as a genuine addition against the decoy-free triple (\cref{tab:s-notation}), for four kinds of added candidate: a searched decoy, a re-render of the fixed decoy, a content-free mid-grey image with independent pixel jitter, and a natural image (\cref{tab:s-k1}).

\paragraph{Natural images.}
Fifty unscreened natural images per set, the human-best generated images of other test prompts in a seeded order, in the added slot, one message each, on q8 and q32.
They need no access to the judge.
They are reported as any flip, as flips to another contender (a natural image may itself win), and one at a time against the grey images.

\paragraph{Content-free images.}
A diverse content-free floor replaces the decoys by distinct solid colours, gradients and noise fields, seeded by set and index, in the added slot, through the same search on q8 and q32; its any-of-$n$ rate is compared with the decoys' and the re-renders' at the same budget.

\paragraph{Targeted endpoint.}
From the stored search rows: per set, the target is the contender drawn by a seeded coin from the two that are not the decoy-free winner; a trial succeeds when the target wins.
Any of $k$ decoys is compared with any of $k$ re-renders at $k\in\{1,3,10,50\}$ by \cref{eq:anyk}.

\subsection{Secondary designs}
\label{sec:s-secondary-designs}

\paragraph{Adaptive decoys.}
On q8, every pool image that \Rating rates below both $A$ and $B$ by less than 1.0 and that passes D-pair; the walk, search and floor are as above over this pool.
Such decoys are dominated in both pair orders but survive \ScreenList's rating screen, so they target it (\cref{sec:s-defences}).
58 of the 300 sets have one.

\paragraph{Native answers.}
On q32, the listwise prompt without the forced first-token readout, generated greedily, the winner parsed from the reply (the reply's only integer in $1,\dots,N$, or the only one on its last line); the reference message, the decoy-free triple, the stored best decoy and best re-render of each set, and a sample of random decoys and grey images are re-judged once.

\paragraph{Far contenders and larger sets.}
On q32, the contenders are human ranks 1, 5 and 9 of each development prompt, and the screen, pool, search and floor are rerun for these triples.
On q8, $N=8$: the contenders are human ranks 1--7 and the decoy takes the eighth slot.

\paragraph{Transfer between open judges.}
For the target judge q32, up to three decoys per set that push q8's probability of $W_0$ furthest below its pool median, chosen on q8's outcomes only; each is D-pair-checked on the target, and only decoys in the target's own screened pool that pass are counted.
Any of $k$ transferred decoys is compared with any of $k$ of the target's random decoys and re-renders, paired.

\paragraph{Text decoys.}
On VL-RewardBench~\cite{li2024vlrewardbench}, q8 ranks $N=3$ text responses to an image; decoys are screened, checked and searched as for images, and the floor is fifty re-renders of the fixed decoy (\cref{sec:s-text}).

\paragraph{Closed-model check.}
Claude Opus~5.5 returns text, so it is not screened and receives q8's strongest dominated decoys, three per set on 120 sets, against three random pool decoys, three re-renders and a re-query of the same message, with $W_0$ the modal answer of repeated reference queries.
It was queried through blinded agents at its default sampled settings and parsed with the native-answer parser.
The comparison was fixed before any answer: transfer had to reach at least twice the re-query rate and at least 10 points above the random-decoy rate, with a positive lower bound (\cref{sec:s-closed}).

\subsection{Dominance-check statistics}
\label{sec:s-dpair}

\Cref{tab:s-dpair} gives the closure statistics of the development chain.
The random-check failure rate is the expected share of unchecked, non-flipping decoys in the random-decoy denominator that would fail D-pair; no counted flip carries an unchecked decoy.

\begin{table}[t]
  \centering
  \scriptsize
  \setlength{\tabcolsep}{3pt}
  \caption{\textbf{Dominance closure on the development sets.} Checked: decoys put through D-pair by the walk, the random sample and the closure. Random-check failure: share of the seeded random checks that failed (95\%). No verified decoy: sets excluded because no qualifier passed.}
  \label{tab:s-dpair}
  \begin{tabular}{@{}lrrrrr@{}}
    \toprule
    judge & sets & checked & failed & random-check failure & no verified decoy \\
    \midrule
    q32 & 299 & 6{,}058 & 72 & 1.1 [0.5, 1.8] of 2{,}990 & 1 / 300 \\
    q8 & 289 & 5{,}248 & 269 & 3.7 [2.7, 4.8] of 2{,}890 & 11 / 300 \\
    ivl38 & 298 & -- & -- & -- & 2 / 300 \\
    mst24 & 298 & -- & -- & -- & 2 / 300 \\
    ur7 & 239 & -- & -- & -- & 61 / 300 \\
    q8, text & 486 & 6{,}680 & 252 & -- & 0 \\
    \bottomrule
  \end{tabular}
\end{table}

Knife-edge sets number 51 of 299 on q32 and 26 of 289 on q8; excluding them, the searched flip rate is 28.2 [21.8, 34.7] on q32 and 34.2 [27.8, 40.7] on q8.
On q32 a decoy is the modal winner in 0 of 299 sets, and a single decoy wins 0.1 [0.0, 0.1]\% of messages.
ur7's 61 sets without a verified decoy (20.3\%) are disclosed beside its row; its searched excess is computed on the remaining 239 sets after a complete closure.

\clearpage
\section{Proofs, aggregation rules and the certified leader}
\label{sec:s-proofs}

\subsection{Definitions and hypothesis}
\label{sec:s-defs}

\paragraph{Pair margins.}
A candidate set $T$ is a finite set of images, each in a slot $0,\dots,|T|-1$.
For every unordered pair $\{x,y\}\subseteq T$, \PWC sends two messages that hold exactly the two images, $x$ first and $y$ first, and reads from each the first-token log-odds of the label of its first image over the label of its second.
The two-order margin of $x$ over $y$, $m(x,y)$, is the sum of the two log-odds for $x$ (the released code rounds it to six decimals), so $m(y,x)=-m(x,y)$ exactly.
A message whose committed label mass is below $\tfrac12$ is unreadable and leaves the pair's margin undefined.
Summing both orders cancels a position bias that is additive and the same in both messages; whether the judge's bias has that form is measured, not assumed, and nothing below depends on it.

\paragraph{Hypothesis H.}
The margin of a pair is a function of its two images, their order and the judge only; it does not depend on which other candidates are in $T$, on their slots, or on $|T|$.
H holds by construction of the two-image message whenever every message is one forward pass, which is the only path the training code accepts for a scored group.
On the batched serving path used for some attack rows, the pair signs agree with the one-pass path in 99.72\% of pairs and the \PWC winner in 99.50\% of 1{,}200 three-image sets, so H holds there up to bf16 batching noise; a winner that differs between the two paths is a serving effect, not a failure of the aggregation statements below.
Under H, inserting a candidate, removing one, or swapping the content of a slot leaves every other pair's margin unchanged, so the three cases below are one statement about the relation.

\paragraph{The relation and the readout.}
$x$ \emph{beats} $y$, written $x\beats y$, when $m(x,y)>0$; a zero or undefined margin is a tie, a win for neither.
$\beats$ is irreflexive and asymmetric and need not be complete.
For $x\in T$ let $W_T(x)=\{z\in T: x\beats z\}$.
$y$ \emph{covers} $x$ in $T$ when $y\beats x$ and $W_T(x)\subseteq W_T(y)$: the downward (Bordes) covering relation of a weak tournament~\cite{bordes1983possibility,brandt2008computing,duggan2013uncovered}, Miller's on complete tournaments~\cite{miller1980new}; a tie never counts toward covering, because a tie is not in $W$.
$C(T)$ is the set of candidates covered in $T$.
The readout $w(T)$ has the three steps of the main paper, with the tie-break of step (3) split in two:
(1) remove $C(T)$ and repeat on the survivors until nothing is covered; the end set is $N(T)$;
(2) give each survivor its Copeland score over $N(T)$, one point per survivor it beats and half per tie;
(3) if several survivors share the top score, give each of them the sum of its margins against the other tied survivors (undefined margins count 0) and keep the maximisers;
(4) if more than one remains, take the lowest slot.

\paragraph{Tie convention.}
Step (2) reads only pairs among $N(T)$ and step (3) only pairs among the tied survivors, a subset of $N(T)$.
Step (4) is the only step that reads a slot; it is reached when both the Copeland score and the margin sum among the tied survivors tie exactly.
With two tied survivors their sums are $m(x,z)$ and $-m(x,z)$, so step (4) is reached only when their own pair is a tie or undefined; with three or more it needs an exact equality of real-valued sums.
In the 384 stored eight-rollout selection groups (256 best-of-8 groups and 128 audit groups), Copeland ties among the survivors occurred in 11 and 9 groups, all among three survivors; the margin sum decided all 20, and step (4) was never reached.
Replacing only the slot key of step (4) by a hash of the image changes no stored selection.
Replacing steps (3) and (4) together by the hash changes the winner in 6 of the 11 and 4 of the 9 tied groups (\cref{sec:s-leader}).

\subsection{Lemmas and the selection theorem}

\begin{lemma}\label{lem:s-props}
(a) Covering is irreflexive and asymmetric. (b) It is transitive. (c) It is hereditary: if $y$ covers $x$ in $T$ and $T'\subseteq T$ holds both, $y$ covers $x$ in $T'$. (d) It is persistent: if $x,w\in C(T)$ and $x\ne w$, then $x\in C(T\setminus\{w\})$.
\end{lemma}
\noindent\emph{Proof.}
(a) $y$ covers $x$ requires $y\beats x$, and $\beats$ is irreflexive and asymmetric.
(b) If $u$ covers $y$ and $y$ covers $x$, then $x\in W_T(y)\subseteq W_T(u)$, so $u\beats x$, and $W_T(x)\subseteq W_T(y)\subseteq W_T(u)$.
(c) $W_{T'}(x)=W_T(x)\cap T'\subseteq W_T(y)\cap T'=W_{T'}(y)$, and $y\beats x$ is unchanged.
(d) Let $y$ cover $x$ in $T$. If $y\ne w$, (c) applies. If $y=w$, let $v$ cover $w$ in $T$; $v\ne w$ by (a) and $v\ne x$, since $x$ covering $w$ and $w$ covering $x$ would contradict (a). By (b) $v$ covers $x$ in $T$ and by (c) in $T\setminus\{w\}$. \hfill$\square$

\begin{lemma}[Order independence]\label{lem:s-order}
Every maximal sequence of single removals $T\to T\setminus\{x\}$ with $x\in C(T)$ ends at the same set $N(T)$; a simultaneous round of step (1) is such a sequence, so the readout computes $N(T)$; and $N(T)\ne\emptyset$ for $T\ne\emptyset$.
\end{lemma}
\noindent\emph{Proof.}
Removals terminate because $|T|$ decreases.
They are locally confluent: for distinct $x,w\in C(T)$, \cref{lem:s-props}(d) gives $w\in C(T\setminus\{x\})$ and $x\in C(T\setminus\{w\})$, so both one-step results reach $T\setminus\{x,w\}$.
By Newman's lemma a terminating, locally confluent rewriting system has unique normal forms.
A simultaneous round removes $C(T)$; by (d) applied repeatedly, the members of $C(T)$ stay covered while the others are removed one at a time, so the round is a sequence of single removals.
Finally, covering is a strict partial order by \cref{lem:s-props}(a, b), so a finite non-empty $T$ has an uncovered element. \hfill$\square$

\begin{theorem}[Selection; restates \mainref{thm:select}]\label{prop:s-ddp}
Let $S'=S\cup\{d\}$ with the members of $S$ in the same relative slot order in both sets and the same margins among them (H).
(a) If $d\in C(S')$, then $N(S')=N(S)$; the Copeland scores, the tied set and the margin sums over the survivors are the same; the slot key selects the same candidate; so $w(S')=w(S)$, and no margin that involves $d$ enters steps (2)--(4).
(b) For any bijection $\pi$ of the slots of $S$ with the margins transported, $w(\pi S)=\pi\,w(S)$ unless step (4) is reached on $S$.
\end{theorem}
\noindent\emph{Proof.}
(a) $S'\to S'\setminus\{d\}=S$ is a legal single removal, so by \cref{lem:s-order} $N(S')=N(S)$.
Steps (2) and (3) read only margins among $N(S)\subseteq S$, which are the same in both sets, so the scores, the tied set and its margin sums coincide.
Step (4) compares slots: inserting $d$ at any slot maps the slots of $S$ to those of $S'$ by a strictly increasing function, and a swap at a fixed slot leaves them unchanged, so the minimum over the same tied set is attained by the same candidate.
Since $d\notin N(S')$, no pair that contains $d$ is read after step (1); the released readout logs every pair it reads, and the log is checked on every stored row.
(b) Under H the transported relation is $\pi(\beats)$, so $W$, covering, $C$, the removal sequence, $N$, the Copeland scores and the margin sums among the tied are all carried by $\pi$: they are defined from the set and the relation, not from the slots.
If step (3) leaves one candidate, it is $\pi\,w(S)$.
If it leaves several, step (4) returns the one with the lowest slot under $\pi$, which depends on $\pi$. \hfill$\square$

\begin{remark}\label{rem:s-select}
(i) The removal direction is the same statement read from $S'$ to $S$: removing a covered candidate that is not the winner leaves the winner.
(ii) Any slot-independent key in step (4), for example the lowest hash of the image bytes, makes (b) unconditional, since every other step is label-free; two candidates with identical bytes are interchangeable.
(iii) A candidate that beats nobody and loses to at least one is covered, ties allowed, because its $W$ is empty; a universal loser, one that loses every pair, is the special case.
(iv) H is where the judge enters: the theorem is about the aggregation given the margins, so it makes the selection decoy-proof end to end exactly when the contender margins do not depend on $d$, which the two-image message gives for swaps and additions alike.
(v) The cover winner is a different selector from the argmax of the Copeland scores over all candidates.
On the 256 best-of-8 groups the cover winner equals a unique Copeland argmax in 95.7\% [93.0, 98.0] of groups and lies in the Copeland top set in 99.2\%; 4.3\% of the groups contain a cycle among their candidates; and best-of-8 recovers 64.2\% of the oracle gain with the cover winner against 62.7\% with the Copeland argmax, ties split.
In the pilot training groups of the reward-path review the two agree in 96--97\% of groups, and 5--6\% of those groups contain a cycle.
\end{remark}

\subsection{Why the hypothesis and the cover step are needed}
\label{sec:s-counter}

\paragraph{Covered is strictly larger than universal loser, and the cover step is what pays for it.}
Let $A\beats B$, $B\beats C$, $C\beats A$ with $m(A,B)=0.5$, $m(B,C)=3$, $m(C,A)=3$, and add $d$ with $A\beats d$, $B\beats d$, $d\beats C$.
$d$ is not a universal loser, yet $B$ covers it ($B$ beats $d$ and beats $C$, the only candidate $d$ beats), so $d\in C(S')$.
On $S$ nothing is covered, every Copeland score is 1, and the margin sums $-2.5$, $2.5$, $0$ pick $B$; on $S'$ the readout removes $d$ first and again picks $B$.
Copeland without the cover step, with the same tie-break among the tied, scores $A$ and $B$ 2, $C$ and $d$ 1, and picks $A$ by $m(A,B)>0$: a merely covered entrant moves its winner from $B$ to $A$.
This is the pattern behind the 12.6--12.9\% of enumerated covered-only configurations in which Copeland with a tie-break among the tied and insertion sort move the contender winner (\cref{tab:s-rules}).

\paragraph{The hypothesis is needed.}
Let $A\beats B$, $A\beats C$, $B\beats C$, so $w(S)=A$.
Add $d$ with $d\beats A$, $B\beats d$, $C\beats d$: no candidate beats both $d$ and $A$, so $d$ is not covered.
$C$ is covered by $B$ and removed; $\{A,B,d\}$ is a three-cycle with Copeland scores $1,1,1$, and the margin sums can pick $B$ or $d$.
A decoy that beats a contender is not irrelevant, and no rule can both honour that win and ignore it.

\paragraph{Why step (1), and not Copeland over all candidates.}
Let $A\beats B$, $B\beats C$, $C\beats A$ with margins such that the three-way tie-break picks $C$, and add $d$ that $A$ and $B$ beat and $C$ ties.
Copeland over all four gives $A$ and $B$ two points and $C$ one and a half, so that winner moves to $A$ although every contender-pair margin is unchanged; $d$ is covered by $A$ ($W(d)=\emptyset$), so the readout removes it first and still picks $C$.
The same mechanism, one more term per contender, is why the rules without a cover step in \cref{tab:s-rules} move a winner even for a universal loser: plain Copeland's tie-break sums every margin including the margin over $d$, and Borda, Bradley--Terry and Elo add each contender's own margin or win probability against $d$, terms that differ between contenders unless every contender's result against $d$ is the same, which is exactly the condition of \cref{prop:s-reward}.

\subsection{The training reward}

\begin{proposition}[Reward; \mainref{sec:reward}]\label{prop:s-reward}
Let $R$ be a group of $G\ge2$ scored rollouts with win counts $W_i=\sum_{j\in R,\,j\ne i}v_{ij}$, where $v_{ij}=1,\tfrac12,0$ for $m(i,j)>0$, $m(i,j)=0$ or undefined, $m(i,j)<0$; the reward is $c_i=W_i/(G-1)$ and the advantage $A_i=(c_i-\bar c)/(\sigma+\varepsilon)$, with $\bar c$ and $\sigma$ the mean and population standard deviation of $(c_j)_{j\in R}$, $\varepsilon=10^{-4}$, then clipped to $[-5,5]$.
Let a candidate $d$ enter the comparisons with the same result from every rollout, $v_{id}=v$ for all $i\in R$, and the margins among $R$ unchanged (H); the universal loser is $v=1$.
Then:
(i) $W'_i=W_i+v$, so $W'_i-W'_j=W_i-W_j$ for all $i,j\in R$;
(ii) $c'_i=\frac{G-1}{G}c_i+\frac vG$, one positive affine map for every $i$, so the ranking of $R$ and all ratios of reward differences are unchanged;
(iii) if the advantage is normalised over $R$ ($d$ unscored), $A'_i=A_i\cdot\frac{\sigma+\varepsilon}{\sigma+\varepsilon G/(G-1)}$: identical at $\varepsilon=0$, otherwise a common positive factor $1-\varepsilon/((G-1)\sigma)+O(\varepsilon^2)$, within $5\cdot10^{-5}$ of one at the observed group standard deviation $\sigma\approx0.32$ and $G=8$; the clip never binds, since $|A_i|\le\sqrt{G-1}=\sqrt7$;
(iv) if $d$ is scored and its own $c'_d$ enters the mean and standard deviation, then $A'_i=\gamma A_i+\delta$ on $R$ with $\gamma>0$ (for $\sigma>0$): the ranking and the sign of every difference $A'_i-A'_j$ are those of $A$, while the sign of an individual $A'_i$ can change.
Conversely, if two rollouts receive different results against $d$, $v_{id}\ne v_{jd}$, then $W'_i-W'_j\ne W_i-W_j$: (i) fails, and (ii)--(iii) fail in general.
\end{proposition}
\noindent\emph{Proof.}
(i) $W'_i=\sum_{j\in R,\,j\ne i}v_{ij}+v_{id}=W_i+v$ by H.
(ii) $c'_i=W'_i/G$, because $d$ adds one opponent.
(iii) Write $c'=ac+b$ with $a=\frac{G-1}{G}$ and $b=\frac vG$; the mean and the population standard deviation over $R$ obey $\bar c'=a\bar c+b$ and $\sigma'=a\sigma$, so $A'_i=a(c_i-\bar c)/(a\sigma+\varepsilon)=(c_i-\bar c)/(\sigma+\varepsilon/a)$, whose ratio to $A_i$ is $(\sigma+\varepsilon)/(\sigma+\varepsilon/a)$; $|A_i|\le\sqrt{G-1}$ is the bound on a population $z$-score.
(iv) Over $R\cup\{d\}$ with mean $\bar c''$ and standard deviation $\sigma''$, $A'_i=(ac_i+b-\bar c'')/(\sigma''+\varepsilon)$ is increasing affine in $c_i$, and so is $A_i$; composing gives $\gamma=a(\sigma+\varepsilon)/(\sigma''+\varepsilon)>0$ and some $\delta$, which is not zero in general because $c'_d$ moves the mean.
The converse is (i) with $v_{id}-v_{jd}\ne0$. \hfill$\square$

\begin{remark}\label{rem:s-reward}
(i) \emph{What the trained arm does.} The \PWC reward queries only pairs of scored rollouts and never shows an extra candidate to the pair judge, so with an extra present its reward is bitwise the same as without: the extra enters no comparison at all. \Cref{prop:s-reward} is the statement for a universal loser that \emph{does} enter the comparisons, as an unscored candidate of a pairwise group (case iii) or as a scored member (case iv). In ordinary GRPO every candidate is a scored rollout, so case (iii) describes mixed-source groups and reused selectors, and case (iv) a group in which one rollout loses every pair, which the exchange test of \cref{sec:s-exchange} constructs from base-model rollouts.
(ii) \emph{Hard wins.} The reward uses $v_{ij}\in\{1,\tfrac12,0\}$. A soft variant, the mean of $\mathrm{sigmoid}(m(i,j))$ over $j$, is logged as a diagnostic and never used; under it a universal loser adds $\mathrm{sigmoid}(m(i,d))/G$ to rollout $i$, which differs between rollouts unless every margin over $d$ is equal, and the differences change. This is the failure of Borda and Bradley--Terry on win probabilities in \cref{tab:s-rules}; with hard wins the within-group win rate, including \PairOne's one-order win rate, is invariant to a universal loser in the sense of (i)--(iii).
(iii) \emph{Selection and reward are different guarantees.} \Cref{prop:s-ddp}(a) holds for every covered $d$; \cref{prop:s-reward} needs the same result from every rollout. A merely covered $d$ beats some rollouts and not others, so it changes the reward differences while it cannot change the selection winner. Neither statement explains the training difference between rewards measured in \mainref{sec:training}; that comparison is empirical, and we do not attribute the training gain to the cover rule.
(iv) \emph{Order.} The reward vector has no tie-break, so under H it is permutation-equivariant without exception; the exception in \cref{prop:s-ddp}(b) concerns the selection winner only.
\end{remark}

\subsection{Cost, with a certificate}
\label{sec:s-leader}

\paragraph{Cost.}
\PWC asks $N(N{-}1)$ two-image messages, 56 at $N=8$, which on q8 take 4.93\,s per group against 0.22\,s for one \List message and 1.66\,s for \ListAvg's eight.
For selection, a proposed leader that beats every other candidate covers them all, so its $2(N-1)$ messages certify a selection identical to that of full \PWC (\cref{prop:s-leader}).
The certificate holds at some leader in 95.7\% of the 256 selection groups, and certified selection averages 24.3 messages per group against the 56 that the training reward needs (\cref{tab:s-leader}).
In the terms of \mainref{tab:protocols}, certified selection takes 24.3 messages, 23.3 pair messages $+$ 1 \List message, with Reorder 0.0 and Best-of-8 64.2; its selections are identical to full \PWC, with the full-matrix fallback in 4.3\% of groups.

\begin{proposition}[Certified leader]\label{prop:s-leader}
If $L\beats x$ for every $x\in T\setminus\{L\}$, then $C(T)=T\setminus\{L\}$, $N(T)=\{L\}$ and $w(T)=L$.
Hence a procedure that queries $L$'s $|T|-1$ pairs in both orders, $2(|T|-1)$ messages, and stops when $L$ wins every one of them returns $w(T)$ without the remaining $(|T|-1)(|T|-2)$ messages, and a run that completes the matrix when the certificate fails returns $w(T)$ in every case.
\end{proposition}
\noindent\emph{Proof.}
For $x\ne L$: $L\beats x$, and $W_T(x)\subseteq T\setminus\{x,L\}\subseteq W_T(L)=T\setminus\{L\}$, because $x$ does not beat $L$ by asymmetry; so $L$ covers $x$.
Nobody beats $L$, so $L$ is uncovered.
One round removes $T\setminus\{L\}$, step (2) is trivial, and $w(T)=L$.
By H the margins of $L$'s pairs are the same whether or not the other pairs are queried, so $L$ is the Condorcet winner of the full relation and the full readout also returns $L$; when the certificate fails, completing the matrix is the full readout by definition. \hfill$\square$

\begin{remark}
A tie or an undefined margin in any of $L$'s pairs voids the certificate.
The Copeland reward of every rollout needs its own $G-1$ pairs, hence all $G(G-1)/2$ unordered pairs, so early stopping reduces the cost of selection and not of the training reward.
\end{remark}

\paragraph{Replay on stored matrices.}
The procedure proposes a leader and issues its pairs against every other candidate in both orders.
If the leader beats all of them, it stops (\cref{prop:s-leader}).
Otherwise the next leader is the candidate with the largest margin over the current one among those that beat it and have not yet led, and pairs already issued are reused; when no such candidate exists (a cycle, or a tie with the leader), the remaining pairs are issued and the full readout returns the winner.
If a Condorcet winner exists it beats every leader before it, so the chain reaches it; the fallback is needed only without one.
We replayed it on the stored pair margins of the 256 best-of-8 groups and the 128 audit groups ($N=8$, 56 pair messages for the full matrix), with no new judge query.
The stored matrices were rebuilt from the pair-query replays with 0 mismatches (7{,}168 and 3{,}584 pairs); the certified output equals the full readout's winner in every group (256 of 256 and 128 of 128, for every proposal); and the selections reproduce the stored best-of-8 of the cover winner, 64.2\% [57.9, 70.2] of the oracle gain.
\Cref{tab:s-leader} gives the message counts.
With the \List proposal the procedure averages 23.34 pair messages, 24.34 with the proposal message, against 56; at the measured per-message time this projects to 2.26\,s per group against 4.93\,s for the full matrix, a projection rather than a timed online run.
The full-matrix fallback ran in the groups without a Condorcet winner, 4.3\% [2.0, 7.0] of the best-of-8 groups and 7.0\% [3.1, 11.7] of the audit groups.

\begin{table*}[t]
  \centering
  \scriptsize
  \setlength{\tabcolsep}{4pt}
  \caption{\textbf{Certified-leader early stopping, replayed on stored pair matrices} ($N=8$; pair messages per group out of 56; 95\% bootstrap intervals over groups). First leader: share of groups in which the proposed leader certifies at once; fallback: share needing the full matrix (the remaining groups certify at a later leader). Proposals: the top pick of the stored \List message in identity order (\List) or in a permuted base order (\List, perm.), slot 0, or the stored \Rating top pick; the \List proposals cost one extra \List message, added in the ``with proposal'' column and in the projected seconds; \Rating's eight single-image messages are not added. Selections are identical to full \PWC in every group.}
  \label{tab:s-leader}
  \begin{tabular}{@{}llrrrrrrr@{}}
    \toprule
    groups & proposal & first leader \% & fallback \% & mean pair msgs & median & p90 & with proposal & projected s \\
    \midrule
    best-of-8 (256) & \List & 43.4 [37.1, 49.6] & 4.3 [2.0, 7.0] & 23.34 [22.09, 24.65] & 26 & 36 & 24.34 & 2.26 \\
    & slot 0 & 10.5 [7.0, 14.5] & 4.3 [2.0, 7.0] & 28.74 [27.66, 29.84] & 26 & 36 & 28.74 & 2.53 \\
    & \Rating top pick & 43.8 [37.5, 49.6] & 4.3 [2.0, 7.0] & 23.16 [21.92, 24.50] & 26 & 36 & 23.16 & 2.04 \\
    & \List, perm. & 48.0 [41.8, 54.3] & 4.3 [2.0, 7.0] & 22.58 [21.36, 23.91] & 26 & 36 & 23.58 & 2.19 \\
    \midrule
    audit (128) & \List & 35.9 [27.3, 44.5] & 7.0 [3.1, 11.7] & 24.98 [23.03, 27.00] & 26 & 36 & 25.98 & 2.42 \\
    & slot 0 & 14.8 [9.4, 21.1] & 7.0 [3.1, 11.7] & 28.84 [27.16, 30.64] & 26 & 38.4 & 28.84 & 2.54 \\
    & \Rating top pick & 46.1 [37.5, 54.7] & 7.0 [3.1, 11.7] & 23.66 [21.70, 25.73] & 26 & 36 & 23.66 & 2.08 \\
    & \List, perm. & 38.3 [29.7, 46.9] & 7.0 [3.1, 11.7] & 24.12 [22.30, 26.11] & 26 & 36 & 25.12 & 2.34 \\
    \bottomrule
  \end{tabular}
\end{table*}

\paragraph{Tie-break variants on the same groups.}
On the 256 best-of-8 groups the implemented rule recovers 64.2\% [57.9, 70.2] of the oracle gain.
Replacing only the slot key by a hash of the stored file name changes no selection (64.2\%).
Replacing the margin sum and the slot key together by the hash among the Copeland-tied survivors gives 61.6\% [54.8, 68.0] (paired difference to the implemented rule $-2.6$ [$-5.9$, $-0.2$]); a uniform choice among the Copeland-tied survivors gives 62.4\% [56.0, 68.4] in exact expectation.
On the 128 audit groups (hash of the PNG bytes) the same three numbers are 59.3\% [48.6, 68.8] (no change), 61.3\% [50.7, 70.7] (paired $+2.1$ [$-0.1$, $+4.8$]) and 58.3\% [47.6, 67.9].
The margin-sum step therefore matters in a few groups in either direction, and the slot key matters in none.
On the 256 groups the implemented rule leads the Copeland arg\,max over all candidates with ties split, 62.7\%, by $+1.5$ [$+0.1$, $+3.1$] points, and the uniform choice among the survivors separates its two steps: the margin tie-break adds 1.8 points (64.2 against 62.4), and the cover step with a uniform choice among the tied survivors gives 0.3 points less than the Copeland arg\,max (62.4 against 62.7).

\subsection{Rankings by successive selection}

Successive selection ranks $S$ by $\rho(S)=(w_1,w_2,\dots)$ with $w_k$ the readout's winner on $S_k=S\setminus\{w_1,\dots,w_{k-1}\}$.

\begin{proposition}[Stagewise covering]\label{prop:s-rank}
Let $\rho'$ be the order of $S$ obtained by successive selection on $S\cup\{d\}$ with $d$'s place dropped. If $d\in C(S_k\cup\{d\})$ for every stage $k$ with $S_k\ne\emptyset$, then $\rho'=\rho(S)$ and $d$ is selected last; in particular when every candidate of $S$ beats $d$.
\end{proposition}
\noindent\emph{Proof.}
By induction on $k$: if the first $k-1$ places with $d$ present are $w_1,\dots,w_{k-1}$, the remaining set is $S_k\cup\{d\}$, and \cref{prop:s-ddp}(a) on $S_k$ gives the winner $w_k$, which is not $d$ since $d$ is covered. If every candidate beats $d$, $W(d)=\emptyset$ at every stage and any remaining candidate covers $d$. \hfill$\square$

\noindent Covering at the first stage is not enough: with $A$ beating $B$, $C$ and $d$, $B\beats C$, $C\beats d$ and $d\beats B$, $A$ covers $d$ at the first stage, but $\{B,C,d\}$ is a three-cycle at the second, where the tie-break can place $C$ above $B$ although $B\beats C$.

\subsection{Which aggregation rules keep which guarantee}
\label{sec:s-aggregation}

\Cref{tab:s-design} classifies the readouts of the main paper and the common pairwise aggregations by the guarantee they keep, beside the measured searched excess and reorder rate; \cref{tab:s-rules} gives the exhaustive check behind the classification.
The enumeration runs each rule over every margin configuration on a grid of three contenders, ties and cycles included, and one added candidate inserted at each of the four slots: 13{,}500 configurations in which the added candidate loses every pair and 23{,}400 in which it is covered but wins some pair, 36{,}900 enumerated margin configurations in all, and counts the configurations in which the winner among the contenders changes.
Only the cover rule is unchanged in both classes.
Copeland with a tie-break among the tied and insertion sort are unchanged by a universal loser but not by a merely covered candidate; plain Copeland, Elo, Borda and Bradley--Terry on win probabilities or margins, a knockout bracket and merge sort move the winner even for a universal loser.
On the stored development rows of q8 and q32 the differences are small, because 99.6\% of the stored decoys are universal losers under the pair winners (\cref{sec:s-pairwise-attack}): the cover rule reproduces \PWC exactly (maximum absolute difference 0.0 on every endpoint), and the knockout bracket leaves no residual but loses pair agreement on K3+ (q8 74.1\% against 80.4\%).

\begin{table*}[t]
  \centering
  \scriptsize
  \setlength{\tabcolsep}{4pt}
  \caption{\textbf{Guarantees kept by each readout or aggregation.} Unchanged by $d$: the selected contender cannot change when a candidate that loses every pair (universal loser) or a merely covered candidate is added, with the margins among the contenders fixed (\yes: never; \no: changes in some configuration; source in the last column). Messages: per group of $N$. Excess: searched excess on the development sets, q8 / q32 (\pp, \cref{sec:s-attack}). Reorder: share of the 128 audit groups ($N=8$, q8) whose top pick changes under a permutation of the same images.}
  \label{tab:s-design}
  \begin{tabular}{@{}llcccll@{}}
    \toprule
    readout / aggregation & messages & univ.\ loser & covered & excess q8 / q32 & reorder & source of the guarantee columns \\
    \midrule
    \List, one message & 1 & \no & \no & 31.1 / 22.4 & 62.5\% & measured (\cref{tab:s-k1}) \\
    \ListAvg, $N$ cyclic rotations & $N$ & \no & \no & 26.3 / 24.7 & 25.0\% & measured (\cref{sec:s-listavg}) \\
    \Rating, each image alone & $N$ & \yes & \yes & 0.3 / 0.3 & 0 & by construction (no other candidate is read) \\
    \PairOne, one order, win count & $N(N{-}1)/2$ & \yes & \no & 2.8 / 1.0 & 48.4\% & \cref{prop:s-reward}(i); counterexample of \cref{sec:s-counter} \\
    \PRP, both orders, hard wins & $N(N{-}1)$ & \yes & \no & 3.8 / 2.3 & 15.6\%$^\dagger$ & \cref{prop:s-reward}(i); counterexample of \cref{sec:s-counter} \\
    pairs + Copeland (tie among tied) / insertion & $N(N{-}1)$ & \yes & \no & -- & -- & enumeration (\cref{tab:s-rules}) \\
    pairs + Borda / Bradley--Terry / Elo & $N(N{-}1)$ & \no & \no & -- & -- & enumeration \\
    pairs + knockout / merge sort & $<N(N{-}1)$ & \no & \no & -- & -- & enumeration \\
    \PWC, pairs + cover & $N(N{-}1)$ & \yes & \yes & 1.4 / 1.3 & 0$^\ddagger$ & \cref{prop:s-ddp}; enumeration \\
    \bottomrule
  \end{tabular}
  \par\smallskip\parbox{\textwidth}{\scriptsize $^\dagger$ \PRP's paper tie rule sends a tied top score to the first-shown position; its top score is tied in 23.4\% of these groups, and with a uniform tie-break its searched excess is 0.7 / 0.5. \PairOne and \PRP keep the universal-loser guarantee because a universal loser adds the same point to every contender's count and insertion preserves the contenders' relative order. $^\ddagger$ Exact unless both the Copeland score and the margin sum tie (never in the 384 stored groups, \cref{sec:s-defs}).}
\end{table*}

\begin{table}[t]
  \centering
  \scriptsize
  \setlength{\tabcolsep}{2.5pt}
  \caption{\textbf{Eleven aggregations over the same pair results.} Changes: configurations in which the winner among the contenders changes when $d$ is added, out of 13{,}500 with $d$ a universal loser and 23{,}400 with $d$ covered but winning some pair. Residual: on the stored development rows (\pp of sets; q8 $n=289$, q32 $n=299$).}
  \label{tab:s-rules}
  \begin{tabular}{@{}lrrcc@{}}
    \toprule
    rule & univ.\ loser & covered & q8 & q32 \\
    \midrule
    cover (\PWC) & 0 & 0 & 0.0 & 0.3 \\
    Copeland, tie among tied & 0 & 2{,}960 (12.6\%) & 0.0 & 0.3 \\
    insertion sort & 0 & 3{,}016 (12.9\%) & 0.0 & 0.0 \\
    Copeland, plain & 1{,}136 (8.4\%) & 2{,}956 (12.6\%) & 0.0 & 1.0 \\
    Elo, wins & 862 (6.4\%) & 3{,}200 (13.7\%) & 0.0 & 0.3 \\
    Elo, win probabilities & 2{,}888 (21.4\%) & 4{,}663 (19.9\%) & 4.8 & 2.3 \\
    Bradley--Terry, probabilities & 2{,}744 (20.3\%) & 4{,}588 (19.6\%) & 4.2 & 2.3 \\
    Borda, probabilities & 2{,}744 (20.3\%) & 4{,}588 (19.6\%) & 4.2 & 2.3 \\
    Borda, margins & 4{,}424 (32.8\%) & 4{,}588 (19.6\%) & 7.6 & 3.3 \\
    knockout bracket & 2{,}461 (18.2\%) & 4{,}969 (21.2\%) & 0.0 & 0.0 \\
    merge sort & 1{,}620 (12.0\%) & 4{,}420 (18.9\%) & 0.0 & 0.3 \\
    \bottomrule
  \end{tabular}
\end{table}

\subsection{Relation to known results and tests}

\paragraph{Relation to known results.}
$N(T)$ is the iterated uncovered set~\cite{laslier1997tournament,brandt2008computing} of the downward covering relation and \cref{lem:s-props}(b) is the standard transitivity of covering~\cite{duggan2013uncovered}; \cref{prop:s-ddp}(a) is a stability property in the spirit of Dutta's covering sets~\cite{dutta1988covering}: an entrant that the relation covers leaves the iterated uncovered set unchanged.
What we add is the reading for relations with ties and undefined results, the Copeland count and the tie convention over the survivors, the separation of the selection statement from the reward statement, and the certificate; we claim no monotonicity.
\PRP~\cite{qin2024prp} also queries both orders but aggregates without a cover step, so the covered-candidate guarantee of \cref{prop:s-ddp}(a) is not available to it by this proof; its measured behaviour is in \cref{M-sec:pair-results,M-tab:protocols} of the main paper and \cref{sec:s-pairwise-attack}.

\paragraph{Tests.}
The released test suite checks \cref{lem:s-props} on 3{,}000 random relations; order independence over every relation on 3 and 4 candidates with wins, losses, ties and undefined results along all removal paths, and over 20{,}000 random relations on 3--7 candidates; \cref{prop:s-ddp}(a) exhaustively for 3 contenders, 4 decoy slots and every decoy result in \{win, loss, tie, none\}, plus 20{,}000 random cases; the ``hypothesis is needed'' example; and, for the aggregation rules of \cref{tab:s-rules}, that the hard-win rate is invariant to a universal loser and that plain Copeland is moved by one through its tie-break.
On the stored rows, every readable \PWC row whose decoy is covered under its own pair winners must return the contenders' winner and log no read of a pair that holds $d$; a violation is a code fault.

\clearpage
\section{Extended attack results}
\label{sec:s-attack}

All rows in this section are on the development sets unless marked held-out; every flip is closed under the dominance check (\cref{sec:s-steps}).
The main paper's per-judge table (\mainref{tab:judges}) prints its intervals at 95\%; some development rows below are only available at 97.5\% and are marked.

\subsection{All endpoints per judge}
\label{sec:s-judges}

\begin{figure*}[t]
  \centering
  \figfile{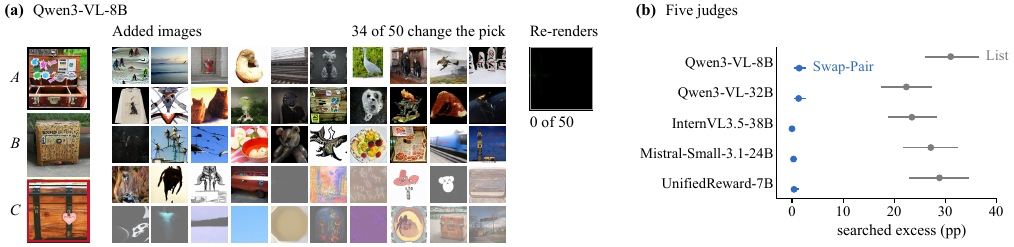}{6.755in}{4.2cm}{figAB\_v11.pdf: (a) one set, fifty added images and fifty re-renders; (b) five-judge searched excess}
  \vspace{-0.6em}
  \caption{\textbf{Fifty added images against fifty re-renders.}
  \emph{(a)}~Qwen3-VL-8B under \List picks $A$ (black frame) from one HPDv2 set.
  Each added image is rated below $A$, $B$ and $C$ by the same judge alone; 34 of the fifty move the pick to $C$ (red frame; the rest faded), and none of fifty re-renders of the first pool image (right; a near-black photograph) does.
  \emph{(b)}~Searched excess of \List and \PWC on five judges, 95\% intervals (\cref{tab:s-judges}).}
  \label{fig:s-fifty}
\end{figure*}

\Cref{tab:s-judges} gives every endpoint per judge, of which \mainref{tab:judges} prints the fifty-try columns and the searched excess of \PWC, and \cref{fig:s-fifty} shows the fifty tries on one set and the searched excess on five judges.
The judge rates each dominated decoy alone on a 1--10 scale (ur7: its own 1--5 score), and the study reads every judge from its first-token label distributions.
\List reads $P(\text{winner}=i)$ from the first-token distribution over the label tokens, and the added image of a set is drawn from a pool of 70 candidates per set.
The re-renders of the fixed decoy that give the noise floor vary JPEG quality 85--95, $\pm1$\,px and $\pm1\%$ brightness (\cref{sec:s-floors}).
\emph{One added image} is the share of sets whose winner among the contenders changes when one image joins the decoy-free triple, for a random decoy, one re-render of a fixed decoy and a grey square.
For \mainref{sec:pairwise} we also read the \emph{residual}, the share of sets in which some decoy moves the winner to another contender after every flipping decoy is re-verified under the readout's own pair winners.
\emph{Human agreement} is the winner agreement with the HPDv2 ranking on 1{,}200 three-image sets minus \List's.
A second task family, VL-RewardBench~\cite{li2024vlrewardbench}, ranks three text responses to one image, one of them a text decoy (486 items, q8).
Estimates are means with 95\% percentile-bootstrap intervals (5{,}000 resamples; sets, prompts or seeds as the unit), and the held-out sets were judged once, after every readout, endpoint and threshold was fixed.

The excess replicates for q32 on the held-out prompts and also appears for q8 with text decoys on VL-RewardBench (\cref{tab:s-judges}).
Unscreened images move the winner too: fifty images of other prompts, which need no access to the judge, change q32's contender choice in 42.4\% of sets, $+25.3$\,\pp above the floor (\cref{sec:s-natural}; transfer in \cref{sec:s-transfer} and targeted flips in \cref{sec:s-budget}).

\begin{table*}[t]
  \centering
  \scriptsize
  \setlength{\tabcolsep}{3.2pt}
  \caption{\textbf{Every endpoint per judge.}
  Read by \PWC, the same judges leave a residual of at most 0.3\,\pp.
  HPDv2 development sets ($N=4$; $n$ sets with a judge-verified dominated decoy), a held-out replication and the text family; \% of sets, 95\% intervals; endpoints as in \mainref{sec:measure-setup} and \cref{sec:s-judges}$^{a,b,d}$.}
  \label{tab:s-judges}
  \begin{tabular}{@{}l r cc cc l c cc l@{}}
    \toprule
    & & \multicolumn{2}{c}{One added image} & \multicolumn{3}{c}{Fifty tries, \List} & \ListAvg$^{d}$ & \multicolumn{2}{c}{\PWC} & $\Delta$Human$^{d}$ \\
    \cmidrule(lr){3-4}\cmidrule(lr){5-7}\cmidrule(lr){8-8}\cmidrule(lr){9-10}
    Judge & $n$ & decoy & re-render & decoys & re-renders & $\Eone$ [95\% CI] & $\Eone$ & $\Eone$ & residual & \PWC $-$ \List [95\% CI] \\
    \midrule
    q8          & 289 & 26.8 & 26.2 & 40.1 & 9.0  & $+31.1$ [$+26.0$, $+36.7$] & 26.3 & 1.4 & 0.0 & $+1.4$ [$+0.1$, $+2.8$] \\
    q32         & 299 & 22.0 & 22.4 & 39.5 & 17.1 & $+22.4$ [$+17.4$, $+27.4$] & 24.7 & 1.3 & 0.3 & $+3.4$ [$+2.1$, $+4.8$] \\
    ivl38$^{c}$ & 298 & 18.6 & 18.9 & 33.6 & 10.1 & $+23.5$ [$+18.8$, $+28.5$] & --   & 0.0 & 0.0 & -- \\
    mst24       & 298 & 20.6 & 21.3 & 45.3 & 18.1 & $+27.2$ [$+21.8$, $+32.6$] & --   & 0.3 & 0.0 & -- \\
    ur7$^{c}$   & 239 & 23.6 & 24.5 & 46.4 & 17.6 & $+28.9$ [$+23.0$, $+34.7$] & --   & 0.4 & 0.0 & -- \\
    \midrule
    q32 held-out   & 100 & -- & -- & --   & --  & $+20.0$ [$+12.0$, $+28.0$] & -- & 0.0 & 0.0 & $+3.3$ [$+1.3$, $+5.3$] \\
    q8 text, $N=3$ & 486 & -- & -- & 45.9 & 9.7 & $+36.2$ [$+31.9$, $+40.5$] & -- & --  & --  & -- \\
    \bottomrule
  \end{tabular}
  \vspace{2pt}
  \begin{minipage}{\textwidth}\scriptsize
  $^{a}$A grey square moves the winner at similar rates (q8 22.4, q32 20.6, ivl38 14.7, mst24 19.9, ur7 19.6\%; decoy minus grey at most $+4.4$\,\pp); decoy minus re-render lies within $\pm1$\,\pp on every judge.
  $^{b}$The \List and \ListAvg columns use the dominance check of the listwise search; the \PWC columns use the later check of the pairwise study, under which \List's $\Eone$ reads $+31.5$, $+23.4$ and $+22.5$ on q8, q32 and ivl38 (\cref{sec:s-variants}).
  $^{c}$ivl38 abstains often; on the 180 sets readable throughout its $\Eone$ is $+19.4$ [$+13.3$, $+25.6$]. ur7: 61 of 300 sets had no decoy it verified as dominated (\cref{sec:s-ivl}).
  $^{d}$\ListAvg: $\Eone$ over all four rotations. $\Delta$Human: winner agreement with the HPDv2 ranking on 1{,}200 three-image sets, \PWC minus \List.
  \end{minipage}
\end{table*}

\subsection{One added image}
\label{sec:s-k1}

\Cref{tab:s-k1} gives the three $k=1$ rates that \cref{tab:s-judges} prints as two columns and a footnote, plus a natural image on the two Qwen judges.
On every judge one added image of any kind moves \List's decoy-free winner in 15 to 28\% of the sets.
A searched decoy and a re-render of the fixed decoy are indistinguishable at $k=1$ (differences $-0.9$ to $+0.6$ points, every interval containing 0), and a content-free grey image moves the winner almost as often (decoy minus grey $+0.7$ to $+4.4$).
The finding at $k=1$ is therefore that any added sibling moves the winner; dominance is what makes the movement a violation, and the search is what separates decoys from re-renders.
With the whole pool as the budget, the genuine-addition rates, which do not count the added image itself winning, separate: decoys against re-renders 45.0 against 30.8 on q8 (paired $+14.2$ [$+10.4$, $+18.3$]), 43.1 against 30.4 on q32 ($+12.7$ [$+8.4$, $+17.4$]), 36.9 against 24.8 on ivl38 ($+12.1$ [$+8.1$, $+16.8$]), 46.3 against 32.6 on mst24 ($+13.8$ [$+9.7$, $+18.1$]) and 49.8 against 32.6 on ur7 ($+17.2$ [$+12.1$, $+22.2$]).

\begin{table*}[t]
  \centering
  \scriptsize
  \setlength{\tabcolsep}{4pt}
  \caption{\textbf{One added image, \List.} Share of sets (\%) in which one added candidate in the fourth slot changes the winner of the decoy-free triple to another contender (a genuine addition; \cref{tab:s-notation}), averaged over the set's added candidates of each kind. Intervals are 95\% except the differences marked $^\dagger$, which are 97.5\%; the ivl38, mst24 and ur7 rows are recomputed from stored rows.}
  \label{tab:s-k1}
  \begin{tabular}{@{}lrllllll@{}}
    \toprule
    judge & sets & one searched decoy & one re-render & one grey image & one natural image & decoy $-$ re-render & decoy $-$ grey \\
    \midrule
    q8 & 289 & 26.8 [22.3, 31.3] & 26.2 [21.3, 31.2] & 22.4 [17.8, 27.2] & 27.9 [23.4, 32.5] & $+0.6$ [$-1.5$, $+2.7$] & $+4.4$ [$+1.7$, $+7.1$] \\
    q32 & 299 & 22.0 [18.1, 26.1] & 22.4 [18.1, 26.8] & 20.6 [16.5, 24.8] & 23.7 [19.6, 27.8] & $-0.4$ [$-2.6$, $+1.9$]$^\dagger$ & $+1.4$ [$-2.4$, $+5.0$]$^\dagger$ \\
    ivl38 & 298 & 18.6 [14.8, 22.6] & 18.9 [14.8, 23.2] & 14.7 [10.9, 18.6] & -- & $-0.3$ [$-2.4$, $+1.8$]$^\dagger$ & $+3.9$ [$+1.4$, $+6.6$]$^\dagger$ \\
    mst24 & 298 & 20.6 [16.9, 24.3] & 21.3 [17.0, 25.6] & 19.9 [15.8, 24.1] & -- & $-0.6$ [$-2.8$, $+1.5$] & $+0.7$ [$-1.3$, $+2.7$] \\
    ur7 & 239 & 23.6 [19.2, 28.2] & 24.5 [19.5, 29.7] & 19.6 [15.0, 24.6] & -- & $-0.9$ [$-3.5$, $+1.6$] & $+3.9$ [$+0.0$, $+7.8$] \\
    \bottomrule
  \end{tabular}
\end{table*}

\paragraph{What one added image does to the GRPO ranking.}
A listwise reward uses the whole ranking, not only the top pick.
\Cref{tab:s-grpo-rank} reads, from the stored rows, how often one added candidate changes \List's top pick among the contenders, their full order, and the sign of a GRPO advantage computed from the ranks.
With one added decoy the order of the three contenders changes in 44.5\% of q8 sets and the sign of some advantage in 24.9\%; with the whole pool, in 75.1\% and 50.2\%.
Grey images and re-renders do nearly as much at $k=1$.
\PWC and \Rating change none of these quantities at $k=1$ and at most 1.4\% with the whole pool.

\begin{table*}[t]
  \centering
  \scriptsize
  \setlength{\tabcolsep}{5pt}
  \caption{\textbf{One added image and the \List ranking used by GRPO} (\%; stored rows; $k=1$ / any of the pool; 95\% where printed). Top-1: the top pick changes, the added image winning included, which the genuine-addition rates of \cref{tab:s-k1} exclude. Order: the contenders' full order changes. Sign: the sign of some contender's rank-based GRPO advantage changes.}
  \label{tab:s-grpo-rank}
  \begin{tabular}{@{}llccc@{}}
    \toprule
    judge & added & top-1 & order & sign \\
    \midrule
    q8 (289) & decoy & 26.8 [22.3, 31.4] / 45.7 & 44.5 [39.8, 49.3] / 75.1 & 24.9 [20.8, 29.2] / 50.2 \\
    & grey & 22.4 / 25.3 & 40.7 / 46.4 & 24.1 / 28.4 \\
    & re-render & 26.2 / 30.8 & 44.2 / 54.0 & 24.9 / 32.9 \\
    q32 (299) & decoy & 22.1 [18.2, 26.2] / 44.1 & 40.4 / 71.2 & 25.0 [21.0, 29.2] / 48.5 \\
    & grey & 20.6 / 28.8 & 44.0 / 55.9 & 29.2 / 36.8 \\
    & re-render & 22.4 / 30.4 & 41.6 / 54.5 & 26.1 / 36.1 \\
    ivl38 (298) & decoy & 18.6 / 37.6 & 31.2 / 62.1 & 17.3 / 41.3 \\
    \midrule
    q8, q32 & \PWC or \Rating & 0.0 / $\le$1.4 & 0.0 / $\le$0.3 & 0.0 / $\le$0.3 \\
    \bottomrule
  \end{tabular}
\end{table*}

\subsection{Searched excess, budget curves and the addition}
\label{sec:s-budget}

\Cref{tab:s-budget} gives the per-judge searched rate, floor and excess behind \mainref{tab:judges}, and \cref{tab:s-matched} the matched budgets between one try and the whole pool.
On q32, q8 and ur7 the excess of the swap against the modal row grows with the budget: at one decoy against one re-render it is 4.3--6.1 points (as a genuine addition, \cref{tab:s-k1}, the difference is within $\pm1$), at any of three 9.2--13.3, and over the whole pool 22.4--31.1.
Counting only flips to another contender changes little (q32 $+21.1$, q8 $+29.8$), because the decoy itself almost never wins under \List.

\begin{table}[t]
  \centering
  \scriptsize
  \setlength{\tabcolsep}{2pt}
  \caption{\textbf{Searched rate, floor and excess per judge} (\List; any of $n$; \% of sets; 95\%, the level of the main paper).}
  \label{tab:s-budget}
  \begin{tabular}{@{}lrlll@{}}
    \toprule
    judge & sets & searched & floor & excess $\Eone$ \\
    \midrule
    q32 & 299 & 39.5 [33.8, 44.8] & 17.1 [12.7, 21.4] & 22.4 [17.4, 27.4] \\
    q8 & 289 & 40.1 [34.6, 46.0] & 9.0 [5.9, 12.5] & 31.1 [26.0, 36.7] \\
    ivl38 & 298 & 33.6 [28.2, 39.3] & 10.1 [6.7, 13.8] & 23.5 [18.8, 28.5] \\
    mst24 & 298 & 45.3 [39.6, 51.0] & 18.1 [13.8, 22.5] & 27.2 [21.8, 32.6] \\
    ur7 & 239 & 46.4 [40.2, 52.7] & 17.6 [13.0, 22.6] & 28.9 [23.0, 34.7] \\
    \midrule
    q32, held-out & 100 & -- & -- & 20.0 [12.0, 28.0] \\
    q8, text decoys & 486 & 45.9 [41.4, 50.2] & 9.7 [7.2, 12.3] & 36.2 [31.9, 40.5] \\
    \bottomrule
  \end{tabular}
\end{table}

\begin{table*}[t]
  \centering
  \scriptsize
  \setlength{\tabcolsep}{4pt}
  \caption{\textbf{Matched budgets} (\List; \% of sets; 95\%). Decoys against re-renders at the same budget $k$ by \cref{eq:anyk}, and the contender-only excess over the whole pool.}
  \label{tab:s-matched}
  \begin{tabular}{@{}llll@{}}
    \toprule
    judge & $k=1$: one decoy vs one re-render (difference) & any of 3 (difference) & contender-only $\Eone$ \\
    \midrule
    q32 & 7.1 [5.6, 8.6] vs 2.7 [1.9, 3.7] ($+4.3$ [$+2.9$, $+5.8$]) & 15.6 [12.7, 18.6] vs 6.4 [4.5, 8.5] ($+9.2$ [$+6.5$, $+12.0$]) & $+21.1$ [$+16.4$, $+26.1$] \\
    q8 & 7.2 [5.8, 8.7] vs 2.0 [1.2, 3.0] ($+5.2$ [$+3.9$, $+6.6$]) & 16.0 [13.2, 19.0] vs 4.4 [2.6, 6.4] ($+11.6$ [$+8.9$, $+14.4$]) & $+29.8$ [$+24.6$, $+35.3$] \\
    ur7 & 9.3 [7.6, 11.2] vs 3.2 [2.1, 4.5] ($+6.1$ [$+4.3$, $+8.0$]) & 20.4 [17.0, 24.1] vs 7.1 [4.7, 9.8] ($+13.3$ [$+9.7$, $+17.0$]) & $+29.3$ [$+23.4$, $+35.1$]$^\ast$ \\
    \bottomrule
  \end{tabular}
  \par\smallskip\parbox{\textwidth}{\scriptsize $^\ast$ Before ur7's final dominance closure; its closed whole-pool excess is $+28.9$ (\cref{tab:s-budget}). The $k=1$ column here is the swap against the modal row; \cref{tab:s-k1} is the addition against the decoy-free triple.}
\end{table*}

\paragraph{Swap and addition.}
The main estimator compares each decoy's winner with the modal winner over the pool (a swap); the addition compares it with the decoy-free triple.
Counted as in \cref{tab:s-grpo-rank}, they agree: on q32 the mean addition rate is 22.1 [18.2, 26.2] per decoy against 22.4 [18.1, 26.8] per re-render (difference $-0.3$ [$-2.2$, $+1.6$]) and 44.1 [38.5, 49.8] against 30.4 [25.4, 35.5] over the whole pool ($+13.7$ [$+9.7$, $+17.7$]); on q8, 26.8 against 26.2 ($+0.6$ [$-1.5$, $+2.7$]) and 45.7 [40.1, 51.2] against 30.8 [25.6, 36.3] ($+14.9$ [$+10.7$, $+19.4$]).
The modal winner over the pool differs from the decoy-free winner in 21.7 [17.1, 26.4]\% of q32 sets.

\paragraph{Targeted flips.}
Making a fixed, seeded contender other than the decoy-free winner win is harder and grows with the budget: decoys minus re-renders at $k=1$, 3, 10 and 50 are $+1.3$ [$-0.3$, $+3.0$], $+4.6$ [$+2.5$, $+6.7$], $+7.0$ [$+4.5$, $+9.5$] and $+9.7$ [$+6.6$, $+13.1$] points on q8, and $+0.1$ [$-1.5$, $+1.7$], $+2.7$ [$+0.7$, $+4.8$], $+5.4$ [$+3.0$, $+8.0$] and $+9.0$ [$+5.7$, $+12.4$] on q32.

\paragraph{By slot.}
With the decoy in slot 1, 2, 3 or 4, q32's searched excess is $+28.0$ [$+17.3$, $+38.7$] ($n=75$), $+28.4$ [$+18.9$, $+39.2$] (74), $+13.3$ [$+5.3$, $+22.7$] (75) and $+20.0$ [$+10.7$, $+29.3$] (75); q8's is $+31.9$, $+31.0$, $+25.7$ and $+36.1$; ur7's $+49.2$, $+35.0$, $+17.2$ and $+25.0$.
The effect is present in every slot.

\subsection{InternVL readability and UnifiedReward coverage}
\label{sec:s-ivl}

\begin{figure*}[t]
  \centering
  \figfile{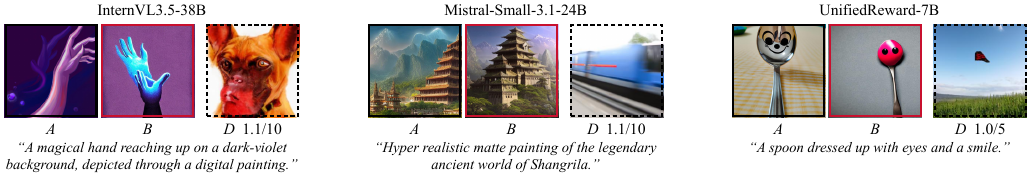}{\textwidth}{5cm}{figS\_c\_v11.pdf: three examples on three other judges}
  \vspace{-1.4em}
  \caption{\textbf{One added image on three more judge families.}
  InternVL3.5-38B, Mistral-Small-3.1-24B and UnifiedReward-7B under \List: $D$ (dashed) is the added image, with the judge's rating of it alone; black frame, the pick without $D$; red, the pick with it.}
  \label{fig:s-cards}
\end{figure*}

\Cref{fig:s-cards} shows one added image on ivl38, mst24 and ur7.
ivl38 abstains often: 54.08\% of pool readouts and 52.38\% of re-render readouts have a first-token label mass below 0.5 (12.42\% of decoy-free readouts); the primary readout reads them through the answer stem (\cref{sec:s-readouts}).
The main paper prints the full-set rates, in which every set counts: searched 33.6 [28.2, 39.3], floor 10.1 [6.7, 13.8], excess $+23.5$ [$+18.8$, $+28.5$].
Counting first-token abstentions as unreadable instead leaves 180 sets with a median effective pool of 42 and 53.26\% of 29{,}685 search readouts excluded; on them the searched rate is 28.9 [22.8, 35.6], the floor 9.4 [5.6, 13.9] and the excess $+19.4$ [$+13.3$, $+25.6$].
On the 37 sets whose decoy-free row abstained the excess is $+48.6$ [$+32.4$, $+64.9$], on the other 261 $+19.9$ [$+14.9$, $+24.9$].
For ur7, 61 of 300 sets have no verified decoy; before its final dominance closure the excess read $+31.8$ [$+25.1$, $+38.9$] (97.5\%) with unchecked flips outstanding, and after it $+28.9$ [$+23.0$, $+34.7$] with none.

\subsection{Transfer between open judges}
\label{sec:s-transfer}

q8's strongest decoys, checked on q32 and counted only where they enter q32's own screened pool, transfer: on 287 sets with 822 qualified decoys (mean $k$ 2.86), any of $k$ transferred decoys flips q32 in 24.0 [19.2, 29.3]\% of sets against 15.2 [12.3, 18.2]\% for $k$ of q32's random decoys ($+8.8$ [$+5.1$, $+12.8$]) and 6.1 [4.2, 8.2]\% for $k$ re-renders ($+18.0$ [$+13.1$, $+22.9$]).
Per decoy, 12.1\% against 7.1\% ($+5.1$ [$+2.8$, $+7.4$]).
Transfer was measured between open-weight judges only; the closed-model check is in \cref{sec:s-closed}.

\subsection{Text decoys}
\label{sec:s-text}

On VL-RewardBench, q8 ranks $N=3$ text responses to an image, and text decoys are searched exactly as image decoys (\cref{sec:s-secondary-designs}).
On 486 items the searched rate is $+45.9$ [$+41.4$, $+50.2$], the floor of fifty re-renders $+9.7$ [$+7.2$, $+12.3$], and the excess $+36.2$ [$+31.9$, $+40.5$]; the contender-only excess is $+36.0$ [$+31.7$, $+40.3$] and the decoy itself wins 0.1 [0.0, 0.2]\%.
At $k=1$ one random decoy flips 9.1 [7.8, 10.4]\% against 2.5 [1.7, 3.3]\% for one re-render ($+6.6$ [$+5.4$, $+7.9$]); at any of 3, 20.2 [17.5, 22.7] against 5.3 [3.8, 7.1] ($+14.9$ [$+12.4$, $+17.3$]).
The dominance check covered 6{,}680 decoys (252 failed); no readout was unreadable and no item incomplete.
This family has one judge and is reported as such in \cref{tab:s-judges}.

\subsection{Natural and content-free images, native answers, far contenders, larger sets}
\label{sec:s-natural}

\paragraph{Natural images.}
Searched decoys are not special.
Fifty unscreened natural images per set move \List about as often as fifty dominated decoys (\cref{tab:s-natural}): flips to another contender 40.3\% on q8 and 42.4\% on q32, 31.3 and 25.3 points above the re-render floor, and decoys minus natural images $-1.5$ [$-5.7$, $+2.7$] and $-4.2$ [$-8.3$, $-0.2$].
The natural image itself wins in 0.2\% and 0.1\% of single messages.
What the dominance check adds is that the movement is a violation of the property, not that the decoys are stronger.
Counting any flip instead of contender flips gives natural minus re-render $+32.4$ [$+26.9$, $+37.9$] (q8) and $+27.0$ [$+21.6$, $+32.4$] (q32), and decoys minus natural $-1.2$ [$-5.8$, $+3.2$] and $-4.6$ [$-8.9$, $-0.2$].

\begin{table*}[t]
  \centering
  \scriptsize
  \setlength{\tabcolsep}{5pt}
  \caption{\textbf{Fifty natural images per set} (\List; \%; 95\%). Contender flip: the winner moves to another contender (any of the fifty). Own win: the natural image wins one message. Differences use contender flips. Last column: one natural image against one grey image as a genuine addition ($k=1$).}
  \label{tab:s-natural}
  \begin{tabular}{@{}lrllllll@{}}
    \toprule
    judge & sets & any flip & contender flip & own win & natural $-$ re-render & decoys $-$ natural & $k=1$: natural $-$ grey \\
    \midrule
    q8 & 289 & 41.4 [35.8, 47.1] & 40.3 [34.8, 45.9] & 0.2 [0.1, 0.4] & $+31.3$ [$+25.9$, $+36.9$] & $-1.5$ [$-5.7$, $+2.7$] & $+5.5$ [$+2.6$, $+8.4$] \\
    q32 & 299 & 44.0 [38.4, 49.7] & 42.4 [36.8, 48.0] & 0.1 [0.0, 0.1] & $+25.3$ [$+20.1$, $+30.7$] & $-4.2$ [$-8.3$, $-0.2$] & $+3.0$ [$-0.4$, $+6.5$] \\
    \bottomrule
  \end{tabular}
\end{table*}

\paragraph{Content-free images.}
Distinct content-free images searched through the same budget flip q8 in 21.1 [16.6, 26.0]\% of sets and q32 in 30.8 [25.4, 36.1]\%, against re-render floors of 9.0 [5.9, 12.5] and 17.1 [12.7, 21.4]: 12.1 [7.3, 17.3] and 13.7 [8.4, 19.1] points above the floor, and 19.0 [14.2, 24.2] and 8.7 [3.3, 14.0] points below the decoys.
Any added content moves \List; content-free images move it less than natural images or searched decoys.

\paragraph{Native answers.}
Reading q32's free-form greedy reply instead of its first-token distribution leaves the effect: the best decoy flips the reference winner in 31.8 [26.4, 37.1]\% of sets against 11.0 [7.4, 14.7]\% for the best re-render (excess $+20.7$ [$+15.7$, $+25.8$]); at $k=1$, decoys 22.3 [18.4, 26.5] against grey images 20.9 [16.8, 25.1] ($+1.4$ [$-1.9$, $+4.7$]).
All 7{,}176 replies parsed, and the reference reply agrees with the forced answer in 98.7 [97.3, 99.7]\% of sets.

\paragraph{Far contenders.}
With contenders of human ranks 1, 5 and 9, the decisions are easy and the excess shrinks to $+8.8$ [$+5.7$, $+12.2$] on q32 ($n=296$; searched 13.9 [10.1, 17.9], floor 5.1 [2.7, 7.8]): the sensitivity is a close-call effect.

\paragraph{Larger sets.}
With seven contenders and the decoy in the eighth slot, q8's excess is $+24.9$ [$+20.1$, $+30.1$] ($n=289$; searched 36.7 [31.5, 42.2], floor 11.8 [8.3, 15.6]).

\subsection{Closed-model check}
\label{sec:s-closed}

Our claims concern open-weight judges with logit readouts.
On the one closed judge we could test, Claude Opus~5.5, through a blinded agent route (120 sets, any of 3, 1{,}740 answers, none unreadable), q8's strongest decoys did not transfer (\cref{tab:s-closed}): transfer minus random $+3.3$ [$-2.5$, $+9.2$], transfer minus re-render $-4.2$ [$-10.8$, $+1.7$], transfer minus re-query $-1.7$ [$-9.7$, $+7.0$].
The comparison fixed before any answer was not met.
Opus picked the decoy itself in 0 of the 1{,}740 answers.
Its sampled answers are unstable on their own (re-query 18.3\%), which bounds what an any-of-3 check can resolve.
This is a statement about black-box transfer of q8's decoys to one closed judge at the probed scale, not evidence that closed judges are insensitive; a logit-level search on a closed judge is not possible through its interface and was not run.

\begin{table}[t]
  \centering
  \scriptsize
  \setlength{\tabcolsep}{3pt}
  \caption{\textbf{Closed-model transfer check} (Claude Opus~5.5; 120 development sets; any of 3; 95\%).}
  \label{tab:s-closed}
  \begin{tabular}{@{}lll@{}}
    \toprule
    arm & sets flipped (\%) & per query (\%) \\
    \midrule
    transfer (q8's strongest decoys) & 15.8 [10.0, 22.5] & 6.7 [3.9, 9.7] \\
    random pool decoy & 12.5 [6.7, 19.2] & 5.6 [3.1, 8.6] \\
    re-render of the fixed decoy & 20.0 [13.3, 27.5] & 9.2 [5.6, 13.3] \\
    re-query (60 sets) & 18.3 [10.0, 27.2] & 8.0 [4.0, 12.3] \\
    \bottomrule
  \end{tabular}
\end{table}

\subsection{Order averaging and label-prior debiasing}
\label{sec:s-listavg}

\ListAvg judges every message in all four cyclic rotations (\cref{sec:s-floors}).
On q32 its excess is $+24.7$ [$+19.7$, $+29.4$] against \List's $+22.4$ [$+17.4$, $+27.4$] ($R=1.10$ [0.84, 1.44]): averaging lowers the floor from 17.1 [12.7, 21.4] to 13.7 [10.0, 17.7] but leaves the searched rate at 38.5 [33.1, 43.8] against 39.5 [33.8, 44.8].
On q8 the excess is $+26.3$ [$+21.1$, $+31.5$] against $+31.1$ [$+26.0$, $+36.7$] ($R=0.84$ [0.66, 1.06]), with residual 37.0\% and the decoy winning 0.1\%.
Permutation self-consistency~\cite{tang2024found}, which aggregates the rankings of all rotations, likewise keeps an excess on both judges, 27.0 on q8 and 22.4 on q32 (four rotations per decoy set).
Descriptively, the averaged winner differs from the decoy-free winner for 30.7\% (q8) and 22.2\% (q32) of the stored decoys, and at $N=8$ on the audit groups one added decoy still changes \ListAvg's top pick in 18.5\% of groups (\cref{sec:s-audit}).
Averaging also changes agreement with the human winner on the decoy-free triples of the attack sets, paired with \List: $+1.4$ [$-3.5$, $+6.2$] on q8 and $+4.0$ [$+0.0$, $+8.4$] on q32.

\PriDe~\cite{zheng2024mcq} removes an estimated label prior from a single \List message (prior estimated on held-out folds, four-fold).
It leaves the excess at $+30.8$ [$+25.6$, $+36.0$] on q8 and $+24.4$ [$+19.4$, $+29.8$] on q32.
Removing the order noise or the estimated label prior thus leaves most of the candidate-set sensitivity; its mechanism is not resolved by either.

\subsection{Pairwise readouts under the attack}
\label{sec:s-pairwise-attack}

Every pairwise readout we tested shows a small searched excess, mostly the added image itself winning.
On q8 and q32 the excess is 1.4 [0.3, 2.8] and 1.3 [0.3, 2.7] for \PWC, 2.8 and 1.0 for \PairOne, and 3.8 and 2.3 for \PRP with its paper tie rule (0.7 and 0.5 with a uniform tie-break among tied top scores).
\PWC's residual, the flips to another contender, is 0.0 [0.0, 0.0] on q8 and 0.3 [0.0, 1.0] on q32 (one set of 299), and its decoy wins 1.4 and 1.0 any of $n$.
Under the pair winners the stored decoy is a universal loser in 99.6\% of the attack messages, which is why aggregations without a cover step leave little residual here (\cref{tab:s-rules}).
\PRP's top score is tied in 28--40\% of pool messages, because the first-shown image wins both orders in 45\% (q8) and 29\% (q32) of contender pairs, and \PRP counts such a pair as a tie; most of its excess comes from how the resulting top-score ties are broken.
\PWC's rows on the other judges are in \cref{sec:s-variants}: ivl38 0.0 [0.0, 0.0], mst24 0.3 [0.0, 1.0], ur7 0.4 [0.0, 1.3], each with residual 0.0.
A pair message never shows a third image and so holds each comparison's inputs fixed; the searched excess left to the pairwise readouts (\mainref{sec:measure-diag}) is the added image itself winning under \PWC and the slot tie-break under \PRP.

\subsection{Defences that failed}
\label{sec:s-defences}

Every defence runs on q8 with the same 289 sets and candidates, and the decoy pool is re-searched and re-closed under each, because a defence changes which decoys flip (\cref{tab:s-defences}).
\begin{itemize}[leftmargin=*,itemsep=1pt,topsep=2pt]
\item \emph{Order averaging} is \ListAvg (\cref{sec:s-listavg}).
\item \emph{Delimiter token scaling}~\cite{lee2026enhancing} scales the hidden states of the image-delimiter tokens at the method's released example setting, in one listwise pass.
\item \emph{Sparse intra-image mask}, after BlindSight~\cite{srikrishnan2025blindsight}: image tokens attend only to the text before the first image and to their own image in the inner language-model layers; the readout is unmasked. This is an efficiency method's mask used as a defence, not a defence its authors proposed.
\item \emph{FOCUS}~\cite{park2026mitigating}: for each candidate, one pass in which every other image is replaced by FOCUS's noise at its strongest masking ($\lambda=1$), plus a reference pass, and the arg\,max of the contrasted scores. At this setting the noise carries no image content, so a contender's score never depends on the decoy, and every FOCUS flip is a win of the decoy itself.
\item \emph{\PriDe}~\cite{zheng2024mcq}: a single \List message debiased by an estimated label prior (\cref{sec:s-listavg}).
\item \emph{\ScreenList}: every candidate is rated alone with \Rating, candidates rated well below the second best are dropped, and \List judges the survivors in their original relative order.
\end{itemize}

\begin{table*}[t]
  \centering
  \scriptsize
  \setlength{\tabcolsep}{4pt}
  \caption{\textbf{Defences} (development sets, q8 $n=289$, q32 $n=299$; \% of sets; 95\%). $R$: excess relative to \List on the same resamples. Residual: flips to another contender. Agreement: agreement with the human winner on the decoy-free triples, paired with \List. Decoy wins: share of a set's decoys that win, averaged over sets (\cref{tab:s-fix} counts the sets in which some decoy wins).}
  \label{tab:s-defences}
  \begin{tabular}{@{}llllllll@{}}
    \toprule
    defence & judge & excess $\Eone$ & $R$ & residual & decoy wins & agreement $-$ \List \\
    \midrule
    \List (reference) & q8 & 31.1 [26.0, 36.7] & -- & 38.8 [33.2, 44.6] & 0.1 & -- \\
    \ListAvg, four rotations & q8 & 26.3 [21.1, 31.5] & 0.84 [0.66, 1.06] & 37.0 & 0.1 & $+1.4$ [$-3.5$, $+6.2$] \\
    & q32 & 24.7 [19.7, 29.4] & 1.10 [0.84, 1.44] & -- & -- & $+4.0$ [$+0.0$, $+8.4$] \\
    delimiter token scaling & q8 & 30.8 [25.6, 36.0] & 0.99 [0.83, 1.19] & 41.2 & 0.1 & $-2.4$ [$-5.9$, $+1.0$] \\
    sparse intra-image mask & q8 & 28.0 [22.8, 33.6] & 0.90 [0.71, 1.12] & 35.3 & 0.1 & $-11.4$ [$-17.0$, $-5.5$] \\
    FOCUS$^\ast$ & q8 & 0.3 [0.0, 1.0] & 0.01 [0.00, 0.04] & 0.3 & 25.6 [20.7, 30.8] & $-25.6$ [$-34.3$, $-17.0$] \\
    \PriDe, one message & q8 & 30.8 [25.6, 36.0] & -- & -- & -- & -- \\
    & q32 & 24.4 [19.4, 29.8] & -- & -- & -- & -- \\
    \ScreenList$^\ddagger$ & q8 & 3.8 [1.7, 6.2] & 0.12 [0.06, 0.20] & 2.4 [0.7, 4.5] & -- & -- \\
    & q32 & 5.4 [3.0, 8.0] & 0.23 [0.12, 0.36] & 4.3 [2.3, 6.7] & -- & -- \\
    \Rating & q8 & 0.3 [0.0, 1.0] & 0.01 & 0.0 & -- & -- \\
    & q32 & 0.3 [0.0, 1.0] & 0.01 & 0.0 & -- & -- \\
    \bottomrule
  \end{tabular}
  \par\smallskip\parbox{\textwidth}{\scriptsize $^\ast$ The decoy itself wins instead; one certificate problem in 28{,}827 FOCUS rows, so the row is reported as incomplete. $^\ddagger$ Fails on adaptive decoys (below).}
\end{table*}

Order averaging, delimiter scaling and the sparse mask leave most of the excess on q8 ($R$ 0.84--0.99), and so does \PriDe ($+30.8$ against $+31.1$).
FOCUS removes the flips to another contender only by letting the decoy itself win, in 25.6\% of the sets, and loses 25.6 points of agreement with the human winners.
\ScreenList removes most of the excess on screened decoys, because the screen drops them before the listwise message; an adaptive attacker defeats it.
Under all these controls, the sensitivity stays with the multi-candidate message.

\paragraph{Adaptive decoys.}
On the 58 q8 sets with an adaptive decoy, rated within 1.0 of $A$ and $B$ (\cref{sec:s-secondary-designs}), \List's excess is $+8.6$ [$+1.7$, $+17.2$] (searched 12.1, floor 3.4), and \ScreenList's is identical, $+8.6$: every adaptive decoy survives its screen, against 1\% of the standard pool; \Rating's is 0.0.
\ScreenList's row is incomplete (7.32\% of its readouts unreadable).
\PWC on the same band has excess 0.0 [0.0, 0.0], residual 0.0 and decoy wins 3.4 [0.0, 8.6]; the closure removed 80 decoys and no \PWC flip.
The band was searched against \Rating and \List, never against \PWC, so it does not test an attacker adapted to \PWC.

\subsection{Label permutation, isolation and shared positions}
\label{sec:s-iso}

\begin{figure}[t]
  \centering
  \figfile{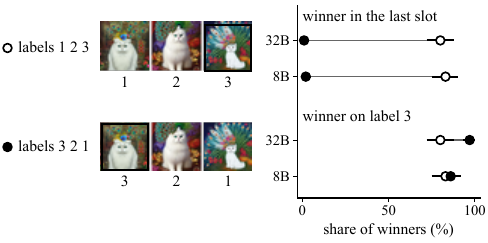}{\linewidth}{4.5cm}{fig2\_v9.pdf: left, label-permutation schematic; right, last-slot and label-3 winner shares under labels 1 2 3 and 3 2 1}
  \vspace{-1.5em}
  \caption{\textbf{The preference follows the label digit, not the slot.}
  Label-permutation control, 100 development sets per judge, isolated pairwise diagnostic readout \IsoPair.
  \emph{Left:}~the digits are reversed while every image keeps its slot; the pick (black frame) follows label~3.
  \emph{Right:}~share of decoy-free winners in the last slot and on label~3 under labels 1\,2\,3 ($\circ$) and 3\,2\,1 ($\bullet$), 95\% intervals; under 1\,2\,3 the two coincide.}
  \label{fig:s-lp}
\end{figure}

\paragraph{Label permutation.}
On 100 development sets per judge, each message is judged with the identity labelling, with the labels reversed, and with a random relabelling, so that slot and label digit separate; the images and their positions are unchanged and the swap is checked bitwise.
\Cref{tab:s-lp} decomposes the diagnostic pair readout \IsoPair into a label term, a slot term and a mention-order term (log-odds, nats).
For \IsoPair the label term dominates on both judges, so its preference follows the label digit, not the position (\cref{fig:s-lp}): under reversed labels its last-slot winner share drops from 83\% to 2\% on q8 and from 80\% to 1\% on q32, while the share of the top-label image stays at 86\% and 97\%.
For \IsoList the winner is label-driven on q32 ($\lambda$ 1.06 [1.02, 1.12]) and mixed on q8 ($\lambda$ 0.60 [0.49, 0.72]).
For \List itself the last-slot winner share falls from 45\% to 19\% (q8) and from 46\% to 24\% (q32) under reversed labels, while the top-label share rises to 66\% and 64\%.
$\lambda$ is the share of the prior that follows the label; both \IsoPair values exceed 1.1, which means the prior reversed by more than it was, so the label and slot components, not the share, are the quantities to read.
A yes/no readout without digits shows no last-slot preference (winner share by slot q8 37 / 35 / 28, q32 48 / 25 / 27).
Discrimination is unchanged within its interval under reversed or random labels (e.g.\ \IsoPair, reversed: q8 $-1$ [$-16$, $+15$], q32 $+8$ [$-8$, $+24$]).
Swap certificates hold bitwise in every arm, every readout parsed, and the identity rows equal the stored rows in 300 of 300 messages.

\begin{table*}[t]
  \centering
  \scriptsize
  \setlength{\tabcolsep}{6pt}
  \caption{\textbf{Label-permutation control} (100 development sets per judge; 95\%). Terms of the diagnostic pair readout \IsoPair in nats; $\lambda$: share of its prior that follows the label digit; winner shares and pair rates in \%, identity labels $\to$ reversed labels; \IsoList's $\lambda$ and \List's shares for comparison.}
  \label{tab:s-lp}
  \begin{tabular}{@{}lll@{}}
    \toprule
    quantity & q8 & q32 \\
    \midrule
    \IsoPair: $\lambda$ & 1.12 [1.03, 1.26] & 1.11 [1.05, 1.19] \\
    \quad label term & $-2.24$ [$-2.41$, $-2.08$] & $-3.85$ [$-4.07$, $-3.62$] \\
    \quad slot term & $+0.15$ [$-0.12$, $+0.44$] & $+0.86$ [$+0.58$, $+1.14$] \\
    \quad mention term & $-2.37$ [$-2.52$, $-2.22$] & $-1.78$ [$-1.87$, $-1.69$] \\
    \quad last-slot winners & 83 [75, 90] $\to$ 2 [0, 5] & 80 [72, 88] $\to$ 1 [0, 3] \\
    \quad top-label winners & 83 $\to$ 86 [79, 92] & 80 $\to$ 97 [93, 100] \\
    \quad lower slot wins a contender pair & 17.3 [12.3, 23.0] $\to$ 90.7 [87.3, 93.7] & 12.7 $\to$ 95.7 \\
    \IsoList: $\lambda$ & 0.60 [0.49, 0.72] (mixed) & 1.06 [1.02, 1.12] \\
    \List: last-slot winners; top-label winners & 45 $\to$ 19; 45 $\to$ 66 [57, 75] & 46 $\to$ 24; 46 $\to$ 64 \\
    \bottomrule
  \end{tabular}
\end{table*}

\paragraph{Slot statistics, by data source.}
On the development sets with no decoy, \List's winner is in the last of three slots in 48.8 [43.3, 54.7]\% of q8 sets and 45.5 [39.8, 51.2]\% of q32 sets (all development sets, 289 and 299; 45\% and 46\% on the 100 label-permutation sets).
On the 128 audit groups of eight rollouts, single \List messages put the top pick in slots 1--8 at 8.5, 2.6, 5.1, 6.1, 9.9, 12.6, 23.5 and 31.7\% (chance 12.5; 2{,}048 messages).
In the pilot training groups of the reward-path review (\cref{sec:s-pilot}), slots 7 and 8 won 59\% of groups against 25\% expected, and slot position accounted for about 29\% of \List's reward variance.
These come from different data and are not combined.

\paragraph{Isolation.}
Encoding the images in isolation does not remove the effect (\cref{tab:s-iso}): isolating each image's encoding with InvariRank's block attention mask~\cite{bito2026one} keeps 110\% of \List's excess on q8 and 191\% on q32.
This is \IsoList; the forward-tail pair readout keeps 46\% on both judges; the yes/no readout and the per-candidate readout, which never compare labels, are near zero.
Every isolated readout loses agreement with the human winner against \List (top-3 agreement \IsoList $-15.6$ [$-21.8$, $-9.3$] on q8 and $-15.4$ [$-21.7$, $-8.7$] on q32; forward-tail pairs $-17.0$ and $-14.7$), while the yes/no and per-candidate readouts lose less ($-3.5$ to $-5.4$, every interval containing 0).
The isolation certificates held: the contender margins were bitwise identical across every decoy and re-render of a set under a swap (0 varying sets of 289 and 299).
\IsoPair, which sums both mention orders of the isolated pair readout, has excess 9.7 [6.2, 13.5], 8.4 [5.4, 11.7] and 4.7 [2.3, 7.4] on q8, q32 and ivl38, residual 8.0 [4.8, 11.4], 0.7 [0.0, 1.7] and 3.7 [1.7, 6.0], and loses 14.2, 14.0 and 9.1 points of top-3 agreement; on K3+ it trails \List by 6.1 and 6.2 points of winner agreement (8.0 on the held-out sets).
Isolation therefore removes neither the label prior nor the set dependence, and costs accuracy.

\begin{table*}[t]
  \centering
  \scriptsize
  \setlength{\tabcolsep}{4pt}
  \caption{\textbf{Isolated readouts} (development sets; searched excess in \% of sets; 97.5\%). $R$: excess relative to \List on the same resamples. Contender flips: the forward-tail pair readout's excess counting only flips to another contender.}
  \label{tab:s-iso}
  \begin{tabular}{@{}lllllllll@{}}
    \toprule
    judge & \List & \IsoList & $R$ & pairs, fwd.\ tail & $R$ & contender flips & yes/no & per-cand. \\
    \midrule
    q8 (289) & 31.5 [25.6, 37.7] & 34.6 [28.4, 41.2] & 110\% [84, 144] & 14.5 [10.0, 19.4] & 46\% [31, 67] & 15.9 [11.1, 20.8] & 0.7 [0.0, 2.1] & 0.0 \\
    q32 (299) & 23.4 [17.7, 29.1] & 44.8 [38.5, 51.2] & 191\% [148, 255] & 10.7 [6.7, 15.1] & 46\% [27, 71] & 3.0 [1.0, 5.4] & 2.0 [0.3, 4.0] & 2.0 [0.3, 4.0] \\
    \bottomrule
  \end{tabular}
\end{table*}

\paragraph{Shared positions.}
Set-Based Prompting~\cite{mcilroyyoung2024orderindependence} gives parallel candidates the same positions.
We tested whether shared positions remove the slot preference of the isolated readouts, on 100 development sets per judge (\cref{tab:s-sp}).
They do not: the last-slot share of \IsoPair's decoy-free winners rises from 83\% to 92\% on q8 and from 80\% to 98\% on q32, \IsoList reaches 100\% on q32, the lower slot wins a contender pair less often, and the decoy itself starts to win (q32 \IsoList: 0.7 [0.3, 1.1]\% $\to$ 18.8 [12.0, 25.8]\% per decoy).
The yes/no readout without digits shows no comparable concentration (last slot 28 $\to$ 35\% on q8, 27 $\to$ 37\% on q32).
Sharing positions destroys the binding between a label and its image in label-referenced comparative readouts, so this control cannot say whether unshared positions cause the slot preference; the label-permutation control above, which keeps positions and permutes digits, is the informative one.
As a negative result in its own right, the Set-Based-Prompting remedy does not carry over to multi-image judges read through label digits.

\begin{table}[t]
  \centering
  \scriptsize
  \setlength{\tabcolsep}{2pt}
  \caption{\textbf{Shared-position control} (100 development sets per judge; \%; 95\%). Sequential $\to$ shared positions. Last-slot winners: share of decoy-free winners in the last slot (yes/no: share by slot). Pair: a contender pair in the four-image rows (\cref{tab:s-lp}: in the decoy-free triples).}
  \label{tab:s-sp}
  \begin{tabular}{@{}lll@{}}
    \toprule
    readout & last-slot winners & lower slot wins a pair \\
    \midrule
    q8 \IsoPair & 83 [75, 90] $\to$ 92 [86, 97] & 13.9 [9.5, 18.8] $\to$ 6.4 [3.4, 9.8] \\
    q8 pairs, forward tail & 89 $\to$ 96 [92, 99] & 7.1 $\to$ 2.0 \\
    q8 \IsoList & 82 $\to$ 54 [44, 64] & -- \\
    q8 yes/no (by slot) & 37 / 35 / 28 $\to$ 30 / 35 / 35 & -- \\
    q32 \IsoPair & 80 $\to$ 98 [95, 100] & 11.8 [7.5, 16.4] $\to$ 0.3 [0.0, 1.0] \\
    q32 pairs, forward tail & 89 $\to$ 100 & 7.7 [4.4, 11.4] $\to$ 0.0 \\
    q32 \IsoList & 92 $\to$ 100 & -- \\
    q32 yes/no (by slot) & 48 / 25 / 27 $\to$ 26 / 37 / 37 & -- \\
    \bottomrule
  \end{tabular}
\end{table}

\subsection{Variants of the fix and the held-out confirmation}
\label{sec:s-variants}

\paragraph{Variants.}
Before the readout of the main paper was chosen, several variants were compared on the development rows with one rule fixed in advance (\cref{sec:s-protocol}): residual at most 2.0 points with a 95\% upper bound at most 4.0 on q8 and q32, decoy wins at most 3.0 points, and a K3+ winner-agreement lower bound above $-1.0$ point against \List on q8, whose K3+ rows had not been looked at.
\GCC (gate, compare, cover) first drops every candidate whose per-candidate yes/no log-odds is at least 1.0 nat below the second highest, then runs \PWC on the survivors; \GCCR gates with \Rating instead; \GCC{}+$\lambda$ adds the gate log-odds difference to each pair margin.
\CHIP calibrates \IsoPair's summed margins by offsets estimated on the other half of the sets and adds the yes/no log-odds difference; \CalPair drops that term; \CHIP-P uses the per-candidate readout instead.
\GCC met the rule (residual 0.0 / 0.0, decoy wins 1.04 / 1.34, K3+ lower bound $-0.03$); \PWC met its attack-row conditions as well (residual 0.0 on q8 and 0.33 [upper 1.0] on q32, decoy wins 1.38 / 1.0); \CHIP was also eligible.
The held-out confirmation tested \PWC, the readout with no gate, and the main paper uses it; the gate's effect is a development-only observation.
\Cref{tab:s-fix} gives every attack row.

\begin{table*}[t]
  \centering
  \scriptsize
  \setlength{\tabcolsep}{4pt}
  \caption{\textbf{Variants of the fix on the attack rows} (development sets; \pp; 95\%; every closure complete). Top-3: agreement with the human winner on the decoy-free triples; $\Delta$: paired against \List. For mst24 and ur7 the \List reference is \cref{tab:s-budget}; \GCC was not run there and \GCCR has no ivl38 rows.}
  \label{tab:s-fix}
  \begin{tabular}{@{}llllll@{}}
    \toprule
    readout & excess $\Eone$ & residual & decoy wins & top-3 & $\Delta$ top-3 \\
    \midrule
    \multicolumn{6}{@{}l}{\emph{q8, 289 sets}} \\
    \PWC & 1.4 [0.3, 2.8] & 0.0 [0.0, 0.0] & 1.4 [0.3, 2.8] & 58.5 [52.6, 64.0] & $+1.7$ [$-3.8$, $+7.6$] \\
    \GCC & 1.0 [0.0, 2.4] & 0.0 [0.0, 0.0] & 1.0 [0.0, 2.4] & 57.1 [51.2, 63.0] & $+0.3$ [$-5.5$, $+6.2$] \\
    \GCC{}+$\lambda$ & 0.7 [0.0, 1.7] & 0.0 & 0.7 & 58.8 & $+2.1$ [$-4.2$, $+8.3$] \\
    \GCCR & 1.0 [0.0, 2.4] & 0.0 [0.0, 0.0] & 1.0 & 57.4 & $+0.7$ [$-4.8$, $+6.6$] \\
    \CHIP & 0.7 [0.0, 1.7] & 0.0 & 0.7 & 54.7 & $-2.1$ [$-8.7$, $+4.5$] \\
    \CalPair & 4.5 [2.4, 6.9] & 1.4 [0.3, 2.8] & 3.1 & 50.9 & $-5.9$ [$-12.5$, $+0.7$] \\
    \CHIP-P & 0.7 & 0.0 & 0.7 & 54.0 & $-2.8$ [$-9.0$, $+3.5$] \\
    \List (ref.) & 31.5 [26.3, 37.0] & 38.8 [33.6, 44.6] & 3.1 [1.4, 5.2] & 56.7 [50.9, 62.6] & -- \\
    \IsoPair (ref.) & 9.7 [6.2, 13.5] & 8.0 [4.8, 11.4] & 4.5 [2.4, 6.9] & 42.6 & $-14.2$ [$-20.1$, $-8.3$] \\
    \midrule
    \multicolumn{6}{@{}l}{\emph{q32, 299 sets}} \\
    \PWC & 1.3 [0.3, 2.7] & 0.3 [0.0, 1.0] & 1.0 [0.0, 2.3] & 53.2 [47.5, 58.5] & $+1.7$ [$-2.7$, $+6.0$] \\
    \GCC & 1.3 [0.3, 2.7] & 0.0 [0.0, 0.0] & 1.3 [0.3, 2.7] & 50.2 [44.5, 55.5] & $-1.3$ [$-6.4$, $+3.7$] \\
    \GCC{}+$\lambda$ & 1.0 [0.0, 2.3] & 0.0 & 1.0 & 49.8 & $-1.7$ [$-6.7$, $+3.3$] \\
    \GCCR & 1.3 [0.3, 2.7] & 0.3 [0.0, 1.0] & 1.0 & 52.2 & $+0.7$ [$-3.7$, $+5.0$] \\
    \CHIP & 2.3 [0.7, 4.3] & 0.0 & 2.3 & 44.8 & $-6.7$ [$-12.0$, $-1.0$] \\
    \CalPair & 7.4 [4.7, 10.4] & 2.3 [0.7, 4.0] & 6.4 & 47.8 & $-3.7$ [$-9.4$, $+2.0$] \\
    \CHIP-P & 3.3 [1.3, 5.7] & 0.7 [0.0, 1.7] & 2.7 & 46.8 & $-4.7$ [$-10.0$, $+1.0$] \\
    \List (ref.) & 23.4 [18.7, 28.4] & 38.8 [33.4, 44.5] & 3.0 [1.3, 5.0] & 51.5 [45.8, 57.2] & -- \\
    \IsoPair (ref.) & 8.4 [5.4, 11.7] & 0.7 [0.0, 1.7] & 8.0 [5.4, 11.4] & 37.5 & $-14.0$ [$-20.4$, $-7.7$] \\
    \midrule
    \multicolumn{6}{@{}l}{\emph{ivl38, 298 sets}} \\
    \PWC & 0.0 [0.0, 0.0] & 0.0 [0.0, 0.0] & 0.0 & 57.7 [52.0, 63.4] & $+3.7$ [$-0.7$, $+8.1$] \\
    \GCC & 0.0 [0.0, 0.0] & 0.0 [0.0, 0.0] & 0.0 [0.0, 0.0] & 57.7 [52.0, 63.4] & $+3.7$ [$-1.0$, $+8.4$] \\
    \GCC{}+$\lambda$ & 0.3 [0.0, 1.0] & 0.0 & 0.3 & 55.0 & $+1.0$ [$-3.7$, $+5.7$] \\
    \CHIP & 1.0 [0.0, 2.3] & 0.7 [0.0, 1.7] & 0.7 & 52.0 & $-2.0$ [$-7.4$, $+3.4$] \\
    \List (ref.) & 22.5 [17.8, 27.5] & 32.2 [26.8, 37.6] & 0.7 [0.0, 1.7] & 54.0 [48.3, 59.7] & -- \\
    \IsoPair (ref.) & 4.7 [2.3, 7.4] & 3.7 [1.7, 6.0] & 1.0 & 45.0 & $-9.1$ [$-15.4$, $-2.7$] \\
    \midrule
    \multicolumn{6}{@{}l}{\emph{mst24 (298 sets; \GCCR 297) and ur7 (239 sets)}} \\
    \PWC, mst24 & 0.3 [0.0, 1.0] & 0.0 [0.0, 0.0] & 0.3 [0.0, 1.0] & -- & -- \\
    \GCCR, mst24 & 0.3 [0.0, 1.0] & 0.0 [0.0, 0.0] & 0.3 [0.0, 1.0] & -- & -- \\
    \PWC, ur7 & 0.4 [0.0, 1.3] & 0.0 [0.0, 0.0] & 0.4 [0.0, 1.3] & -- & -- \\
    \GCCR, ur7 & 0.4 [0.0, 1.3] & 0.0 [0.0, 0.0] & 0.4 [0.0, 1.3] & -- & -- \\
    \bottomrule
  \end{tabular}
\end{table*}

The \List reference rows of \cref{tab:s-fix} use the closure of the isolation study and therefore read $+31.5$, $+23.4$ and $+22.5$ instead of the $+31.1$, $+22.4$ and $+23.5$ of \cref{tab:s-budget}; the conventions differ in the dominance closure, not in the data.
On mst24 the closure removed 265 decoys and 4 flips in 4 sets; without it \PWC's excess would read 1.7 [0.3, 3.4] with residual 0.0; on ur7 it removed 566 decoys and one flip (0.8 [0.0, 2.1] without it).
\GCCR on mst24 excluded 238 unreadable candidate rows (below the 2\% cap) and is computed on 297 sets.
The gate rarely matters: a decoy survives \GCC's gate in 0.07\% of q8 rows (19 of 28{,}538; 11 sets), 0.23\% of q32 rows (68 of 29{,}692; 23 sets) and 0.15\% of ivl38 rows (45 of 29{,}724; 25 sets); on q32's 23 survivor sets \PWC's residual is 4.3 [0.0, 13.0] where \GCC's is 0.0.
On the 58 adaptive sets, \GCC's excess is 3.4 [0.0, 8.6] with decoy wins 6.9 [1.7, 13.8], and \GCCR equals \PWC on every endpoint.

\paragraph{Agreement with human rankings.}
\Cref{tab:s-k3} gives K3+ for every readout.
\PWC raises winner agreement over \List by 1.4 [0.1, 2.8] points on q8 and 3.4 [2.1, 4.8] on q32, and pair agreement by 3.5 and 4.2.
Against \Rating the differences are smaller: winners $+0.2$ [$-1.4$, $+1.9$] (q8) and $+2.1$ [$+0.6$, $+3.6$] (q32), pairs $+0.8$ [$-0.1$, $+1.7$] and $+1.8$ [$+1.0$, $+2.6$].
Against \Rating, whose scores are too coarse to separate close candidates, \PWC orders more close pairs as the human ranking does.
On pairs whose \Rating expectations differ by less than 0.5, \PWC orders more pairs correctly than \Rating: $+4.8$ [$+2.0$, $+7.5$] on q8 ($n=2{,}431$) and $+4.9$ [$+2.3$, $+7.5$] on q32 ($n=2{,}224$); on pairs with a rating gap of at least 2 the difference is $+0.1$ [$-0.3$, $+0.5$] and $+0.4$ [$+0.0$, $+0.8$].
A pointwise rating cannot order close pairs that a comparison separates.

\begin{table*}[t]
  \centering
  \scriptsize
  \setlength{\tabcolsep}{2pt}
  \caption{\textbf{K3+ agreement with the HPDv2 human order} (1{,}200 development sets from 300 prompts; \%; 95\% prompt-cluster bootstrap). W: winner agreement; P: pair agreement; $\Delta$: paired against \List.}
  \label{tab:s-k3}
  \begin{tabular}{@{}lllllllll@{}}
    \toprule
    readout & q8 W & q8 P & q8 $\Delta$W & q8 $\Delta$P & q32 W & q32 P & q32 $\Delta$W & q32 $\Delta$P \\
    \midrule
    \List & 68.7 [66.8, 70.5] & 76.9 [75.8, 78.1] & -- & -- & 67.0 [65.1, 68.9] & 76.8 [75.6, 77.9] & -- & -- \\
    \Rating & 69.9 [68.0, 71.8] & 79.6 [78.4, 80.8] & $+1.2$ [$-0.4$, $+2.9$] & $+2.7$ [$+1.7$, $+3.7$] & 68.3 [66.3, 70.4] & 79.2 [78.0, 80.4] & $+1.3$ [$-0.2$, $+2.9$] & $+2.4$ [$+1.4$, $+3.3$] \\
    \PWC & 70.1 [68.1, 72.1] & 80.4 [79.2, 81.5] & $+1.4$ [$+0.1$, $+2.8$] & $+3.5$ [$+2.7$, $+4.2$] & 70.5 [68.5, 72.4] & 81.0 [79.8, 82.1] & $+3.4$ [$+2.1$, $+4.8$] & $+4.2$ [$+3.4$, $+5.0$] \\
    \GCC & 70.0 [68.0, 72.0] & 79.8 [78.7, 81.0] & $+1.3$ [$-0.0$, $+2.7$] & $+2.9$ [$+2.1$, $+3.7$] & 69.7 [67.8, 71.6] & 78.8 [77.6, 80.0] & $+2.7$ [$+1.4$, $+4.1$] & $+2.0$ [$+1.2$, $+2.9$] \\
    \GCCR & 70.2 [68.2, 72.1] & 80.5 [79.3, 81.6] & $+1.5$ [$+0.2$, $+2.9$] & $+3.6$ [$+2.8$, $+4.3$] & 70.3 [68.4, 72.2] & 80.7 [79.6, 81.8] & $+3.3$ [$+2.0$, $+4.6$] & $+3.9$ [$+3.2$, $+4.7$] \\
    \CHIP & 69.6 [67.7, 71.5] & 79.6 & $+0.9$ [$-0.6$, $+2.4$] & $+2.6$ [$+1.7$, $+3.6$] & 67.1 [65.2, 69.0] & 78.4 & $+0.1$ [$-1.3$, $+1.6$] & $+1.6$ [$+0.8$, $+2.5$] \\
    \IsoPair & 62.6 [60.6, 64.5] & 75.3 [74.2, 76.5] & $-6.1$ [$-7.7$, $-4.5$] & $-1.6$ [$-2.5$, $-0.7$] & 60.8 [58.9, 62.7] & 74.6 [73.4, 75.8] & $-6.2$ [$-8.0$, $-4.5$] & $-2.2$ [$-3.2$, $-1.2$] \\
    \bottomrule
  \end{tabular}
\end{table*}

\paragraph{Held-out confirmation.}
The 100 held-out sets of q32 were judged once, after three one-sided tests and their margins had been fixed (\cref{sec:s-protocol}), with a step-down adjustment for the three tests.
\List's excess on them is $+20.0$ [$+12.0$, $+28.0$] ($p=0.0002$); \PWC's residual is 0.0 [0.0, 0.0] against a margin of 4.0 points ($p=0.0002$); and \PWC minus \List in K3+ winner agreement on the 1{,}200 held-out sets is $+3.25$ [$+1.25$, $+5.25$] against a margin of $-1.0$ point ($p=0.0004$); every adjusted $p$ is 0.0006.
Descriptively, \PWC's excess, residual and decoy wins are all 0.0 (also with no closure; the closure removed 25 decoys and no \PWC flip), and \GCCR's are identical; \IsoPair's excess is 7.0 [3.0, 12.0], residual 1.0 [0.0, 3.0], decoy wins 7.0 and $R=0.35$ [0.12, 0.73].
Held-out K3+ winner agreement is 67.2 [63.7, 70.6] for \List, 69.1 [65.5, 72.7] for \Rating, 70.4 [67.5, 73.3] for \PWC, 70.5 [67.5, 73.4] for \GCCR and 59.2 [55.9, 62.5] for \IsoPair; pair agreement 77.0, 79.6, 81.1, 81.0 and 73.6.
\PWC minus \List in pairs is $+4.2$ [$+3.0$, $+5.3$]; \PWC minus \Rating is $+1.3$ [$-1.3$, $+4.0$] in winners and $+1.5$ [$-0.0$, $+3.3$] in pairs; \IsoPair minus \List is $-8.0$ [$-10.9$, $-5.2$] in winners.

\paragraph{Batched serving.}
The held-out rows were judged on a batched serving path and replayed once more one message per forward pass.
The two paths agree on 99.42\% of \List winners (1{,}193 of 1{,}200 K3+ sets) and 99.53\% of its pair signs, 99.50\% of \PWC winners and 99.72\% of its pair signs, and 99.25\% / 99.36\% for \IsoPair; on the attack rows, 98.99\% of \PWC winners (9{,}834 of 9{,}934), 98.94\% of \IsoPair's and 98.83\% of \List's (9{,}917 of 10{,}034).
The training rewards always use the one-pass path.

\paragraph{Latency.}
At $N=4$ the median decision on q32 takes 0.253\,s [IQR 0.251, 0.254] for one \List message, 1.524\,s [1.512, 1.536] for \PWC (6.02 times) and 1.079\,s [0.583, 1.087] for \GCCR (4.26 times).
At the training group size, $N=8$ on q8, one \List message takes 0.215\,s (0.207 with eight images, 0.223 with nine), \ListAvg's eight messages 1.66\,s, \Rating's eight 0.54\,s and \PWC's 56 pair messages 4.93\,s, about 23 times one \List message.
Certified early stopping lowers \PWC's selection cost to 24.3 messages per group including the proposal, 2.26\,s projected from per-message timing (\cref{sec:s-leader}); the Copeland training reward still needs all 56.

\subsection{Public benchmarks: MJ-Bench pairs and GenAI-Rank groups}
\label{sec:s-public}

\paragraph{MJ-Bench position test.}
On the 729 alignment pairs of MJ-Bench~\cite{chen2024mjbench}, each a caption, two images and the image that annotators prefer, every judge favours the image shown first (\mainref{sec:pair-results}): depending on the judge, the preferred image is picked 7 to 43\,\pp more often when it is shown first than when it is shown second.
Asked once, q8 picks the preferred image in 86.6\% of the pairs when that image is shown first and in 45.1\% when it is shown second; the two shares are 78.6 and 57.2\% on q32, 86.4 and 43.3\% on ivl38, 83.2 and 49.7\% on mst24, and 68.4 and 61.0\% on ur7.
The two-order margin is the same for both orders by construction.
It is right in 70.1\% \ci{66.8}{73.4} of the pairs on q8, 70.2\% \ci{66.9}{73.4} on q32, 70.5\% \ci{67.2}{73.8} on ivl38, 69.4\% \ci{66.1}{72.7} on mst24 and 64.5\% \ci{61.0}{67.9} on ur7, against 65.8, 67.9, 64.8, 66.4 and 64.7\% for one message asked in a random order.
The difference is $+4.3$ \ci{+1.9}{+6.6}\,\pp on q8, $+2.3$ \ci{+0.3}{+4.3} on q32, $+5.7$ \ci{+3.4}{+8.2} on ivl38 and $+3.0$ \ci{+0.8}{+5.1} on mst24; on ur7, whose position gap is the smallest, the two agree.

\paragraph{Real groups of GenAI-Rank.}
For eight of the nine SD-XL samples of a GenAI-Rank prompt~\cite{li2024genaibench} (\mainref{sec:measure-groups}), the searched excess under q8 grows with the number $K$ of added images, both families cut to their first $K$ members: $+4.9$ \ci{+0.8}{+9.5}\,\pp at $K=2$, $+12.8$ \ci{+7.8}{+18.1} at 5, $+17.3$ \ci{+11.9}{+22.6} at 10 and $+22.3$ \ci{+16.9}{+28.1} at 20, up to $+24.3$ \ci{+18.5}{+30.0} with all decoys of a group.
Averaging each message over eight orders leaves $+19.8$ \ci{+14.4}{+25.1}\,\pp.
Under one \List message, $1.23$ \ci{1.05}{1.40} more of the eight samples change the sign of their rank-based GRPO advantage across the decoys than across the re-renders.

\clearpage
\section{Training details and extended results}
\label{sec:s-training}

\subsection{Reward audits and best-of-8, extended}
\label{sec:s-audit}

\paragraph{Audit groups.}
128 HPDv2 training prompts with eight SD3.5-Medium rollouts each, sampled with the deterministic ODE sampler, judged by q8 on the one-pass path; no message was unreadable in any family.
\Cref{tab:s-audit-perturb} shows how much the \List reward of a group moves when the same eight images are reordered, when one image is added, or when the message is repeated.
Reordering moves it most: the top pick changes in 62.5\% of groups, against 29.0\% for an added decoy (difference $-33.1$ [$-43.5$, $-22.6$]), and a decoy moves it no more than a re-render ($+0.8$ [$-1.6$, $+4.0$]).
Because an added image moved the \List reward less than reordering did, the training study compares reward protocols on ordinary groups instead of training with injected decoys.
\PWC's and \Rating's rewards were unchanged by the extra in 128 of 128 groups (maximum absolute difference 0.0), because neither ever shows the extra to the judge; 124 of the 128 groups had a qualifying decoy.

\begin{table*}[t]
  \centering
  \scriptsize
  \setlength{\tabcolsep}{5pt}
  \caption{\textbf{How much the \List reward moves} (128 audit groups of eight rollouts, q8; \%; 95\%). Sign flip: the sign of some rollout's GRPO advantage changes, from the label probabilities (P) or from the ranks.}
  \label{tab:s-audit-perturb}
  \begin{tabular}{@{}lrllll@{}}
    \toprule
    perturbation & groups & top-1 change & Kendall $\tau$ & sign flip (P) & sign flip (rank) \\
    \midrule
    reorder & 128 & 62.5 [53.9, 70.3] & $+0.329$ [$+0.286$, $+0.373$] & 75.0 [68.0, 82.0] & 89.1 [83.6, 93.8] \\
    add a grey image & 128 & 21.1 [14.1, 28.9] & $+0.719$ [$+0.686$, $+0.750$] & 45.3 [36.7, 53.9] & 53.1 [44.5, 61.7] \\
    add a decoy & 124 & 29.0 [21.0, 37.1] & $+0.687$ [$+0.649$, $+0.723$] & 53.2 [45.2, 62.1] & 58.9 [50.0, 67.7] \\
    add a re-render & 124 & 28.2 [20.2, 36.3] & $+0.686$ [$+0.647$, $+0.722$] & 51.6 [42.7, 60.5] & 58.9 [50.0, 67.7] \\
    repeat the message & 128 & 0.0 & $+1.000$ & 0.0 & 0.0 \\
    \bottomrule
  \end{tabular}
\end{table*}

\paragraph{Readouts on the audit groups.}
\Cref{tab:s-w3x} compares the readouts as rewards on the same 128 groups.
Averaging \List over more cyclic shifts lowers its reorder rate from 62.5\% (one message) through 46.9\% (two shifts) and 38.3\% (four) to 25.0\% (eight, \ListAvg), and its added-decoy rate to 18.5\%; it never reaches zero.
Batch calibration of the \List logits lowers the reorder rate only to 54.7\%; a content-free calibration is not interpretable here, because the grey-message prior is extremely uneven across slots, and is not reported.
\PWC's Copeland reward correlates more closely with HPSv2.1 than \ListAvg's ($\rho$ 0.51 against 0.39); the difference in HPSv2.1 top-1 agreement between \ListAvg and \PWC is $-2.99$ [$-10.94$, $+4.95$], and \ListAvg's top pick equals \PWC's cover winner in 55.1 [47.3, 62.9]\% of groups (one \List message: 40.6\%).
Of the pairwise readouts, \PRP reaches $\rho$ 0.48 and changes its top pick under reordering in 15.6\% of groups through its tie rule (its top score is tied in 23.4\% of groups), and \PairOne reaches $\rho$ 0.41 and changes its top pick under reordering in 48.4\%.
For comparison, HPSv2.1 and PickScore agree on the top image in 41.4 [32.8, 50.0]\% of groups and correlate at $\rho=0.518$.

\begin{table*}[t]
  \centering
  \scriptsize
  \setlength{\tabcolsep}{3pt}
  \caption{\textbf{Readouts as rewards on the 128 audit groups} ($N=8$, q8; \%; 95\% where printed). Messages and judge seconds per group; top-1 change under reordering and under one added image; agreement of the top pick with HPSv2.1's and PickScore's; within-group Spearman $\rho$ with HPSv2.1. BC: batch-calibrated.}
  \label{tab:s-w3x}
  \begin{tabular}{@{}lrrlllllll@{}}
    \toprule
    reward & msgs & s & reorder & + grey & + decoy & + re-render & = HPS & = Pick & $\rho$ HPS \\
    \midrule
    \List, one message & 1 & 0.22 & 62.5 [53.9, 70.3] & 21.1 [14.1, 28.9] & 29.0 [21.0, 37.1] & 28.2 [20.2, 36.3] & 26.6 [18.8, 34.4] & 35.9 & $+0.36$ [$+0.30$, $+0.42$] \\
    \List, two shifts & 2 & 0.42 & 46.9 [38.3, 55.5] & 24.2 & 29.8 & 30.6 & 32.0 [25.0, 39.1] & 30.9 & $+0.38$ \\
    \List, four shifts & 4 & 0.83 & 38.3 [29.7, 46.9] & 21.9 & 22.6 & 21.0 & 33.6 [26.2, 41.0] & 33.6 & $+0.38$ \\
    \ListAvg, eight shifts & 8 & 1.66 & 25.0 [18.0, 32.8] & 13.3 [7.8, 19.5] & 18.5 [12.1, 25.8] & 16.9 [10.5, 24.2] & 33.6 [25.8, 41.4] & 31.6 & $+0.39$ [$+0.34$, $+0.44$] \\
    \List, BC & 1 & 0.22 & 54.7 [46.1, 63.3] & 28.9 & 32.3 & 31.5 & 29.7 [23.0, 36.7] & 30.5 & $+0.36$ \\
    \ListAvg, BC & 8 & 1.66 & 24.2 [17.2, 31.2] & 16.4 & 18.5 & 19.4 & 36.3 [28.5, 44.5] & 31.6 & $+0.41$ [$+0.36$, $+0.46$] \\
    \Rating & 8 & 0.54 & 0 & 0 & 0 & 0 & 29.7 [21.9, 37.5] & 30.3 & $+0.33$ [$+0.26$, $+0.39$] \\
    \PWC, Copeland & 56 & 4.93 & 0 & 0 & 0 & 0 & 36.6 [28.5, 44.8] & 34.5 & $+0.51$ [$+0.45$, $+0.56$] \\
    \bottomrule
  \end{tabular}
\end{table*}

\paragraph{Best-of-8 on the audit groups (post hoc).}
Best-of-8 was first computed on the audit groups after their rewards were known, and then repeated on 256 new groups (below).
On the audit groups (oracle gain 2.59 [2.34, 2.85] HPSv2.1 $\times100$) one \List message recovers 36.4 [24.0, 48.1]\% of the oracle gain (43.3 [33.5, 52.8]\% with both base orders), two, four and eight shifts 49.1, 53.0 and 55.1 [44.1, 64.5]\%, \Rating 47.7 [36.8, 57.8]\%, and \PWC 57.9 [47.3, 67.5]\% with the Copeland argmax and 59.3 [48.6, 68.8]\% with the cover winner.
Paired, \PWC's Copeland argmax gains $+21.6$ [$+8.6$, $+34.5$] points over one \List message, $+2.8$ [$-4.8$, $+10.4$] over \ListAvg and $+10.2$ [$-0.1$, $+20.7$] over \Rating.
On PickScore the shares are 35.2 (\List), 36.3 (\ListAvg), 39.0 (\Rating) and 42.4--42.6\% (\PWC).
With an added decoy counted as a loss when picked, one \List message recovers 32.4 [19.5, 44.5]\% ($-4.4$ [$-14.0$, $+5.0$] against the clean message), \ListAvg 49.8\%, \PWC 57.3--58.7\% and \Rating 47.3\%; a decoy tops a single nine-image \List message in 1 of 992 shifted messages.

\paragraph{Best-of-8 on 256 new groups.}
On 256 new prompts, with the analysis fixed before their rollouts were generated, the selection shares are (\% of the oracle HPSv2.1 gain, 95\% prompt bootstrap): \List 47.5 [41.5, 53.1], \ListAvg 58.0 [52.0, 63.8], \PRP 56.1 [49.1, 62.1] with its paper tie rule and 55.3 [48.9, 61.1] with ties split, \PairOne 47.7 [41.6, 53.6], and \PWC 62.7 with the Copeland argmax and 64.2 [57.9, 70.2] with the cover winner.
Paired against the cover winner on the same groups, the cover winner gains $+21.0$ [$+13.2$, $+29.3$] points over \Rating, $+16.7$ [$+10.0$, $+23.6$] over \List, $+6.2$ [$+0.4$, $+12.0$] over \ListAvg, $+8.1$ [$+3.3$, $+13.4$] over \PRP (ties split: $+8.8$ [$+4.2$, $+13.7$]), $+16.4$ [$+11.0$, $+22.0$] over \PairOne and $+1.5$ [$+0.1$, $+3.1$] over the Copeland argmax; against the Copeland argmax instead, \PRP is $-6.6$ [$-11.7$, $-1.7$] and \PairOne $-15.0$ [$-20.4$, $-9.5$].
On the same 256 groups \List's within-group $\rho$ with HPSv2.1 is 0.345 [0.307, 0.382] and its top pick changes under reordering in 57.8 [51.6, 64.1]\% of groups; for \ListAvg, 0.444 [0.402, 0.483] and 28.9 [23.4, 34.4]\%.
On the 128 audit groups, analysed after the fact, \PWC reached 57.9--59.3 against \PRP's 56.7--56.9 ($+1.2$ [$-5.7$, $+8.1$]) and \PairOne's 41.5--43.7 (14.3 to 16.4 points below \PWC, every interval excluding 0).
On the new groups the cover winner selects better than every other readout, with intervals that exclude 0 against \PRP and \ListAvg as well, and much better than \PairOne, the weakest pairwise selector (\cref{sec:s-pairone}).

\paragraph{Published selection baselines.}
Three published readouts were added to Tab.~\ref{M-tab:protocols} of the main paper after the shares above were known; they are computed from the stored judge answers of the same groups, with no new judge query.
Permutation self-consistency~\cite{tang2024found} ranks the candidates in each rotation of the \List message by their label probabilities and returns a Kemeny-optimal aggregate of these rankings, found by exhaustive search; the rotations stand in for the random shuffles of the original method, because they are the orders whose answers are stored.
Several optimal rankings with different tops occur in 115 of the 512 cases (22.5\%; the 256 groups in two base orders), where best-of-8 splits the pick among these tops and the Reorder column takes the one standing first in the base order; the cover winner gains $+6.3$ [$+0.7$, $+12.0$] points over it.
The knockout follows the best-of-$N$ procedure of MMRB2~\cite{hu2025mmrb2}: a single-elimination bracket over the eight rollouts in the base order, with one stored pair query per match and the earlier-positioned rollout shown first.
The first-shown image wins 64.7\% of the 14{,}336 single-order queries of these groups, and the pick of the knockout differs between the two stored base orders in 51.6\% of the groups; its best-of-8 share is the mean over the two, and the cover winner gains $+15.5$ [$+9.7$, $+21.5$] points over it.
The repeated tournament follows the pairwise structure of Tournament-GRPO~\cite{yang2026tournamentgrpo}: each of $M$ repeats shuffles the active candidates before every round, pairs them, asks one stored query per pair and gives the winner a point, and the pick is the rollout with the most points over the $M$ repeats, with ties split.
Because the shuffle is part of this readout, its best-of-8 cells are means over 200 seeded runs per group (20 per decoy set for the searched excess below), and its Reorder cell is the probability that two runs pick different rollouts.
With $M=1$, 3 and 8 repeats, which cost 7, 21 and 56 messages, it recovers 48.5 [43.5, 53.2], 50.1 [44.9, 54.8] and 52.3 [46.8, 57.3]\% of the oracle gain, and the cover winner gains $+15.6$ [$+11.5$, $+19.9$], $+14.1$ [$+10.1$, $+18.2$] and $+11.9$ [$+7.9$, $+16.1$] points over it; the main paper prints $M=3$.
On the 128 audit groups (\mainref{tab:protocols}), the top pick changes under reordering in 32.8\% of groups for permutation self-consistency, 50.8\% for the knockout and 44.7\% for the repeated tournament, and their within-group $\rho$ with HPSv2.1 is 0.43, 0.30 and 0.39.
Bradley--Terry strengths fitted to the 56 single-order votes~\cite{xu2025nontransitivity} have the same top candidates as \PRP in all 256 groups and recover 55.3 [48.9, 61.1]\% with ties split, so they have no row of their own.
On the decoy sets of \mainref{tab:judges}, the searched excess on q8 and q32 is 27.0 and 22.4 for permutation self-consistency, 3.1 and 1.3 for the knockout and 1.1 and 0.1 for the repeated tournament; \Rating's is 0.3 on both, not 0 because the dominance screen rates under another rubric (\cref{sec:s-notation}).
The eight paired best-of-8 contrasts of the cover winner with the other readouts of \mainref{tab:protocols} are reported with their 95\% intervals and carry no multiplicity correction.

\paragraph{Exchanging one rollout.}
\label{sec:s-exchange}
The Exch.\ column of \mainref{tab:protocols} asks whether the pick among seven rollouts depends on which dominated rollout fills the eighth slot.
For each audit group the base model draws 50 further rollouts of the same prompt with new noises and the sampler of the audit rollouts; the first seven rollouts of the group keep their slots, and the eighth slot takes one of the new rollouts.
A new rollout qualifies when q8 rates it alone, with the screen of \cref{sec:s-rubrics}, at least 1.0 below the two top-rated kept rollouts and prefers each of the seven kept rollouts to it in both orders of the pair message of \cref{sec:s-pair-prompts}.
119 of the 128 groups have at least ten qualifying rollouts; the other nine are excluded and counted.
In each of the 119 groups every readout is read once with each qualifying rollout in the eighth slot, at most fifty tries, and once with each of as many re-renders of the first qualifying rollout (\cref{sec:s-search}); $f=1$ when its pick among the seven kept rollouts is not the same for every try, and the endpoint is the excess of the exchanged rollouts over the re-renders, as for the searched excess of \mainref{sec:measure-setup}, with a 95\% bootstrap interval over groups.
The repeated tournament uses the same shuffle seeds for every try.

Under one \List message the pick among the seven is not the same for every exchanged rollout in 43.7\% of the groups, against 26.1\% for the re-renders, an excess of $+17.6$ [$+10.9$, $+24.4$]\,\pp (\cref{tab:s-exchange}); \ListAvg and permutation self-consistency (\textsf{PSC}), which read the same listwise message in eight rotations, keep $+11.8$ and $+10.9$.
Every other readout shows 0.0 in both families, as its construction implies: \Rating judges each rollout alone, and the exchanged rollout loses every pair, so under \PairOne and \PRP it adds the same win to the count of every kept rollout (\cref{prop:s-reward}(i)), in the knockout and the repeated tournament it loses its first match and leaves every other match as it was, and under \PWC it is covered, so the cover winner stays the same (\cref{prop:s-ddp}(a)).
The training rewards move in the same way.
Per group we count the kept rollouts whose GRPO advantage, computed from the reward ranks as in \cref{tab:s-audit-perturb}, does not have the same sign for every try.
Across the exchanged rollouts, on average 1.05 more of the seven change sign than across the re-renders under the \List reward, and 0.71 more under the \ListAvg reward.
Under the \PWC reward the number is 0: every exchanged rollout gives each kept rollout one more win and wins nothing itself, so the reward vector of the group, and with it every advantage, is the same for every try (\cref{prop:s-reward}, \mainref{sec:reward}).

\begin{table}[t]
  \centering
  \scriptsize
  \setlength{\tabcolsep}{4pt}
  \caption{\textbf{Exchanging one dominated rollout} (119 audit groups, q8; \% of groups whose pick among the seven kept rollouts is not the same for every try; excess: exchanged minus re-renders, \pp, 95\%).}
  \label{tab:s-exchange}
  \begin{tabular}{@{}lrrl@{}}
    \toprule
    readout & exchanged & re-renders & excess \\
    \midrule
    \List & 43.7 & 26.1 & $+17.6$ [$+10.9$, $+24.4$] \\
    \ListAvg & 30.3 & 18.5 & $+11.8$ [$+5.9$, $+17.6$] \\
    \textsf{PSC} & 29.4 & 18.5 & $+10.9$ [$+5.0$, $+16.8$] \\
    \Rating & 0.0 & 0.0 & 0.0 \\
    \PRP & 0.0 & 0.0 & 0.0 \\
    \PairOne & 0.0 & 0.0 & 0.0 \\
    knockout & 0.0 & 0.0 & 0.0 \\
    repeated tournament & 0.0 & 0.0 & 0.0 \\
    \PWC & 0.0 & 0.0 & 0.0 \\
    \bottomrule
  \end{tabular}
\end{table}

\subsection{Training recipe and arms}
\label{sec:s-recipe}

\paragraph{Policy and recipe.}
Stable Diffusion 3.5 Medium~\cite{esser2024sd3,stability2024sd35} with LoRA adapters is trained with the GRPO recipe of Flow-GRPO's public SD3.5 configuration~\cite{liu2025flowgrpo} for 200 steps (600 for the longer runs of \cref{sec:s-further}), with no checkpoint selection, at 64 images per step: eight rollouts per prompt, judged as one group of $G=8$.
The KL weight is 0.4 instead of the configuration's 0.04, for the reason given in \cref{sec:s-pilot}; the configuration's own batch and KL weight are run in \cref{sec:s-publicbatch}.
The per-group advantage is $(c_i-\bar c)/(\sigma+10^{-4})$ over the eight scored rollouts, clipped to $\pm5$ (\cref{prop:s-reward}); with $G=8$ the clip never binds.
Training prompts are HPDv2 training prompts, disjoint from every evaluation prompt.
A seed fixes the prompt order, the initial noise and the per-step sampler noise and nothing arm-dependent, so the arms of a seed see the same prompts and start from the same rollouts, and their step-0 rollout HPSv2.1 is identical.
The judge is q8 on the one-pass path; the trainer refuses any scored group that was not judged one message per forward pass.

\paragraph{Arms.}
Every arm of the q8 study uses the same judge, rubric and recipe; only the reward readout differs (\cref{sec:s-readouts}). The arms with other reward sources are described in \cref{sec:s-further}.
\begin{itemize}[leftmargin=*,itemsep=1pt,topsep=2pt]
\item \List: one message with the eight rollouts; reward the normalised rank of $P(\text{winner})$; 1 message per group.
\item \ListAvg: the \List message in all eight cyclic shifts, the label distributions mapped back to the rollouts and averaged, then the same rank transform; shift 0 is \List's message byte for byte; 8 messages.
\item \PairOne: the 28 unordered pairs, lower rollout index shown first, one order each; reward the win count divided by 7 (win 1, tie or unreadable 0.5); 28 messages; the same images and pair prompt as \PWC, whose same-order margins it reproduces bitwise at step 0.
\item Tournament-GRPO: the pairwise structure of Tournament-GRPO~\cite{yang2026tournamentgrpo} (\cref{sec:s-audit}) as a reward: three repeats of a single-elimination bracket, each shuffling the rollouts before every round and asking one single-order pair message per match; a rollout gains one point per match won, and its reward is its points divided by 3; 21 messages.
\item \PRP: the 56 ordered pair messages; reward the hard two-order votes against the other seven rollouts (1 when both orders prefer the rollout, $\tfrac12$ on a conflict, a tie or an unreadable answer, 0 when both prefer the opponent), summed and divided by 7; 56 messages, the same as \PWC's.
\item \PWC: the 56 ordered pair messages; reward the Copeland score over the eight rollouts divided by 7; 56 messages.
\item \Rating: each rollout alone; reward the within-group rank of the expected rating; 8 messages.
\item Random-reward control: the same trainer with a random reward; 3 seeds.
\end{itemize}
The arms differ in reward construction as well as in readout (\List, \ListAvg and \Rating use within-group ranks, \PairOne, \PRP and \PWC win fractions, Tournament-GRPO tournament points), so a difference between arms is attributed to the complete reward protocol.
Every arm of the q8 study except the controls has 16 seeds, and \List, \ListAvg, \PairOne, \PRP and \PWC have 8 more seeds trained to step 600; the arms of \cref{sec:s-further} have 8 seeds (FLUX.1-dev: 6; public batch and KL weight: 3).

\subsection{Evaluation and statistics}
\label{sec:s-train-eval}

Each trained policy generates one image per prompt for 1{,}200 pinned evaluation prompts (HPDv2 test prompts and withheld HPDv2 training prompts) with the deterministic ODE sampler and a fixed noise per prompt, and is compared with the base model's image from the same noise.
The endpoint is the HPSv2.1 win rate, the share of prompts on which the policy's image scores higher (ties count one half), at step 200, and for the longer runs also at steps 100, 400 and 600; PickScore~\cite{kirstain2023pickapic} is read the same way.
A contrast between two arms is two-sided; its interval is the wider of a two-level bootstrap (seeds, then prompts; 5{,}000 resamples) and a Welch $t$ interval over seeds, and its $p$-value the larger of the two.
Seed-paired versions (df 15) are reported for four of the q8 contrasts (\cref{tab:s-contrasts}).
A contrast whose interval contains 0 is reported as not establishing a difference, never as equivalence.

\subsection{Seed statistics and contrasts}
\label{sec:s-seeds}

\Cref{tab:s-seeds} gives the mean and spread over seeds of every q8 arm and \cref{tab:s-contrasts} the contrasts between them.
\PWC raises the mean HPSv2.1 win rate over \List by 11.39 [+7.32, +15.51] points and has the highest mean of the seven arms.
At 16 against 16 seeds its contrast with each of the other six excludes 0 on HPSv2.1 and on PickScore, except with \PRP, where it is $+3.2$ [$-0.5$, $+6.9$] on HPSv2.1 and $+1.5$ [$-0.1$, $+3.1$] on PickScore (\cref{tab:s-contrasts}); with seeds 17 to 24 added, the HPSv2.1 contrast with \PRP is $+2.6$ [$-0.4$, $+5.6$] at 24 against 24 seeds, an interval that also includes 0 (\cref{sec:s-further}).
An earlier first stage with 8 \List, 4 \PWC and 5 \Rating seeds was inconclusive, and the three arms were run to 16 seeds (\cref{sec:s-protocol}).

\begin{table}[t]
  \centering
  \scriptsize
  \setlength{\tabcolsep}{4pt}
  \caption{\textbf{Win rates against the base by arm} (\%; step 200; 1{,}200 pinned prompts; KL weight 0.4). Mean (SD over seeds) of the HPSv2.1 and PickScore win rates.}
  \label{tab:s-seeds}
  \begin{tabular}{@{}lrll@{}}
    \toprule
    arm & seeds & HPSv2.1 & PickScore \\
    \midrule
    \List & 16 & 55.02 (6.20) & 53.05 (3.04) \\
    \ListAvg & 16 & 62.12 (5.54) & 54.92 (3.10) \\
    \PairOne & 16 & 62.57 (5.02) & 56.83 (2.73) \\
    Tournament-GRPO & 16 & 60.40 (5.70) & 55.60 (2.90) \\
    \PRP & 16 & 63.20 (5.30) & 57.60 (2.40) \\
    \PWC & 16 & 66.41 (5.00) & 59.06 (1.97) \\
    \Rating & 16 & 56.20 (6.00) & 53.33 (3.24) \\
    random reward & 3 & 42.23 (6.55) & 50.50 (1.48) \\
    \bottomrule
  \end{tabular}
\end{table}

\begin{table*}[t]
  \centering
  \scriptsize
  \setlength{\tabcolsep}{2.1pt}
  \caption{\textbf{Contrasts between training arms} (win-rate points on the 1{,}200 pinned prompts; 16 against 16 seeds; 95\%; two-sided $p$). Unpaired: the wider of the two-level bootstrap and Welch $t$; seed-paired: $t$ over 16 seed differences.}
  \label{tab:s-contrasts}
  \begin{tabular}{@{}lllll@{}}
    \toprule
    contrast & HPSv2.1, unpaired & HPSv2.1, seed-paired & PickScore, unpaired & PickScore, seed-paired \\
    \midrule
    \PWC $-$ \List & $+11.39$ [$+7.32$, $+15.51$], $p$ .0004 & $+11.39$ [$+7.02$, $+15.77$], $p$ .0001 & $+6.04$ [$+3.98$, $+8.14$]$^\ast$, $p$ .0004 & $+6.01$ [$+4.12$, $+7.88$], $p<$.0001 \\
    \ListAvg $-$ \List & $+7.11$ [$+2.86$, $+11.36$], $p$ .002 & -- & -- & -- \\
    \PairOne $-$ \List & $+7.55$ [$+3.47$, $+11.63$], $p$ .0008 & $+7.55$ [$+3.56$, $+11.55$], $p$ .0011 & $+3.78$ [$+1.48$, $+6.03$], $p$ .0009 & $+3.78$ [$+1.23$, $+6.17$], $p$ .0042 \\
    \PWC $-$ \ListAvg & $+4.28$ [$+0.47$, $+8.10$], $p$ .029 & -- & $+4.14$ [$+2.10$, $+6.27$], $p$ .0004 & -- \\
    \PWC $-$ \PairOne & $+3.84$ [$+0.22$, $+7.46$], $p$ .038 & $+3.84$ [$+0.04$, $+7.64$], $p$ .048 & $+2.23$ [$+0.27$, $+4.13$], $p$ .025 & $+2.23$ [$+0.37$, $+4.09$], $p$ .022 \\
    \ListAvg $-$ \PairOne & $-0.44$ [$-4.26$, $+3.38$], $p$ .81 & $-0.44$ [$-3.32$, $+2.44$], $p$ .75 & $-1.92$ [$-4.19$, $+0.34$], $p$ .098 & $-1.92$ [$-4.01$, $+0.09$], $p$ .066 \\
    \PWC $-$ \PRP & $+3.2$ [$-0.5$, $+6.9$], $p$ .09 & -- & $+1.5$ [$-0.1$, $+3.1$], $p$ .07 & -- \\
    \PWC $-$ Tourn. & $+6.0$ [$+2.1$, $+9.9$], $p$ .004 & -- & $+3.5$ [$+1.7$, $+5.3$], $p$ .0005 & -- \\
    \PWC $-$ \Rating & $+10.21$ [$+6.22$, $+14.20$], $p$ .0004 & -- & $+5.74$ [$+3.78$, $+7.70$]$^\ast$, $p$ .0004 & -- \\
    \Rating $-$ \List & $+1.18$ [$-3.23$, $+5.59$], $p$ .59 & -- & $+0.30$ [$-1.97$, $+2.57$]$^\ast$, $p$ .79 & -- \\
    \bottomrule
  \end{tabular}
  \par\smallskip\parbox{\textwidth}{\scriptsize $^\ast$ From the re-scored evaluation of \cref{sec:s-fresh} (the same images scored again; the \List and \PWC PickScore means differ from the training evaluation by 0.02 and 0.01 points). Tourn.: Tournament-GRPO. ``--'': not computed for this contrast.}
\end{table*}

\subsection{The one-order pairwise reward}
\label{sec:s-pairone}

The \PairOne arm was checked for implementation faults after its result was read; none was found.
Its 16 configurations name the right arm, protocol and seed; it shares \PWC's evaluation set, base images and sampler; every run completed 200 steps; and its per-seed win rates all differ from \PWC's.
At step 0, with the same prompts and rollouts as \PWC, its reward is a different function of the same images: the reward vectors are identical in 10 of 128 groups, the pair signs agree in 85\% of pairs, the top picks in 56\% of groups, and the rewards correlate at Spearman 0.84.
In \PairOne the first-shown image wins 62.5\% of the 716{,}800 pair messages of its 16 seeds to step 200, so its mean reward falls with the rollout index, from 0.625 for the first rollout to 0.376 for the last, while \PWC's is flat at 0.50.
The rollout index is independent of image content, which is consistent with the bias adding advantage noise to GRPO, where \PairOne trails \PWC by 3.84 points (\cref{tab:s-contrasts}), while in best-of-8 it translates directly into picking low-index images (16.4 points below \PWC's cover winner, \cref{sec:s-audit}).
We report this as consistent with the observed pattern, not as a demonstrated cause.

\subsection{Drift controls, the pilot and the KL weight}
\label{sec:s-pilot}

\paragraph{Pilot at KL weight 0.04.}
The design was first run at the KL weight of the public configuration, 0.04, with 64 images per step.
Seven runs finished (two were stopped early): \List 33.4, 44.7 and 19.8\%; \PWC 17.8 and 78.8\%; \Rating 19.2 and 41.2\% HPSv2.1 win rate against the base.
A learning check on \List's training-rollout gain gave $-0.810$, 95\% $t$ interval [$-4.03$, $+2.41$], and every arm showed a painterly high-frequency texture.
We compute no contrast on these runs.
At the configuration's own batch of 1{,}152 images per step the same KL weight of 0.04 trains (\cref{sec:s-publicbatch}).

\paragraph{Oracle-reward diagnostics.}
To separate the recipe from the judge, the unmodified trainer was run with the reward replaced, for 80 steps on seed 1 (\cref{tab:s-oracle}).
At 0.04 even an HPSv2.1 reward lowered the probe score and a random reward lowered it further; at 0.4 an HPSv2.1 reward raised the probe at all eight probe points, a random reward stayed within [$-0.0039$, $+0.0015$], and the images were clean.
We read this as reward-independent noise drift of a sample-starved step: 64 images per step against 1{,}152 in the public 32-GPU configuration, where the KL term's gradient is 0.045 of the policy-gradient term's at 0.04.
The limits are one seed per cell, 80 steps against 200, a 32-prompt probe, only 0.04 and 0.4 tried, and a KL weight chosen with an oracle reward on a prefix of the evaluation prompts after the pilot's per-arm win rates had been seen.
The 16-seed study uses the same design at 0.4.

\begin{table}[t]
  \centering
  \scriptsize
  \setlength{\tabcolsep}{1.8pt}
  \caption{\textbf{Oracle-reward diagnostics} (no judge; 80 steps; seed 1; HPS: HPSv2.1; probe: mean HPSv2.1 difference to the base images on 32 pinned prompts at step 80, 95\%; rollouts: HPSv2.1 of the training rollouts, steps 0--9 $\to$ 70--79).}
  \label{tab:s-oracle}
  \begin{tabular}{@{}llll@{}}
    \toprule
    KL, variant & reward & probe, step 80 & rollouts \\
    \midrule
    0.04 & HPS & $-0.0328$ [$-0.0409$, $-0.0248$] & 0.2654 $\to$ 0.2659 \\
    0.04 & random & $-0.0506$ & 0.2623 $\to$ 0.2352 \\
    0.04 & HPS rank & $+0.0040$ [$-0.0019$, $+0.0100$] & $\to$ 0.2839 \\
    0.04, EMA 0.95 & HPS & $-0.0060$ [$-0.0130$, $+0.0006$] & -- \\
    0.04, exact KL & HPS & $+0.0017$ [$-0.003$, $+0.006$] & -- \\
    0.4 & HPS & $+0.0070$ [$+0.0017$, $+0.0120$] & 0.2680 $\to$ 0.2864 \\
    0.4 & random & $-0.0039$ & 0.2659 $\to$ 0.2740 \\
    \bottomrule
  \end{tabular}
\end{table}

\paragraph{Learning beyond drift at 0.4.}
On the first-stage seeds 1--3, the training-rollout gain $G$ ($100\times$ HPSv2.1 of steps 190--199 minus steps 0--9) of each arm minus that of the random-reward control of the same seed averaged $+0.757$ (\List), $+1.026$ (\PWC) and $+0.536$ (\Rating); over the nine differences $+0.773$ [$+0.488$, $+1.059$] (df 8; anti-conservative, since the three arms of a seed share one control), and over the three seed means [$-0.051$, $+1.597$] (df 2).
The arms beat their seed's control in win rate by $+24.2$, $+21.4$ and $+1.7$ (\List), $+19.5$, $+23.8$ and $+7.2$ (\PWC), and $+16.3$, $+19.4$ and $-4.0$ (\Rating) points; the three controls themselves reach 36.8, 40.4 and 49.5\%.
None of the 17 first-stage runs ended with a final 64-prompt probe below $-0.01$.
Every first-stage run started once with no resume, and the image grids show no painterly or mosaic texture.

\paragraph{Reward-path checks on the pilot groups.}
A read-only review of the stored training groups of the pilot (and the first 38 steps of one \PWC run at 0.4) found no fault in the reward path.
In those groups the within-group $\rho$ of the reward with HPSv2.1 is 0.27--0.29 for \List, 0.42--0.47 for \PWC (0.49 at 0.4) and 0.32--0.36 for \Rating, against 0.51--0.57 between HPSv2.1 and PickScore; the reward correlates with the rollout's slot at $+0.44$ to $+0.48$ for \List and 0.00 for \PWC and \Rating; and every pair and rating was readable.
These pilot numbers are not combined with the audit numbers above, which come from different groups.

\paragraph{KL weight.}
We repeated \List and \PWC at KL weights 0.2 and 0.8 on seeds 1--3, with the 0.4 runs of the same seeds for parity and one random-reward control (seed 1) per weight (\cref{tab:s-kl}); seeds share prompts, initial noise and sampling noise across weights and arms, so the arm difference is paired by seed.
\PWC minus \List in HPSv2.1 win rate is positive at all three weights, $+13.4$ [$-37.8$, $+64.6$] at 0.2, $+1.1$ [$-11.9$, $+14.0$] at 0.4 and $+1.9$ [$-4.5$, $+8.3$] at 0.8, so its sign is stable across a $4\times$ range of the KL weight; three seeds resolve no difference at any weight (at 0.4 the same three seeds give $+1.1$ against $+11.39$ over 16 seeds), and the PickScore difference is $+0.8$, $+1.5$ and $-0.5$.
At 0.2 training is unstable: seed 1 of both arms ends at 29.8 and 27.2\%, the level of that weight's random-reward control (29.7\%), while the other four runs reach 39.2 to 75.8\%.
At 0.4 and 0.8 each arm's mean exceeds the control of its weight by 14.9 to 22.3 points.
The image grids of these runs were not inspected.

\begin{table}[t]
  \centering
  \scriptsize
  \setlength{\tabcolsep}{2pt}
  \caption{\textbf{KL weight} (seeds 1--3; step 200; 1{,}200 pinned prompts). Win rate against the base (\%), mean over seeds with a 95\% $t$ interval (df 2, not truncated to [0, 100]); $\Delta$: \PWC $-$ \List paired by seed; control: random reward, seed 1, same weight.}
  \label{tab:s-kl}
  \begin{tabular}{@{}lllll@{}}
    \toprule
    KL & \List & \PWC & $\Delta$ & control \\
    \midrule
    \multicolumn{5}{@{}l}{\emph{HPSv2.1}} \\
    0.2 & 43.4 [3.4, 83.4] & 56.8 [$-7.8$, 121.4] & $+13.4$ [$-37.8$, $+64.6$] & 29.7 \\
    0.4 & 58.0 [43.4, 72.6] & 59.1 [48.0, 70.2] & $+1.1$ [$-11.9$, $+14.0$] & 36.8 \\
    0.8 & 57.5 [48.9, 66.1] & 59.4 [46.2, 72.6] & $+1.9$ [$-4.5$, $+8.3$] & 42.6 \\
    \multicolumn{5}{@{}l}{\emph{PickScore}} \\
    0.2 & 49.2 [39.2, 59.3] & 50.0 [$-0.2$, 100.1] & $+0.8$ [$-43.1$, $+44.6$] & 46.7 \\
    0.4 & 55.4 [45.8, 65.1] & 56.9 [54.8, 59.1] & $+1.5$ [$-7.7$, $+10.7$] & 51.2 \\
    0.8 & 57.0 [50.5, 63.5] & 56.5 [46.3, 66.6] & $-0.5$ [$-4.3$, $+3.3$] & 49.2 \\
    \bottomrule
  \end{tabular}
\end{table}

\subsection{Fresh prompts: every metric and diversity}
\label{sec:s-fresh}

\paragraph{Protocol.}
The 992 fresh prompts, 793 PartiPrompts~\cite{yu2022parti} and 199 DrawBench~\cite{saharia2022imagen} prompts, were never used in development or training.
Each policy and the base model generate one image per prompt from the same noise, and a policy wins a prompt when its image scores higher (ties count one half).
Six automatic scorers are read: HPSv2.1~\cite{wu2023hpsv2}, PickScore~\cite{kirstain2023pickapic}, ImageReward~\cite{xu2023imagereward}, the LAION aesthetic predictor v2~\cite{schuhmann2022laion}, CLIPScore~\cite{hessel2021clipscore} with ViT-L/14 and UnifiedReward-qwen-7b's point score~\cite{wang2025unifiedreward}.
The same code also re-scored the 1{,}200 pinned evaluation prompts.
Contrasts follow \cref{sec:s-train-eval}; \cref{tab:s-fresh} gives every metric for the \List, \PWC and \Rating policies.

\begin{table*}[t]
  \centering
  \scriptsize
  \setlength{\tabcolsep}{4pt}
  \caption{\textbf{Every automatic metric on fresh and pinned prompts.} Win rate against the base (\%), mean (SD) over 16 seeds per arm. Contrasts: unpaired, the wider of the two-level bootstrap and Welch $t$, 95\%, two-sided $p$.}
  \label{tab:s-fresh}
  \begin{tabular}{@{}llllll@{}}
    \toprule
    metric & \List & \PWC & \Rating & \PWC $-$ \List & \Rating $-$ \List \\
    \midrule
    \multicolumn{6}{@{}l}{\emph{992 fresh prompts}} \\
    HPSv2.1 & 51.95 (4.87) & 62.36 (4.40) & 53.56 (6.23) & $+10.42$ [$+7.07$, $+13.77$], $p$ .0004 & $+1.61$ [$-2.44$, $+5.66$], $p$ .42 \\
    PickScore & 53.12 (1.90) & 57.81 (2.19) & 53.25 (3.16) & $+4.70$ [$+2.76$, $+6.62$], $p$ .0004 & $+0.13$ [$-1.77$, $+2.03$], $p$ .89 \\
    ImageReward & 57.57 (2.30) & 61.85 (2.04) & 55.93 (2.54) & $+4.28$ [$+2.35$, $+6.22$], $p$ .0004 & $-1.64$ [$-3.39$, $+0.11$], $p$ .065 \\
    LAION aesthetic v2 & 46.69 (2.90) & 53.42 (3.48) & 46.73 (2.02) & $+6.73$ [$+4.16$, $+9.27$], $p$ .0004 & $+0.04$ [$-1.77$, $+1.85$], $p$ .96 \\
    CLIPScore (ViT-L/14) & 49.73 (1.80) & 50.42 (1.88) & 50.20 (0.63) & $+0.69$ [$-1.11$, $+2.44$], $p$ .45 & $+0.47$ [$-0.53$, $+1.47$], $p$ .34 \\
    UnifiedReward-qwen-7b & 50.55 (0.82) & 51.20 (0.76) & 50.51 (0.64) & $+0.65$ [$-0.14$, $+1.44$], $p$ .10 & $-0.03$ [$-0.56$, $+0.50$], $p$ .91 \\
    \midrule
    \multicolumn{6}{@{}l}{\emph{1{,}200 pinned prompts (re-scored)}} \\
    HPSv2.1 & 55.02 (6.20) & 66.41 (5.00) & 56.20 (6.00) & $+11.39$ [$+7.32$, $+15.51$], $p$ .0004 & $+1.18$ [$-3.23$, $+5.59$], $p$ .59 \\
    PickScore & 53.03 (3.03) & 59.07 (1.98) & 53.33 (3.24) & $+6.04$ [$+3.98$, $+8.14$], $p$ .0004 & $+0.30$ [$-1.97$, $+2.57$], $p$ .79 \\
    ImageReward & 57.19 (2.88) & 61.72 (2.12) & 54.98 (2.97) & $+4.53$ [$+2.46$, $+6.58$], $p$ .0004 & $-2.20$ [$-4.31$, $-0.09$], $p$ .041 \\
    LAION aesthetic v2 & 48.44 (3.49) & 53.50 (3.09) & 48.65 (3.16) & $+5.06$ [$+2.56$, $+7.52$], $p$ .0004 & $+0.21$ [$-2.19$, $+2.61$], $p$ .86 \\
    CLIPScore (ViT-L/14) & 49.66 (1.64) & 50.72 (1.11) & 50.97 (1.57) & $+1.06$ [$-0.48$, $+2.61$], $p$ .17 & $+1.31$ [$+0.15$, $+2.47$], $p$ .028 \\
    UnifiedReward-qwen-7b & 50.84 (0.85) & 51.78 (0.67) & 50.83 (0.42) & $+0.95$ [$+0.20$, $+1.71$], $p$ .014 & $-0.01$ [$-0.50$, $+0.48$], $p$ .97 \\
    \bottomrule
  \end{tabular}
\end{table*}

\paragraph{Results.}
On fresh prompts all six \PWC$-$\List point estimates are positive, and four intervals exclude 0: HPSv2.1, PickScore, ImageReward and the aesthetic score.
After a step-down adjustment over the six metrics their $p$-values are 0.0024 each, and those of CLIPScore and UnifiedReward 0.45 and 0.20.
The seed-paired contrasts agree (HPSv2.1 $+10.42$ [$+6.58$, $+14.25$]).
The gain holds on both prompt sources: HPSv2.1 $+9.93$ [$+5.12$, $+14.64$] on DrawBench ($n=199$) and $+10.54$ [$+7.23$, $+14.02$] on PartiPrompts ($n=793$); PickScore $+4.15$ [$+0.54$, $+7.61$] and $+4.83$ [$+2.81$, $+6.88$]; ImageReward $+4.54$ [$+1.03$, $+8.28$] and $+4.22$ [$+2.27$, $+6.32$]; aesthetic $+7.99$ [$+3.75$, $+11.94$] and $+6.41$ [$+3.85$, $+9.08$].
CLIPScore does not move ($+1.98$ [$-1.57$, $+5.72$] and $+0.36$ [$-1.53$, $+2.31$]).
On the same 992 prompts the \ListAvg and \PairOne policies won 58.01\% and 58.03\% of HPSv2.1 comparisons against the base and 54.16\% and 55.37\% of PickScore comparisons (\cref{tab:s-fresh-corr}); \PWC minus \ListAvg was $+4.35$ [$+1.05$, $+7.66$] and \PWC minus \PairOne $+4.34$ [$+0.81$, $+7.97$] HPSv2.1 points.
Against both arms the PickScore, ImageReward and aesthetic intervals exclude 0 as well, as do the pinned PickScore contrasts ($+4.14$ [$+2.10$, $+6.27$] and $+2.23$ [$+0.27$, $+4.13$]); the seed-paired contrasts give the same pattern.
The GenEval scores of these two arms are in \cref{tab:s-geneval}.
The \PRP- and Tournament-GRPO-trained policies won 59.4\% and 56.6\% of the HPSv2.1 comparisons on the fresh prompts (\cref{tab:s-fresh-corr}, lower block).
All eight \PWC contrasts with them on HPSv2.1, PickScore, ImageReward and the aesthetic score are positive.
Against \PRP the PickScore interval excludes 0 ($+1.8$ [$+0.2$, $+3.4$]) and the HPSv2.1, ImageReward and aesthetic intervals include it ($+3.0$ [$-0.4$, $+6.4$], $+1.3$ [$-0.3$, $+2.9$] and $+0.8$ [$-1.5$, $+3.1$]); against Tournament-GRPO the HPSv2.1, PickScore and ImageReward intervals exclude 0 and the aesthetic interval includes it ($+1.7$ [$-0.6$, $+4.0$]).

\begin{table*}[t]
  \centering
  \footnotesize
  \setlength{\tabcolsep}{4pt}
  \caption{\textbf{Fresh-prompt metrics of the \ListAvg-, \PairOne-, \PRP- and Tournament-GRPO-trained policies.} Win rate against the base (\%) on the 992 fresh prompts, mean (SD) over 16 seeds; \List and \PWC are in \cref{tab:s-fresh}. Contrasts as in \cref{tab:s-fresh}: unpaired, the wider of the two-level bootstrap and Welch $t$, 95\%, upper block with the two-sided $p$ before adjustment. On the pinned prompts these arms were scored with HPSv2.1 and PickScore only (\cref{tab:s-seeds}).}
  \label{tab:s-fresh-corr}
  \begin{tabular}{@{}lllll@{}}
    \toprule
    metric & \ListAvg & \PairOne & \PWC $-$ \ListAvg & \PWC $-$ \PairOne \\
    \midrule
    HPSv2.1 & 58.01 (4.74) & 58.03 (5.32) & $+4.35$ [$+1.05$, $+7.66$], $p$ .012 & $+4.34$ [$+0.81$, $+7.97$], $p$ .019 \\
    PickScore & 54.16 (2.08) & 55.37 (2.62) & $+3.66$ [$+1.64$, $+5.60$], $p$ .0007 & $+2.44$ [$+0.36$, $+4.63$], $p$ .027 \\
    ImageReward & 58.58 (2.40) & 59.76 (1.64) & $+3.27$ [$+1.41$, $+5.26$], $p$ .0016 & $+2.09$ [$+0.38$, $+3.96$], $p$ .024 \\
    LAION aesthetic v2 & 49.87 (3.08) & 50.69 (2.63) & $+3.55$ [$+1.02$, $+6.10$], $p$ .0077 & $+2.73$ [$+0.32$, $+5.18$], $p$ .029 \\
    CLIPScore (ViT-L/14) & 49.61 (1.44) & 50.44 (1.60) & $+0.81$ [$-0.87$, $+2.51$], $p$ .34 & $-0.03$ [$-1.74$, $+1.66$], $p$ .99 \\
    UnifiedReward-qwen-7b & 50.76 (0.65) & 50.99 (0.79) & $+0.44$ [$-0.28$, $+1.21$], $p$ .24 & $+0.21$ [$-0.56$, $+0.97$], $p$ .58 \\
    \midrule
    metric & \PRP & Tournament-GRPO & \PWC $-$ \PRP & \PWC $-$ Tournament-GRPO \\
    \midrule
    HPSv2.1 & 59.4 (5.10) & 56.6 (5.45) & $+3.0$ [$-0.4$, $+6.4$] & $+5.8$ [$+2.2$, $+9.4$] \\
    PickScore & 56.0 (2.20) & 54.4 (2.45) & $+1.8$ [$+0.2$, $+3.4$] & $+3.4$ [$+1.7$, $+5.1$] \\
    ImageReward & 60.6 (2.25) & 58.8 (2.35) & $+1.3$ [$-0.3$, $+2.9$] & $+3.1$ [$+1.5$, $+4.7$] \\
    LAION aesthetic v2 & 52.6 (2.80) & 51.7 (2.70) & $+0.8$ [$-1.5$, $+3.1$] & $+1.7$ [$-0.6$, $+4.0$] \\
    \bottomrule
  \end{tabular}
\end{table*}

\begin{figure*}[t]
  \centering
  \includegraphics[width=\textwidth]{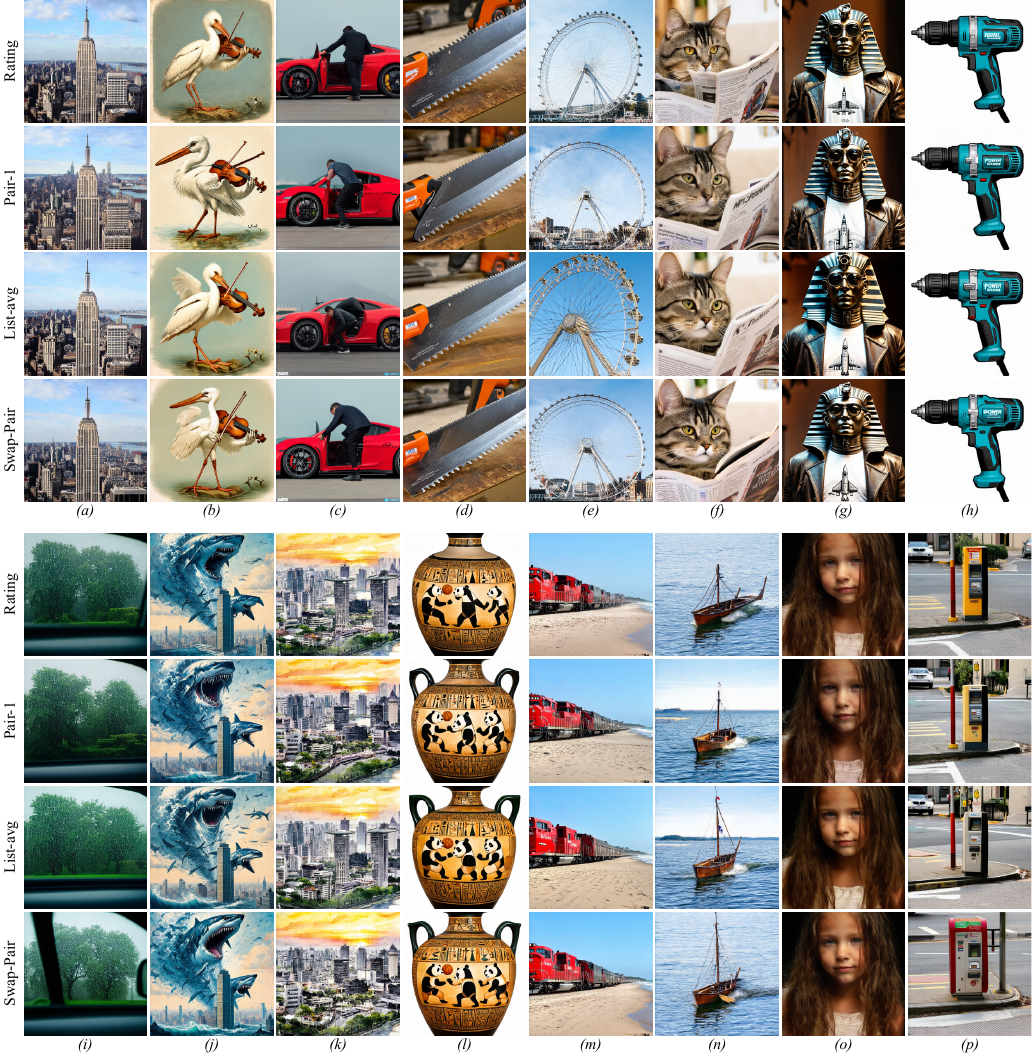}
  \caption{\textbf{Unselected samples of the \Rating-, \PairOne-, \ListAvg- and \PWC-trained policies.} Sixteen pairs of a fresh prompt and a training seed (1 to 5), drawn at random with a fixed seed from the 992 fresh prompts and shown without any selection on scores or images. Each column shares the prompt, the initial noise and the training seed. Prompts: (a)~an overhead view of the Empire State Building; (b)~a stork playing a violin; (c)~a tall man stooping down to enter a low red sports car; (d)~a handsaw; (e)~the Millennium Wheel; (f)~a cat reading a newspaper; (g)~A portrait of a metal statue of a pharaoh wearing steampunk glasses and a leather jacket over a white t-shirt that has a drawing of a space shuttle on it; (h)~a drawing of a power drill; (i)~trees seen through a car window on a rainy day; (j)~A tornado made of sharks crashing into a skyscraper. painting in the style of Hokusai; (k)~Downtown Singapore at sunrise. detailed ink wash; (l)~A photo of an Athenian vase with a painting of pandas playing basketball in the style of Egyptian hieroglyphics; (m)~a red train is coming down the beach; (n)~A small vessel propelled on water by oars, sails, or an engine; (o)~a girl; (p)~A machine next to a parking space in a street, into which the driver puts money so as to be authorized to park the vehicle for a particular length of time.}
  \label{fig:s-samples}
\end{figure*}

\paragraph{Diversity.}
Diversity is the mean pairwise $1-\cos$ of CLIP ViT-L/14 image embeddings among four images from distinct noises, on 100 prompts.
The base model scores 0.1373; the \List, \PWC and \Rating policies 0.1320, 0.1303 and 0.1333 (0.961, 0.949 and 0.971 of the base), the \ListAvg and \PairOne policies 0.1316 and 0.1322 (0.959 and 0.963), the \PRP and Tournament-GRPO policies 0.1310 and 0.1318 (0.954 and 0.960), and the least diverse single policy 0.895 of the base.
\PWC minus \List is $-0.0016$ [$-0.0048$, $+0.0014$] ($p=0.28$; seed-paired $-0.0016$ [$-0.0048$, $+0.0016$]): no sign of a collapse in diversity relative to \List.
\Cref{fig:s-samples} shows unselected samples of the \Rating-, \PairOne-, \ListAvg- and \PWC-trained policies.

\subsection{GenEval by task}
\label{sec:s-geneval}

GenEval's~\cite{ghosh2023geneval} official evaluator scores 553 prompts with four images each for every policy (\cref{tab:s-geneval}).
\PWC exceeds \List overall by $+0.98$ points [$+0.60$, $+1.37$] (Welch $t(25.5)=5.24$, $p<0.0001$, 16 against 16 runs), \ListAvg by $+0.91$ [$+0.52$, $+1.30$], \PairOne by $+0.81$ [$+0.42$, $+1.20$] and \Rating by $+1.11$ [$+0.72$, $+1.50$]; the \PRP and Tournament-GRPO policies reach 0.686 and 0.682 overall, 0.5 and 0.9 points below \PWC (\mainref{tab:training}).
After a step-down adjustment over the six tasks, two objects ($+1.52$ [$+0.60$, $+2.43$], adjusted $p=0.011$) and colors ($+1.15$ [$+0.51$, $+1.78$], adjusted $p=0.006$) differ from \List.
Every arm improves on the base overall: \List $+1.18$ [$+0.84$, $+1.51$], \ListAvg $+1.26$, \PairOne $+1.36$, Tournament-GRPO $+1.2$, \PRP $+1.6$, \PWC $+2.17$ [$+1.96$, $+2.39$] and \Rating $+1.05$ [$+0.76$, $+1.34$] points.
Our base score, 0.6696, is 0.0396 above the published 0.63, a difference we report rather than correct.

\begin{table*}[t]
  \centering
  \scriptsize
  \setlength{\tabcolsep}{1.7pt}
  \caption{\textbf{GenEval by task.} Mean score over runs (0--1); differences in points ($100\times$), Welch 95\% intervals over runs, 16 against 16 runs; $p$: two-sided, \PWC$-$\List. Base (one generation): 0.6696 overall.}
  \label{tab:s-geneval}
  \begin{tabular}{@{}llllllllllll@{}}
    \toprule
    task & \List & \ListAvg & \PairOne & \PWC & \Rating & \PWC $-$ \List & $p$ & \Rating $-$ \List & \List $-$ base & \PWC $-$ base & \Rating $-$ base \\
    \midrule
    overall & 0.6814 & 0.6822 & 0.6832 & 0.6913 & 0.6802 & $+0.98$ [$+0.60$, $+1.37$] & $<$.0001 & $-0.13$ [$-0.43$, $+0.17$] & $+1.18$ & $+2.17$ & $+1.05$ \\
    single object & 0.9977 & 0.9970 & 0.9972 & 0.9967 & 0.9956 & $-0.10$ [$-0.30$, $+0.10$] & .33 & $-0.20$ [$-0.39$, $-0.01$] & $+0.08$ & $-0.02$ & $-0.12$ \\
    two objects & 0.8578 & 0.8625 & 0.8641 & 0.8730 & 0.8606 & $+1.52$ [$+0.60$, $+2.43$] & .002 & $+0.28$ [$-0.44$, $+1.00$] & $+2.70$ & $+4.21$ & $+2.98$ \\
    counting & 0.6092 & 0.6060 & 0.6110 & 0.6191 & 0.5888 & $+0.99$ [$-0.19$, $+2.17$] & .096 & $-2.04$ [$-2.72$, $-1.36$] & $+2.79$ & $+3.78$ & $+0.75$ \\
    colors & 0.8200 & 0.8240 & 0.8228 & 0.8315 & 0.8250 & $+1.15$ [$+0.51$, $+1.78$] & .001 & $+0.50$ [$+0.17$, $+0.83$] & $+0.62$ & $+1.76$ & $+1.12$ \\
    position & 0.2477 & 0.2510 & 0.2498 & 0.2585 & 0.2585 & $+1.08$ [$+0.25$, $+1.91$] & .013 & $+1.08$ [$+0.64$, $+1.52$] & $-0.48$ & $+0.60$ & $+0.60$ \\
    color attribution & 0.5563 & 0.5527 & 0.5543 & 0.5688 & 0.5525 & $+1.26$ [$+0.04$, $+2.47$] & .043 & $-0.38$ [$-0.97$, $+0.21$] & $+1.38$ & $+2.63$ & $+1.00$ \\
    \bottomrule
  \end{tabular}
\end{table*}

\subsection{The Rating arm}
\label{sec:s-rating}

\Rating is invariant to other candidates by construction, since each image is judged alone, and it trains above the random-reward controls, but \PWC leads it on each use we measured.
As a reward on the audit groups its within-group $\rho$ with HPSv2.1 is 0.33 (\List 0.36, \PWC 0.51), and as a selector it recovers 43.1\% [34.9, 51.0] of the oracle best-of-8 gain on the 256 selection groups, 21.0 [13.2, 29.3] points below the cover winner (47.7\% on the audit groups, \cref{sec:s-audit}).
Its 16 trained policies reach 56.20\% on the pinned prompts and 53.56\% on fresh prompts, above the random-reward controls (36.8 to 49.5\%) and level with \List (55.02\% and 51.95\%; $+1.18$ [$-3.23$, $+5.59$] and $+1.61$ [$-2.44$, $+5.66$]); on PickScore it is level with \List as well.
On GenEval it is level with \List overall ($-0.13$ [$-0.43$, $+0.17$]), lower on counting ($-2.04$ [$-2.72$, $-1.36$]) and higher on position ($+1.08$ [$+0.64$, $+1.52$]).
As a judge of human-ranked triples it beats \List by about one point of winner agreement and 2.4--2.7 points of pair agreement (\cref{tab:s-k3}), but trails \PWC on pairs whose ratings are close (\cref{sec:s-variants}).
\PWC leads it by $+10.21$ [$+6.22$, $+14.20$] on the pinned prompts and $+8.80$ [$+4.91$, $+12.69$] on fresh prompts (16 against 16 seeds; \mainref{tab:training}).

\subsection{GenAI-Rank and GenAI-Bench}
\label{sec:s-genai}

\paragraph{Best-of-9 selection on GenAI-Rank.}
GenAI-Rank~\cite{li2024genaibench} holds nine SD-XL and nine DALL-E~3 images for each of 800 prompts, with three human ratings per image; the oracle gain is the gain in mean human rating of the best-rated image over a random one (\mainref{sec:validity}).
The \PWC pick recovers 38.0\% \ci{33.9}{42.0} of it with q8 and 46.4\% \ci{42.5}{50.4} with q32.
Paired over prompts, it leads the pick of one \List message by $+10.4$ \ci{+6.4}{+14.4} points on q8 and $+7.9$ \ci{+4.2}{+11.6} on q32, \ListAvg by $+4.2$ \ci{+0.5}{+7.8} and $+3.9$ \ci{+0.4}{+7.3}, \PairOne by $+7.3$ \ci{+3.6}{+11.0} and $+8.5$ \ci{+5.6}{+11.4}, the knockout by $+8.2$ \ci{+4.5}{+11.7} and $+8.5$ \ci{+5.6}{+11.6}, and Tournament-GRPO's structure by $+7.6$ \ci{+5.0}{+10.2} and $+8.2$ \ci{+5.8}{+10.4}.
With q8, one \List message recovers 27.6\% of the gain, the knockout 29.8\%, Tournament-GRPO's structure 30.4\%, \PairOne 30.7\%, \Rating 31.0\%, permutation self-consistency (\textsf{PSC}) 33.1\%, \ListAvg 33.8\% and \PRP 35.1\% (\mainref{tab:protocols}).
The \PWC pick leads \Rating by $+7.0$ \ci{+3.1}{+11.0} and \textsf{PSC} by $+4.9$ \ci{+1.3}{+8.6}, and \PRP by $+2.9$ \ci{-0.4}{+6.2}, an interval that includes 0.
Against the picks of the published scorers PickScore, HPSv2.1, ImageReward and CLIPScore, it leads by $+18.5$ \ci{+12.8}{+24.3}, $+22.8$ \ci{+17.0}{+28.3}, $+22.9$ \ci{+17.2}{+28.6} and $+32.4$ \ci{+26.6}{+38.3} points with q8, and by $+27.0$ \ci{+21.5}{+32.5}, $+31.3$ \ci{+25.7}{+36.8}, $+31.4$ \ci{+25.8}{+36.7} and $+40.9$ \ci{+35.1}{+46.8} with q32.
With q32 it also leads the pick of VQAScore, the benchmark's own alignment metric, by $+6.5$ \ci{+1.4}{+11.6}.
Shown the same nine images in a second order, one \List message changes its pick in 63.9\% of the 1{,}600 groups on q8 and in 55.0\% on q32, and \PairOne in 62.8\% and 52.1\%, while the \PWC pick is the same in every group.
When the \List pick changes, the two picked images differ by $0.48$ \ci{0.44}{0.52} points of mean human rating on q8 and by $0.45$ \ci{0.41}{0.49} on q32, on a five-point scale where the best-rated image exceeds a random one by $0.62$.

\paragraph{The trained generative models on GenAI-Bench.}
On the 800 GenAI-Bench prompts of GenAI-Rank, HPSv2.1 prefers the \PWC-trained model to the base model on 62.2\% of the prompts and the \List-trained model on 53.9\%, a lead of $+8.2$ \ci{+4.5}{+11.8}\,\pp over 16 seeds each (\mainref{sec:training}).
PickScore and ImageReward agree ($+3.5$ \ci{+1.1}{+5.9} and $+4.3$ \ci{+2.1}{+6.7}), and VQAScore, the alignment metric of the benchmark, prefers the \PWC-trained model to the base model on 52.6\% of the prompts and the \List-trained one on 48.4\%, a lead of $+4.2$ \ci{+2.4}{+6.0}\,\pp.
The other five q8 rewards stay below \PWC on all four scores (\cref{tab:s-genaibench}); in HPSv2.1 win rate \PWC leads \Rating by $+7.1$ \ci{+3.5}{+10.7}, Tournament-GRPO by $+5.8$ \ci{+2.2}{+9.4} and \ListAvg by $+4.0$ \ci{+0.6}{+7.4}\,\pp, and \PairOne by $+3.9$ \ci{-0.1}{+7.9} and \PRP by $+2.9$ \ci{-0.6}{+6.4}, two intervals that include 0.

\begin{table}[t]
  \centering
  \scriptsize
  \setlength{\tabcolsep}{1.6pt}
  \caption{\textbf{Trained generative models on the 800 GenAI-Bench prompts.} Win rate against the base model (\%), mean over 16 seeds; last column: \PWC minus the arm in HPSv2.1 win rate, 95\%. Tourn.: Tournament-GRPO.}
  \label{tab:s-genaibench}
  \begin{tabular}{@{}llllll@{}}
    \toprule
    reward & HPSv2.1 & PickScore & ImageReward & VQAScore & \PWC $-$ arm \\
    \midrule
    \Rating & 55.1 & 51.6 & 56.4 & 49.6 & $+7.1$ [$+3.5$, $+10.7$] \\
    \List & 53.9 & 51.5 & 57.2 & 48.4 & $+8.2$ [$+4.5$, $+11.8$] \\
    \ListAvg & 58.2 & 52.4 & 58.3 & 50.1 & $+4.0$ [$+0.6$, $+7.4$] \\
    \PairOne & 58.3 & 52.9 & 58.6 & 50.4 & $+3.9$ [$-0.1$, $+7.9$] \\
    Tourn. & 56.4 & 52.3 & 58.0 & 49.8 & $+5.8$ [$+2.2$, $+9.4$] \\
    \PRP & 59.3 & 53.6 & 59.6 & 51.3 & $+2.9$ [$-0.6$, $+6.4$] \\
    \PWC & 62.2 & 55.0 & 61.5 & 52.6 & -- \\
    \bottomrule
  \end{tabular}
\end{table}

\subsection{Published rewards, longer training, public batch, people and FLUX.1-dev}
\label{sec:s-further}

\paragraph{Published rewards and the second judge.}
The Pref-GRPO arm uses the reward that Pref-GRPO~\cite{wang2025prefgrpo} releases: its judge, its pair prompt, its answer parsing and its within-group win rate, with each unordered pair asked once in the released order.
\PWC on the same judge asks each pair in both orders with the same pair prompt and answer parsing, and rewards each rollout with its Copeland score over the group (\mainref{sec:reward}).
The scorer arms reward each rollout with its scorer's score against the prompt, normalised within the group as in Flow-GRPO~\cite{liu2025flowgrpo}.
Every arm shares the recipe, the prompts and seeds 1 to 8 of the q8 study, so it is seed-paired with the q8 arms.
On the 128 audit groups of \cref{sec:s-audit}, here judged by the UnifiedReward judge, the within-group $\rho$ with HPSv2.1 is 0.38 for the reward released with Pref-GRPO and 0.40 for \PWC on the same judge, computed as for the q8 readouts in \mainref{tab:protocols}.
\Cref{tab:s-further} gives the contrasts behind \mainref{tab:training}.
On the UnifiedReward judge, \PWC has the higher mean than Pref-GRPO on five of the six yardsticks of \mainref{tab:training}, all but GenEval ($-0.1$ [$-0.7$, $+0.5$]), and none of its intervals in \cref{tab:s-further} and the table note excludes 0.
\PWC on q8 has the higher mean than each scorer reward on every yardstick that no arm optimises; the intervals exclude 0 against the PickScore and HPSv2.1 rewards on all three yardsticks and against the UnifiedReward score on ImageReward, and include 0 for GenEval and VQAScore against the UnifiedReward score.

\begin{table*}[t]
  \centering
  \scriptsize
  \setlength{\tabcolsep}{2.4pt}
  \caption{\textbf{Contrasts with the published rewards} (win-rate points, GenEval points; unpaired, the wider of a Welch and a two-level bootstrap 95\% interval; 16 or 8 seeds against 8).}
  \label{tab:s-further}
  \begin{tabular}{@{}llll@{}}
    \toprule
    contrast & ImageReward & GenEval & VQAScore \\
    \midrule
    \PWC (q8) $-$ PickScore reward & $+2.3$ [$+0.4$, $+4.2$] & $+1.6$ [$+0.9$, $+2.3$] & $+3.4$ [$+1.2$, $+5.6$] \\
    \PWC (q8) $-$ HPSv2.1 reward & $+2.8$ [$+0.7$, $+4.9$] & $+1.8$ [$+1.0$, $+2.6$] & $+4.0$ [$+1.5$, $+6.5$] \\
    \PWC (q8) $-$ UnifiedReward score & $+3.2$ [$+1.4$, $+5.0$] & $+0.4$ [$-0.1$, $+0.9$] & $+1.1$ [$-1.2$, $+3.4$] \\
    \PWC $-$ Pref-GRPO, same judge & $+0.8$ [$-1.6$, $+3.2$] & $-0.1$ [$-0.7$, $+0.5$] & $+0.7$ [$-2.0$, $+3.4$] \\
    \bottomrule
  \end{tabular}
  \par\smallskip\parbox{\textwidth}{\scriptsize On the UnifiedReward judge, \PWC $-$ Pref-GRPO is also $+2.2$ [$-3.5$, $+7.9$] in pinned HPSv2.1 win rate and $+2.0$ [$-2.2$, $+6.2$] on the fresh prompts.}
\end{table*}

\paragraph{Longer training and equal compute.}
Seeds 17 to 24 of \List, \ListAvg, \PairOne, \PRP and \PWC run to step 600 with the recipe unchanged and a constant learning rate, and are evaluated on the 1{,}200 pinned prompts at steps 100, 200, 400 and 600, with no checkpoint selection.
Their step-200 checkpoints are exchangeable with those of seeds 1 to 16, so each of these arms has 24 seeds at step 200, and the step-200 contrasts below pool seeds 1 to 16 with seeds 17 to 24: \PWC leads \List by $+11.4$ [$+8.1$, $+14.7$], \ListAvg by $+4.6$ [$+1.6$, $+7.7$] ($p=0.004$), \PairOne by $+3.4$ [$+0.5$, $+6.4$] ($p=0.023$) and \PRP by $+2.6$ [$-0.4$, $+5.6$] ($p=0.09$; with seeds 1 to 16 only, $+3.2$ [$-0.5$, $+6.9$], $p=0.09$).
With the 16 seeds of \Rating and Tournament-GRPO, the six step-200 contrasts of \PWC, against \List, \Rating, Tournament-GRPO, \ListAvg, \PairOne and \PRP, have $p=0.0004$, $0.0004$, $0.0035$, $0.004$, $0.023$ and $0.09$; after a Holm--Bonferroni correction their adjusted $p$-values are $0.0024$, $0.0024$, $0.014$, $0.014$, $0.046$ and $0.09$, so the five contrasts against \List, \Rating, Tournament-GRPO, \ListAvg and \PairOne stay below 0.05 and the contrast against \PRP does not.
\Cref{tab:s-horizon} gives the means and SDs of seeds 17 to 24 at the four checkpoints.
At step 600 \PWC leads \List by $+15.7$ [$+9.3$, $+22.0$], \ListAvg by $+7.1$ [$+1.3$, $+12.8$] and \PairOne by $+6.2$ [$+1.0$, $+11.3$], and \PRP by $+4.4$ [$-1.2$, $+9.9$], an interval that includes 0 (8 against 8 seeds).
One training step takes 43.6\,s with the \List reward and 81.2\,s with the \PWC reward on one GPU, of which generation and the policy update take 41.8\,s and the judge 1.8\,s and 39.4\,s.
Six hundred \List steps therefore take 7.3 GPU-hours and two hundred \PWC steps 4.5, and on seeds 17 to 24 the \List policies at step 600 trail the \PWC policies at step 200 by 8.5 points; \PWC exceeds \List's step-600 mean already at step 100.
A step takes 55.1\,s with the \ListAvg reward (judge 13.3\,s, 1.66\,s for each of the eight groups), 61.5\,s with the \PairOne reward (judge 19.7\,s), 56.6\,s with the Tournament-GRPO reward (judge 14.8\,s), and 81.2\,s with the \PRP reward, the same as with \PWC, whose 56 messages it shares.
At \PWC's 4.5 GPU-hours, \ListAvg runs 295 steps and \PairOne 264; interpolated linearly between their step-200 and step-400 means over seeds 17 to 24, these budgets give 63.6\% and 65.0\%, 3.5 and 2.1 points below \PWC's step-200 mean of 67.1\% over the same seeds (descriptive).

\begin{table}[t]
  \centering
  \scriptsize
  \setlength{\tabcolsep}{4pt}
  \caption{\textbf{Longer training} (seeds 17 to 24; pinned HPSv2.1 win rate against the base, \%, on the 1{,}200 pinned prompts; mean (SD over the eight seeds) at each checkpoint).}
  \label{tab:s-horizon}
  \begin{tabular}{@{}lllll@{}}
    \toprule
    arm & step 100 & step 200 & step 400 & step 600 \\
    \midrule
    \List & 54.1 (6.5) & 55.6 (6.5) & 57.3 (6.5) & 58.6 (6.8) \\
    \ListAvg & 59.8 (5.7) & 61.9 (5.7) & 65.4 (5.7) & 67.2 (5.9) \\
    \PairOne & 60.2 (5.1) & 64.6 (5.1) & 66.0 (5.1) & 68.1 (5.1) \\
    \PRP & 60.6 (5.5) & 65.7 (5.5) & 67.5 (5.5) & 69.9 (5.6) \\
    \PWC & 63.9 (4.8) & 67.1 (4.8) & 71.2 (4.8) & 74.3 (4.6) \\
    \bottomrule
  \end{tabular}
\end{table}

\paragraph{Public batch and KL weight.}
\label{sec:s-publicbatch}
Four arms are trained at the batch and KL weight of Flow-GRPO's public SD3.5 configuration~\cite{liu2025flowgrpo}: SD3.5-Medium with LoRA at 512\,px, 1{,}152 images per step as 144 groups of eight rollouts, KL weight 0.04 and 1{,}000 steps, with 3 seeds per arm and the recipe of \cref{sec:s-recipe} otherwise.
The arms are the PickScore reward as in Flow-GRPO, Pref-GRPO as released, and \ListAvg and \PWC read from q8.
Each run is evaluated once, at step 1{,}000, on the 1{,}200 pinned prompts, with no checkpoint selection: mean HPSv2.1 ($\times100$), mean PickScore, GenEval ($\times100$) and the HPSv2.1 win rate against the base model's image from the same noise (\mainref{tab:scale}).
\PWC reaches 31.47, 23.05, 72.3 and 79.1, the highest HPSv2.1, GenEval and win rate of the four arms; the PickScore reward has the highest PickScore, 23.38, the score it optimises.
In win rate \PWC leads \ListAvg by $+6.7$ [$+0.9$, $+12.5$] points and Pref-GRPO by $+5.2$ [$-0.6$, $+11.0$], an interval that includes 0, and in GenEval it leads Pref-GRPO by $+1.9$ [$+0.4$, $+3.4$] points (3 against 3 seeds, Welch).
At this batch the public configuration's KL weight trains, which it did not at the 64 images per step of the pilot (\cref{sec:s-pilot}): as a diagnostic, the trainer with an HPSv2.1 reward at KL weight 0.04 reaches an HPSv2.1 win rate of 88.5\% against the base after 300 steps (one seed).

\paragraph{People.}
Three hundred prompts were drawn at random from the 992 fresh prompts, with no metric involved.
The models are the step-200 checkpoints, except in the last comparison of \mainref{tab:human}, where both the \PWC- and the \List-trained models are the step-600 checkpoints of seeds 17 to 24.
For each comparison and prompt, both models generate one image from the same initial noise; an arm with $n$ seeds uses the adapter of its $((i \bmod n)+1)$-th training seed for prompt $i$, so that the result speaks for the arm and not for one adapter.
Five raters who had not seen the paper see the prompt and the two images side by side in random order and answer two questions, which image is better overall and which matches the prompt better; ``no preference'' counts one half.
Each comparison thus has 1{,}500 judgments per question; intervals come from a bootstrap over prompts and raters, and a rater agrees with the majority of the five in 76\% of the judgments.
The nine comparisons set the \PWC-trained model against the base model, the models trained with the \List, \ListAvg, \PairOne and \PRP rewards of q8, with Pref-GRPO's released reward and with the PickScore and HPSv2.1 rewards, and the \List-trained model at step 600 (\mainref{tab:human}).
Against the HPSv2.1-trained model, which wins the metric it optimises (\mainref{tab:training}), raters prefer the \PWC-trained image in 58.1\% [54.6, 61.6] of the overall judgments and in 59.7\% of the prompt-match judgments.
When both models train for 600 steps, they prefer it to the \List-trained image in 61.7\% [58.0, 65.4] and 58.0\%, against 58.6\% [55.0, 62.2] and 56.4\% at step 200.

\paragraph{Drift over longer training.}
For the 600-step runs we also track two signs of reward hacking on the pinned images, the mean HSV saturation and the diversity of \cref{sec:s-fresh} (four noises, 100 prompts). Saturation stays within 2\% of the base model for every arm (\PWC $+1.4$\%, \PRP $+1.2$\%, \List $+0.9$\%), diversity at step 600 is 0.94 of the base for \PWC and 0.95 for \List, and ImageReward and GenEval rise with the HPSv2.1 win rate at every checkpoint.

\paragraph{Absolute scores.}
\Cref{tab:s-absolute} gives the mean scores behind the win rates on the 1{,}200 pinned prompts; the base row is the base model's images from the same noise.

\begin{table*}[t]
  \centering
  \scriptsize
  \setlength{\tabcolsep}{3pt}
  \caption{\textbf{Mean scores on the 1{,}200 pinned prompts} (HPSv2.1 $\times$100; PickScore as in its release). Grey: the arm's own reward or a close relative.}
  \label{tab:s-absolute}
  \begin{tabular}{@{}lccc@{}}
    \toprule
    reward & HPSv2.1, 200 steps & PickScore, 200 steps & HPSv2.1, 600 steps \\
    \midrule
    base model & 29.79 & 22.51 & 29.79 \\
    \Rating (q8) & 29.92 & 22.55 & -- \\
    \List (q8) & 29.90 & 22.55 & 30.03 \\
    \ListAvg (q8) & 30.10 & 22.58 & 30.44 \\
    \PairOne (q8) & 30.11 & 22.63 & 30.47 \\
    Tournament-GRPO (q8) & 30.05 & 22.60 & -- \\
    \PRP (q8) & 30.20 & 22.66 & 30.71 \\
    \PWC (q8) & \textbf{30.29} & \textbf{22.70} & \textbf{30.86} \\
    Pref-GRPO, as released & 30.02 & 22.61 & -- \\
    \PWC, UnifiedReward judge & 30.09 & 22.64 & -- \\
    PickScore reward & \cellcolor{black!8}30.38 & \cellcolor{black!8}23.27 & -- \\
    HPSv2.1 reward & \cellcolor{black!8}31.12 & \cellcolor{black!8}22.84 & -- \\
    UnifiedReward score & 29.98 & 22.57 & -- \\
    \bottomrule
  \end{tabular}
\end{table*}

\paragraph{FLUX.1-dev.}
FLUX.1-dev~\cite{flux2024} is trained with LoRA of rank 32 at 512\,px under the FLUX.1-dev settings of the Flow-GRPO code base, for 200 steps of 8 prompts $\times$ 8 rollouts at KL weight 0.4, with \List, \PairOne and \PWC read from q8 and Pref-GRPO as released, 6 seeds each (\mainref{tab:scale}, lower block).
The seed SDs of the pinned HPSv2.1 win rate are 5.9 (\List), 4.8 (\PairOne), 5.3 (Pref-GRPO) and 4.4 (\PWC).
In pinned HPSv2.1 win rate, \PWC leads \List by $+9.3$ [$+2.4$, $+16.1$] and Pref-GRPO by $+6.8$ [$+0.4$, $+13.1$], and \PairOne by $+5.1$ [$-0.8$, $+11.0$], an interval that includes 0 (6 against 6 seeds).

\clearpage
\section{Protocol and deviations}
\label{sec:s-protocol}

\paragraph{Records.}
Every result row carries a hash of the code and of the data manifest that produced it, and a job refuses to write next to rows of another build.
Changes to the plan were written to a dated log before the rows they affect, together with what had been seen at that time; the log is released with the code.
It was not deposited with a third party, so its timing rests on the hashes and the dates.
Below, ``fixed before'' means that the analysis, its population and its reading were written down before the named data existed or were looked at; ``descriptive'' means that no decision rests on the analysis; ``post hoc'' means that the analysis was defined after the data were seen.

\paragraph{Timeline.}
\begin{itemize}[leftmargin=*,itemsep=1pt,topsep=2pt]
\item \emph{Before any development row.} The development and held-out split (300 and 100 sets), the decoy construction, the dominance closure, the matched re-render floor, the exclusion rules, and the attack endpoint with its reading: a searched excess of at least 10 points with a lower interval bound above 0, decided first on q32. Because two judges could decide, those intervals are 97.5\%. The closed-model check, its blinded route and its pass rule were fixed before any closed-model answer (14 October).
\item \emph{14 and 20 October.} q32 met the attack reading, $+22.7$ [17.1, 28.8] on the first run and $+22.4$ [$+17.1$, $+28.1$] when the whole chain was regenerated on new hardware on 20 October; the paper reports the second, with its 95\% interval [$+17.4$, $+27.4$]. The other four judges are reported with the same statistic but decided nothing.
\item \emph{18--19 October.} The secondary development analyses (one added image with grey and natural images, the content-free floor, far contenders, $N=8$, native answers, the defences other than order averaging and \PriDe, \ScreenList and its adaptive decoys) were fixed before their rows.
\item \emph{20 October.} The isolated readouts, the original candidate fix, did not meet their fixed reading on either judge (\cref{sec:s-iso}). The same day, before any of their rows were computed, the pairwise variants of \cref{sec:s-variants} and the rule that would choose among them on development data were fixed, as was the shared-position control.
\item \emph{21 October.} The development rule chose the gated variant; \PWC met the same attack conditions. Before any held-out set was judged, the held-out tests were rewritten for \PWC (\cref{sec:s-variants}), and the held-out sets were then judged once. The label-permutation control and the extension of \PWC to mst24 and ur7 were fixed the same day, before their rows.
\item \emph{22 October.} The reward audit, its consequence for the training design, and the training design itself (\List, \PWC and \Rating as rewards; contrasts against \List) were fixed before any training row.
\item \emph{23 October.} The pilot at KL weight 0.04 failed its learning check (\cref{sec:s-pilot}). The KL weight 0.4 was chosen with oracle rewards after the pilot's per-arm win rates had been seen, and the same design was rerun at 0.4.
\item \emph{24 October.} With 8 \List, 4 \PWC and 5 \Rating seeds the first stage gave \PWC$-$\List $+6.19$ [$-1.78$, $+14.15$] ($p=0.11$) and \Rating$-$\List $-4.80$ [$-13.49$, $+3.89$] ($p=0.24$): neither difference was established, and both are reported as such. The KL-weight sweep was fixed the same day.
\item \emph{25 October.} After the first stage had been seen, \List, \PWC and \Rating were extended to 16 seeds each, with the test on all 16 against 16 at a stricter threshold, $p<0.0294$, chosen so that the two looks together keep the false-positive rate near 0.07. The best-of-8 study on 256 new prompts was fixed the same day, before the prompts were drawn.
\item \emph{26 October.} The 16-seed result: \PWC$-$\List $+11.39$ [$+7.32$, $+15.51$], $p=0.0004$, below both thresholds.
\item \emph{27 October.} Fixed before any of their outputs: the fresh-prompt metrics and GenEval of the \List, \PWC and \Rating policies; the pair-query baselines (\PRP, \PairOne); the \ListAvg training arm and, later that day, the \PairOne training arm, each with 16 seeds and its contrast against \PWC; and the order-averaged attack on q32.
\item \emph{28 October.} Fixed before any of their outputs: the fresh-prompt metrics and GenEval of the \ListAvg and \PairOne policies; the certified-leader replay on stored matrices (\cref{sec:s-leader}), whose self-check was corrected before any count was seen to compare against the cover-winner selection it reproduces; and the recomputation of table intervals at one level from stored rows.
\end{itemize}

\paragraph{Held-out sets.}
The 100 held-out sets were held back from every development analysis: none of them was judged and their contenders entered no row before the held-out tests had been fixed; by the pool construction, their prompts' human-last images could appear as cross-prompt decoys in development pools.
They were judged once, under three one-sided tests with a step-down adjustment (\cref{sec:s-variants}).
The held-out tests first written for the isolated pair readout were replaced before any held-out row existed, because isolation had not met its development reading; the frozen report still printed the replaced lines, which read $-5.96$, $-9.92$ and $-8.00$ and were not confirmed.

\paragraph{Training seeds.}
The 16-seed \List, \PWC and \Rating arms consist of the first-stage seeds plus the second-stage seeds, analysed together with the test fixed on 25 October; the first-stage estimate above is not used for any claim.
The \ListAvg and \PairOne arms were added after the 16-seed result and run with the same 16 seeds; their contrasts with \PWC were fixed before their runs and are reported with intervals, not as tests of a hypothesis about the earlier result.
Seeds 17 to 24 of \List, \ListAvg, \PairOne and \PWC were added after the 16-seed results; the pooled step-200 contrasts and their Holm--Bonferroni correction over the four contrasts of \PWC, against \List, \Rating (16 seeds), \ListAvg and \PairOne, were fixed before these runs (\cref{sec:s-further}).
The \PRP and Tournament-GRPO arms were added later, with seeds 1 to 16 each and seeds 17 to 24 of \PRP run to step 600; before any of their runs, their contrasts with \PWC were fixed and the Holm--Bonferroni family was extended to the six step-200 contrasts of \PWC.
The KL-weight sweep is descriptive.

\paragraph{Studies added after the 16-seed results.}
The published-reward arms and the second judge, the 600-step runs, the human comparisons with their prompt draw, and the FLUX.1-dev arms (\cref{sec:s-further}) were designed after the results above had been seen; their arms, seeds, metrics and contrasts were fixed before any of their runs, and the drift measures and absolute scores of these runs are descriptive.
The four studies on public benchmarks, the searched excess and best-of-9 on GenAI-Rank, the MJ-Bench position test (\cref{sec:s-public}) and the GenAI-Bench scores of the trained models (\cref{sec:s-genai}), were fixed on 15 and 16 November, each before any of its rows.
Later still, each fixed before any of its runs or rows: the exchange test on the audit groups (\cref{sec:s-exchange}); the \PRP and Tournament-GRPO training arms with their metrics and contrasts (\cref{sec:s-recipe,sec:s-seeds}); the runs at the public batch and KL weight with their arms, seeds, metrics, contrasts and oracle diagnostic (\cref{sec:s-publicbatch}); the human study, enlarged to five raters and nine comparisons; and the new GenAI-Rank rows, best-of-9 for \Rating, permutation self-consistency and \PRP (\cref{sec:s-genai}).

\paragraph{Descriptive and post hoc analyses.}
Descriptive, fixed before their rows: the secondary attack rows (other judges, matched budgets, natural and content-free images, native answers, far contenders, $N=8$, transfer, text decoys, defences other than order averaging and \PriDe), the label-permutation and shared-position controls, the order-averaged attack, the permutation audit, the fresh-prompt metrics beyond HPSv2.1, GenEval by task, the certified-leader replay and the KL-weight sweep.
Descriptive and computed from existing results, with no new run: the equal-GPU-hours interpolation of \cref{sec:s-further} and the split of the cover winner's lead into the cover step and the margin tie-break (\cref{sec:s-leader}).
Post hoc: the aggregation enumeration and the stored-row residuals of \cref{tab:s-rules}, the rating-gap strata, the effect of one added image on the GRPO ranking (\cref{tab:s-grpo-rank}), the reward-path review of the pilot groups, best-of-8 on the audit groups (repeated on new prompts with the analysis fixed in advance), \PriDe, \PRP and \PairOne on the stored attack and audit rows, and the three published selection baselines added to Tab.~\ref{M-tab:protocols} of the main paper (\cref{sec:s-audit}).
Also descriptive: the random-reward control and the gain on training rollouts, diversity, the tie-break variants, the targeted flips and the by-slot rows, and the batched-serving replay.
Intervals recomputed at one level from stored rows are descriptive.

\paragraph{Exclusions and completeness.}
Unreadable readouts are excluded and counted; above 2\% of an arm's rows a problem is reported and the arm is kept.
A judge's attack row is complete only when every counted flip carries a complete dominance check.
ur7 has no verified decoy in 61 of 300 sets, above the 15\% that a deciding judge was allowed; it decided nothing and is reported with that count.
ivl38's first-token abstentions are disclosed with a sensitivity analysis (\cref{sec:s-ivl}).
FOCUS (one certificate problem in 28{,}827 rows) and \ScreenList on the adaptive decoys (7.32\% unreadable) are reported as incomplete.

\end{document}